\documentclass{article}

\usepackage{microtype}
\usepackage{graphicx}
\usepackage{subfigure}
\usepackage{booktabs} 
\usepackage{enumitem}
\usepackage[preprint]{neurips_2025}

\usepackage[utf8]{inputenc} 
\usepackage[T1]{fontenc}    
\usepackage{hyperref}       
\usepackage{url}            
\usepackage{booktabs}       
\usepackage{amsfonts}       
\usepackage{nicefrac}       
\usepackage{microtype}      
\usepackage{xcolor}         
\usepackage{pifont}
\usepackage{wrapfig}
\usepackage{comment}
\newcommand{\cxmark}{\ding{55}}
\newcommand{\cymark}{\ding{51}}

\title{Single-Pass Object-Focused Data Selection}

\author{Niclas Popp\\
Bosch Center for Artificial Intelligence\\
University of Tübingen\\
{\tt\small niclas.popp@de.bosch.com}
\And
Dan Zhang\\
Bosch Center for Artificial Intelligence\\
{\tt\small dan.zhang2@de.bosch.com}
\AND
Jan Hendrik Metzen \thanks{Work done while at Bosch Center for Artificial Intelligence} \\
IPAI Aleph Alpha Research\\
{\tt\small janhendrik.metzen@aleph-alpha-ip.ai}
\And
Matthias Hein\\
University of Tübingen\\
{\tt\small matthias.hein@uni-tuebingen.de}
\AND
Lukas Schott\\
Bosch Center for Artificial Intelligence\\
{\tt\small lukas.schott@de.bosch.com}
\vspace{-.3cm}
}

\usepackage{amsmath}
\usepackage{amssymb}
\usepackage{mathtools}
\usepackage{amsthm}
\usepackage{algorithm}

\usepackage{algpseudocode}
\usepackage[textsize=tiny]{todonotes}
\renewcommand{\algorithmicrequire}{\textbf{Input:}}
\renewcommand{\algorithmicensure}{\textbf{Output:}}

\newcommand{\mhn}[1]{{\color{black}#1}}

\newcommand{\lsn}[1]{{\color{black}#1}}

\begin{document}

\maketitle

\begin{abstract} 
\lsn{While unlabeled image data is often plentiful, the costs of high-quality labels pose an important practical challenge: Which images should one select for labeling to use the annotation budget for a particular target task most effectively?}
\lsn{To address this problem, we focus on \textit{single-pass data selection}, which refers to the process of selecting all data to be annotated at once before training a downstream model. }
Prior methods for single-pass data selection rely on image-level representations and  fail to \lsn{reliably} outperform random selection for object detection and segmentation.
We propose Object-Focused Data Selection (OFDS) which leverages object-level \mhn{features} from foundation models 
\mhn{and ensures} semantic coverage of all target classes. In extensive experiments across tasks and target domains, OFDS consistently outperforms random selection and all baselines. The best results for constrained annotation budgets are obtained by combining human labels from OFDS with autolabels from foundation models. Moreover, using OFDS to \lsn{select} the initial labeled set for active learning yields consistent improvements.
\end{abstract}

\section{Introduction}
\label{sec:intro}

\begin{figure*}
    \centering
\includegraphics[width=0.99\linewidth]{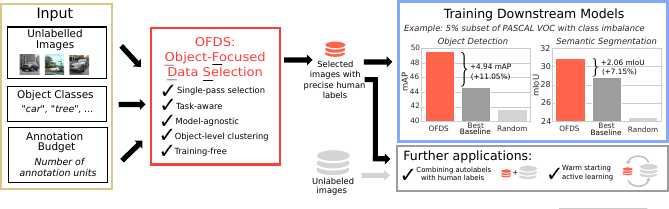}
    \caption{\textbf{Setup for Object-Focused Data Selection.} OFDS selects a subset of images to be labeled from an unlabeled dataset given a set of object classes and an annotation budget for object-level annotations. Training downstream models with the data selected by OFDS improves over all data selection baselines and random selection (results are for an annotation budget of 5\% of all annotation units \mhn{for} the PASCAL VOC dataset with class imbalance).
    The detailed experimental setup is discussed in Section \ref{sec:experiments}.}
    \label{fig:illustration}
    \vspace{-0.5cm}
\end{figure*}

The performance of machine learning systems critically depends on the availability and quality of training data \cite{zha2023data, quality_vs_quantity}. 
The main cost factor in the creation of a dataset is usually not the acquisition of unlabeled images, 
but the labeling. \lsn{This is particularly true for object detection and segmentation, where instance- and pixel-level annotations are required. Labeling a single image for these tasks} can take from seconds for simple cases to over 90 minutes for complex scenes \cite{annotation_duration}. This cost motivates \textbf{the problem of data selection:} given a pool of unlabeled data, a target task, and a fixed annotation budget, \lsn{the goal is to} select a subset of images for labeling that maximizes the downstream performance of models trained on it. We consider object detection and segmentation as our target tasks, for which the set of classes is known via their names. This setup is \mhn{illustrated} in Figure \ref{fig:illustration}.

A standard approach that \lsn{combines} data selection \lsn{simultaneously} with training a downstream model is active learning. In contrast, our goal is to perform model-agnostic data selection, where data is selected before training the downstream model. 
\mhn{The iterative setup of active learning requires multiple model training cycles}, which is up to several hundred times slower than data selection in a single-pass \cite{freesel}. Moreover, active learning methods require an initial dataset with labels, which is typically chosen at random (“cold start”). Using data selection methods to guide this initial choice (“warm start”) has been shown to improve downstream performance in active learning \cite{hacon}. \lsn{This highlights the complementary nature of data selection and active learning.}

Existing methods for model-agnostic and single-pass data selection rely on image-level representations and constrain the selection by the number of images to be annotated \cite{elfs,updp,freesel}. For object detection and segmentation, where the target task entails predictions on instance- or pixel-level, we demonstrate that these methods fail to consistently outperform random selection. Furthermore, the annotation budget is \lsn{usually} not determined per image but per annotation unit (bounding box or mask). Consequently, the annotation cost of an image varies depending on the target classes. Therefore, selecting a subset that meets a given annotation budget requires knowledge of the target classes, which is not accounted for in existing methods.

In this paper, we introduce Object-Focused Data Selection (OFDS), a data selection method which aims to achieve a semantic covering of all target classes at the level of objects. \lsn{While OFDS is model-agnostic and can be used for training or fine-tuning larger models, our main goal is the selection of datasets for training compact downstream models for resource-constrained environments such as autonomous driving or mobile robots, where foundation models cannot be deployed.}
We summarize our main contributions as follows:


\setlist{nosep}
\begin{enumerate}[leftmargin=*,noitemsep]
    \item \textbf{Object-Focused Data Selection (OFDS)} leverages object-level representation from foundation models and adaptive clustering to guide the data selection. Unlike prior work \cite{freesel,updp,hacon,elfs}, OFDS takes the semantic similarity on the level of objects rather than entire images into account. Furthermore, it accounts for a labeling budget given by a number of annotation units.
    \item \textbf{Empirical Demonstration of the Effectiveness of OFDS.} We compare OFDS to 8 baselines for single-pass data selection across three tasks and four datasets. OFDS consistently improves performance
    , in particular in scenarios with imbalanced class distributions.
    \item \textbf{Combining Autolabeling with OFDS.}
    Existing works on data selection ignore the fact that foundation models can be used to generate machine-generated \textit{autolabels} that could be potentially replace human labels. We first \lsn{reiterate} that autolabels can rival human annotations only on simple datasets but fall short as task complexity increases. Subsequently, we propose a holistic training framework that integrates autolabels with OFDS and yields the best performance even for very small annotation budgets.
    \item \textbf{Warm starting active learning.} 
    Active learning methods require a small initial labeled set which is typically chosen randomly (“cold start”). We show that using OFDS for this initial selection consistently improves performance over random selection and a recent baseline \cite{hacon}. 
    
\end{enumerate}

\section{Related Work}\label{sec:related_work}
\textbf{Single-Pass Data Selection} refers to the process of selecting the entire subset of data to be annotated at once before training a downstream model. Recent \lsn{methods} \cite{freesel,updp,hacon,elfs} use a single or fixed number of representations per image and assume that the annotation cost is \lsn{solely} based on the number of images without considering the target classes. 
However, for tasks such as object detection and segmentation, the cost of annotating an image depends on the number of objects to be labeled, which in turn depends on the target classes. Therefore, the target classes are practically always required in order to align the data selection with a given annotation budget defined at the level of individual annotation units (bounding boxes or masks). 
\\ 
\textbf{Active Learning.} Active learning 
aims to select the most informative instances from an unlabeled pool given the current state of a model which is being trained \cite{al_survey}. 
There are several key distinctions between single-pass data selection and active learning. First, the selected data in active learning is specific to the trained model and the selection process is done sequentially, which can further increase the overall cost for the annotation process (see App. \ref{sec:app_cost}). In contrast, OFDS is independent of the downstream model being trained, and the selection is performed "passively" before training. Second, most active learning methods for object-level prediction tasks are specific to single tasks such as object detection \cite{ppal} \textit{or} semantic segmentation \cite{al_segmentation,active_semseg, active_label_correction} while OFDS \lsn{is applicable to both}. Third, active learning methods typically assume the presence of an initial labeled dataset (``cold start problem'') before selecting additional data points to be labeled \cite{3dal,3dal2,hacon}, whereas OFDS works without any labeled data. The cold start problem has been addressed specifically for 3D semantic segmentation by 
\cite{3dal,3dal2}. 
To the best of our knowledge, HaCON \cite{hacon} is the only task-agnostic method for the cold start problem that can be used comparably to OFDS.\\  
\textbf{Coreset Selection.} The goal of coreset selection is to select a subset of a large dataset to approximate the entire dataset. 
\lsn{The main difference to single-pass data selection is that typical coreset methods} \cite{coreset_SGD,deepcore,coreset_survey} require having the full dataset labeled or a model trained on a labeled subset. \lsn{Therefore, coreset selection does not address the problem of constrained annotation budgets.} \\
\textbf{Data and Coreset Selection Beyond Image Classification. } Most approaches for data selection have been evaluated on image classification. However, the labeling costs for object-level prediction tasks are higher, which motivates specific approaches for these tasks. To the best of our knowledge, \cite{reidentification}, \cite{metric_selection} and \cite{coreset_object_selection} are the only works considering coreset selection for tasks beyond image classification. However, they require fully labeled datasets for their selection. USL \cite{usl} considers the combination of selected data with annotations and unlabeled data for semi-supervised learning. Similarly, ReCo \cite{reco} considers the selection of reference images from an unlabeled dataset but specifically targets co-segmentation. Li et al. \cite{updp} and Xie et al. \cite{freesel} include evaluations of their methods on dense prediction tasks but perform the selection using image-level representations which we observe to be inferior in settings with class imbalance.\\ 
\textbf{Pre-Training with Autolabels.} Training small models for downstream tasks with autolabels from a foundation model can be viewed as training under weak or noisy supervision \cite{noisy_supervision,active_label_correction}. This has been explored for specific domains such as segmentation in remote sensing \cite{remote_sensing} or tasks such as local feature learning \cite{local_feature}. While these works focus purely on training with autolabels on entire datasets, the data selection problem considers \textit{which} images to label.

\vspace{-.15cm}
\section{Method: Object-Focused Data Selection}\label{sec:formatting}
In this section, we introduce Object-Focused Data Selection (OFDS), a data selection method for object-level prediction tasks like object detection or segmentation. By \textit{data selection}, we refer to the process of choosing a subset of images $\mathcal{S}$ to be labeled, from a large unlabeled pool ${\textbf{I}_1,\dots, \textbf{I}_N}$, given a fixed annotation budget $B$ in annotation units and a set of $M$ target classes ${C_1, ..., C_M}$. The target classes are given by their class names. We begin with a high-level overview of OFDS before describing each step in detail.  
\vspace{-.15cm}
\subsection{Overview}
\vspace{-.1cm}

\begin{figure}[ht]
    \centering
    \includegraphics[width=.95\linewidth]{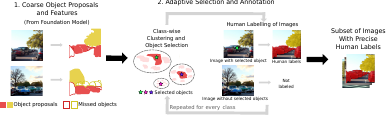}
    \caption{\textbf{Two Stage Selection Process of OFDS.} In the first step, coarse object-level features are extracted from foundation models. These features are grouped per class and clustered. The number of clusters is chosen adaptively based on the remaining budget and already labeled objects. For every cluster, we select one object to ensure a semantic covering of the target classes. Only images containing selected objects are labeled by human annotators.}
    \label{fig:illustration_clustering}
    \vspace{-.2cm}
\end{figure}

The objective of OFDS is to select a labeled subset that achieves semantic coverage across all target classes. The overall procedure consists of two stages, as illustrated in Figure~\ref{fig:illustration_clustering}.

\begin{enumerate}[leftmargin=*,noitemsep]
    \item \textbf{\lsn{Coarse} Object Proposals and Feature Extraction:} We use foundation models to detect objects in the unlabeled data pool and
    extract representations for them. These proposals are coarse and not as precise and complete as the human labels obtained in step 2. We assign the object features to classes based on labels predicted by the foundation model.
    \item \textbf{Adaptive Class-level Clustering of Object Features:} For every target class, we cluster the object features to group semantically similar objects. The number of clusters is chosen adaptively given the annotation budget and already labeled objects of that class. To construct a covering of the semantic space of every target class, we select one object from every cluster. For images containing selected objects, all objects from the target categories are considered for human labeling. 
\end{enumerate}

\subsection{Detailed Steps}
\vspace{-.1cm}
\textbf{Step 1: Object Proposals and Feature Extraction.}\label{sec:feature_extraction}
In the first step, a foundation model - referred to as the \textit{object proposer} - is used to detect which objects are present in an image. For this purpose, we use Grounding DINO \cite{groundingDINO,groundingSAM} as pre-trained open-world object detector. Given an image and a set of class names, it returns a set of object detections consisting of bounding boxes, labels and confidences. The bounding boxes are used as queries to SAM 2 \cite{sam2} for generating object features. These models were chosen as they are state-of-the-art open-world detection and segmentation models performing well across various benchmarks \cite{groundingDINO,groundingSAM,sam2}. A critical aspect of using the object proposer is calibrating the confidence threshold that determines which object proposals to consider. For data selection, it is important to obtain reliable, high-quality predictions for class objects instead of noisy predictions for all potential objects. Therefore, the confidence threshold is set on a reference dataset such that the false positive rate of object proposals is 5$\%$. This ensures that our employed object proposals are highly accurate \mhn{at the cost of missing some objects}. 
\mhn{Depending on the task, the quality of the object proposals varies and is not as accurate as human labels (see Section \ref{sec:finetuning_results}). However, as shown in the experiments even for tasks where their quality is low, OFDS works at least as good as random selection.} 
Based on the object proposals, we construct the object features to provide semantic information for clustering similar objects. We leverage object pointers from the SAM2 memory bank as features which contain high-level information of objects and are stored as 256-dimensional vectors \cite{sam2}. \lsn{For image $\textbf{I}_i$, we denote an object feature for class $l_j$ by $\textbf{O}_{l_j}^{\textbf{I}_i}$.} 

\textbf{Step 2: Adaptive Selection and Annotation.} 
\mhn{The clustering of object features and selection of images to be labeled is done for each target class separately. We order the classes according to the number of object proposals and then begin with the class with the least proposals. In this way, we focus on covering potentially rare classes first. }\\
Given the set of object features for class $C_l$ denoted as $\mathcal{D}_l = \{\textbf{O}_{l_j}^{\textbf{I}_i}| l_j = C_l\}$, we apply k-means clustering independently for each class. The goal is to find a partitioning $\mathcal{D}_l = \{D_l^1, ..., D_l^k\}$ which minimizes the within-cluster sum of squares
\begin{equation} 
    \mathcal{V}_l (\{D_l^1, ..., D_l^k\})= \sum_{i=1}^k \sum_{\textbf{O} \in D_l^i} \| \textbf{O} - \boldsymbol{\mu}_j \|^2 \;\; \textrm{where} \;\; \boldsymbol{\mu}_j = \frac{1}{|D_l^i| } \sum_{\textbf{O} \in D_l^i} \textbf{O} 
\end{equation}
Minimizing $\mathcal{V}_l$ ensures that the clusters are semantically coherent and cover the object feature space of class $C_l$. An illustration with example clusters is given in App. \ref{sec:app_clustering_duplicates}.  \\
The number $k$ of clusters for $k$-means is chosen adaptively for every class by taking the remaining annotation budget and already annotated images into account. Given class $N_{C_l}$, we evenly distribute the annotation budget between the remaining classes. This results in a budget of
\begin{equation} \label{eq:cluster_num}
    N_{C_l}=\frac{B-N(\mathcal{S})}{(M-l+1)N_O}
\end{equation}
images to be annotated. Here, $B$ is the total annotation budget, $N(\mathcal{S})$ is the number of already annotated units, $M$ is the total number of target classes, and $N_O$ is an estimate of the average number of target objects per image. 
Since later classes may have clusters that contain object features from previously annotated images, we perform a linear search over \mhn{the number of clusters} $k$ until there are $N_{C_l}$ clusters without object proposals from already annotated images. From these clusters, we select the feature closest to the cluster centroid $\boldsymbol{\mu}_j$ \mhn{and choose the image containing the corresponding object to be labeled}. This adaptive mechanism ensures that the selected set of objects yields a semantic covering over the object feature space and avoids over- or under-representation of individual clusters. In particular, it prevents the explicit inclusion of exact or near-duplicates.
On images with selected objects, objects from all target classes are labeled by the human annotators. Although this may initially seem contrary to the object-focused approach, exhaustive labeling yields information \mhn{about which part of the image belongs to the background}. This background information is required by most common training frameworks for object-level prediction tasks, either as a separate class \cite{psp} or to ensure correct negative samples \cite{missing_labels}. In particular, the negative samples from the background are class-agnostic, which provide higher information value than class-specific negative samples that could show objects from other target classes. While there exist potential solutions for training models for object-level downstream tasks with partial labeling \cite{partial_labelling,partial_labelling_jmlr}, we aim to ensure compatibility with standard setups.

The complete steps for OFDS are summarized in Algorithm \ref{alg:ofds}.

\begin{algorithm}[h]
\small
\caption{OFDS: Object-Focused Dataset Selection}\label{alg:cap}
\begin{algorithmic}[1]
\Require Set of unlabeled images: $\{\textbf{I}_1,\dots, \textbf{I}_N\}$, Annotation budget by number of units: $B$, Estimated number of annotation units per image: $N_O$, Classes to label sorted by ascending number of object proposals per class: $\{C_1, ..., C_M\}$
\Ensure Subset of images $\mathcal{S}$ selected for labeling
\State Generate a set of object features and corresponding labels $\{(\textbf{O}_j^{\textbf{I}_i},l_j^{\textbf{I}_i})\}_{j=1}^{K_i}$ for every image $\textbf{I}_i$ using the object proposer
\State Initialize the subset $\mathcal{S}=\{\}$
\For{$l \in \{1,\dots,M\}$} 
    \State Select the object features predicted as class $C_l$ by the object proposer: $\mathcal{D}_l = \{\textbf{O}_{l_j}^{\textbf{I}_i}| l_j = C_l\}$
    \State Determine the number of images to add $N_{C_l}$ for the current class from Eq. \ref{eq:cluster_num}
    \State Perform $k$-means clustering on $\mathcal{D}_l$ with adaptive $k$ to feature $N_{C_l}$ clusters without images from $\mathcal{S}$
    \State For every cluster without images from $\mathcal{S}$, select the object $\textbf{O}_{l_j}^{\textbf{I}_i^*}$ which is closest to the cluster mean
    \State Annotate the images with the selected objects and update $\mathcal{S}$: $\mathcal{S}=\mathcal{S} \cup \{\textbf{I}_i^*\}$
\EndFor
\State \textbf{Return} $\mathcal{S}$
\end{algorithmic}
\label{alg:ofds}
\end{algorithm}

\section{Experiments}\label{sec:experiments}
We first outline the setup used to conduct our experiments and subsequently discuss the results in three parts. First, we conduct an extensive comparison of OFDS against existing data selection baselines across three tasks and four datasets. Second, we discuss the model performance on downstream tasks when training purely with autolabels and highlight the advantage of pre-training on the full dataset with autolabels and fine-tuning with human annotations on selected subsets. Finally, we demonstrate that OFDS can be used to improve active learning by selecting the initial data to be labeled.

\subsection{Experimental Setup}
\textbf{Tasks and Datasets} As downstream tasks, we consider object detection, semantic and instance segmentation. In the following section, we focus on object detection and semantic segmentation, while the results for instance segmentation are contained in Appendix \ref{sec:app_instance}. For object detection, we use the joint training set from the PASCAL VOC \cite{pascal} 2007 and 2012 splits, evaluated on the validation set from 2012 as well as the Cityscapes \cite{cityscapes} dataset with the classes featuring instance-level annotations. Similarly, for semantic segmentation we consider the PASCAL VOC 2012 dataset, Cityscapes and LoveDA \cite{loveda} dataset. The PASCAL VOC datasets feature a manually balanced object distribution as shown in App. \ref{sec:app_subsets}. This can be attributed to the fact that the dataset has already been selected and labeled by humans. However, unlabeled real-world datasets typically follow class imbalanced distributions \cite{imbalance_survey}. Thus, we construct two additional settings with rare classes. We reduce the number of annotations units for the six smallest classes in the PASCAL VOC dataset by 99$\%$, 95$\%$, 85$\%$, 80$\%$, 75$\%$ and 50$\%$. We refer to this setting as PASCAL VOC with class imbalance (class imbal.). The original and class imbalanced object distributions are shown in App. \ref{sec:app_subsets}. The class distribution of Cityscapes naturally contains rare classes such that we directly use the full dataset. The LoveDA dataset contains remote sensing images which we use to test the performance of OFDS in a specialized domain for which the foundation models used for the object proposals in OFDS might not have seen much training data.

\textbf{Models and Training Setup} For our main experiments on object detection we use a Faster RCNN \cite{faster_rcnn} with ResNet-18 backbone \cite{resnet} and for semantic segmentation a Segmenter \cite{segmenter} with ViT-T backbone \cite{vit}. Ablations with a Deformable DETR \cite{deformable} for object detection and a PSP Net \cite{psp} for semantic segmentation \lsn{further support our findings and} can be found in App. \ref{sec:app_model_ablations}. The backbones were pre-trained on ImageNet. In Section \ref{sec:data_selection_results}, we train the decoder parts from scratch with the obtained human labels to evaluate the influence of the data selection. In Section \ref{sec:finetuning_results}, we initialize the model with the checkpoint obtained after pre-training with autolabels to improve the downstream performance. For every setting consisting of dataset and task we train for the same number of steps on all subsets. We use augmentations consistent with Xie et al. \cite{freesel}. The complete augmentations and hyperparameters can be found in Appendix \ref{sec:app_hyperparameter}. For the experiments for warm starting active learning, we follow experimental setups from \cite{ppal,al_segmentation}. Details can be found in Section \ref{sec:active_learning}. As object proposer we use Grounding DINO-T and Grounding SAM2-T for our main results. An ablation of the object proposer is contained in Appendix \ref{sec:ablation_object_proposer} and the details on the calibration can be found in Appendix \ref{sec:app_cal}. \lsn{In App. \ref{sec:grounding_dino}, we demonstrate that OFDS also yields improvements when fine-tuning Grounding DINO-T on selected subsets.}

\textbf{Baselines}
To provide a fair comparison of our method to existing work, we have compiled a list of 8 existing methods for single-pass data selection as baselines. Table \ref{tab:method_overview} summarizes the similarities and differences between OFDS and the baselines. To the best of our knowledge, OFDS is the first dedicated method for data selection to meet all guiding criteria for single-pass data selection for object-level prediction tasks. Apart from text-to-image retrieval, \mhn{no other method} utilizes class names for the data selection and cannot be easily modified to do so as theys only work in the image feature space. 
Despite Prototypes and ELFS not using the semantic class names for the selection itself, they require the number of classes as input. Wrongly estimating the number of classes for the downstream task when using image-level clustering has been shown to substantially deteriorate performance \cite{beyond,hacon}. Thus, ELFS and prototypes can be viewed as task-specific, even though in a weaker sense. The closest baseline to OFDS is text-to-image retrieval, for which we use a state-of-the-art SigLip 2 ViT-B/16 model \cite{siglip2}. A more detailed summary of all baselines \mhn{and exact results for all methods in the form of tables} can be found in App. \ref{sec:baselines_summary} and \ref{sec:numerical_results} .

\begin{table}[bp]
\vspace{-.2cm}
\centering
\resizebox{.9\columnwidth}{!}{%
\begin{tabular}{|l|c c c c c |}
\hline
 Method & Single- & Model- & Training- & Uses the Number & Classname- \\ 
  & Pass & Agnostic & Free & of Classes & Aware \\\hline
 Random & \cymark & \cymark & \cymark & \cxmark & \cxmark  \\
 FreeSel \cite{freesel} & \cymark & \cymark & \cymark & \cxmark & \cxmark  \\ 
 K-Centers \cite{active_learning_coreset} & \cymark & \cymark & \cymark & \cxmark & \cxmark  \\ 
 UP-DP \cite{updp} & \cymark & \cymark & \cxmark & \cxmark & \cxmark  \\ 
 HaCON \cite{hacon} & \cymark & \cymark & \cxmark & \cxmark & \cxmark  \\ 
 ELFS \cite{elfs} & \cymark & \cymark & \cxmark & \cymark & \cxmark  \\  
 Prototypes \cite{active_learning_coreset} & \cymark & \cymark & \cymark & \cymark & \cxmark  \\ 
 Text-to-Image Retrieval & \cymark & \cymark & \cymark & \cymark & \cymark \\ \hline
 \textbf{OFDS} & \cymark & \cymark & \cymark & \cymark & \cymark \\  \hline
\end{tabular}
}
\caption{Comparison of the properties of OFDS with the baselines.}
\label{tab:method_overview}
\vspace{-.2cm}
\end{table}

\begin{figure*}[h!]
 \centering
 \includegraphics[width=.99\columnwidth]{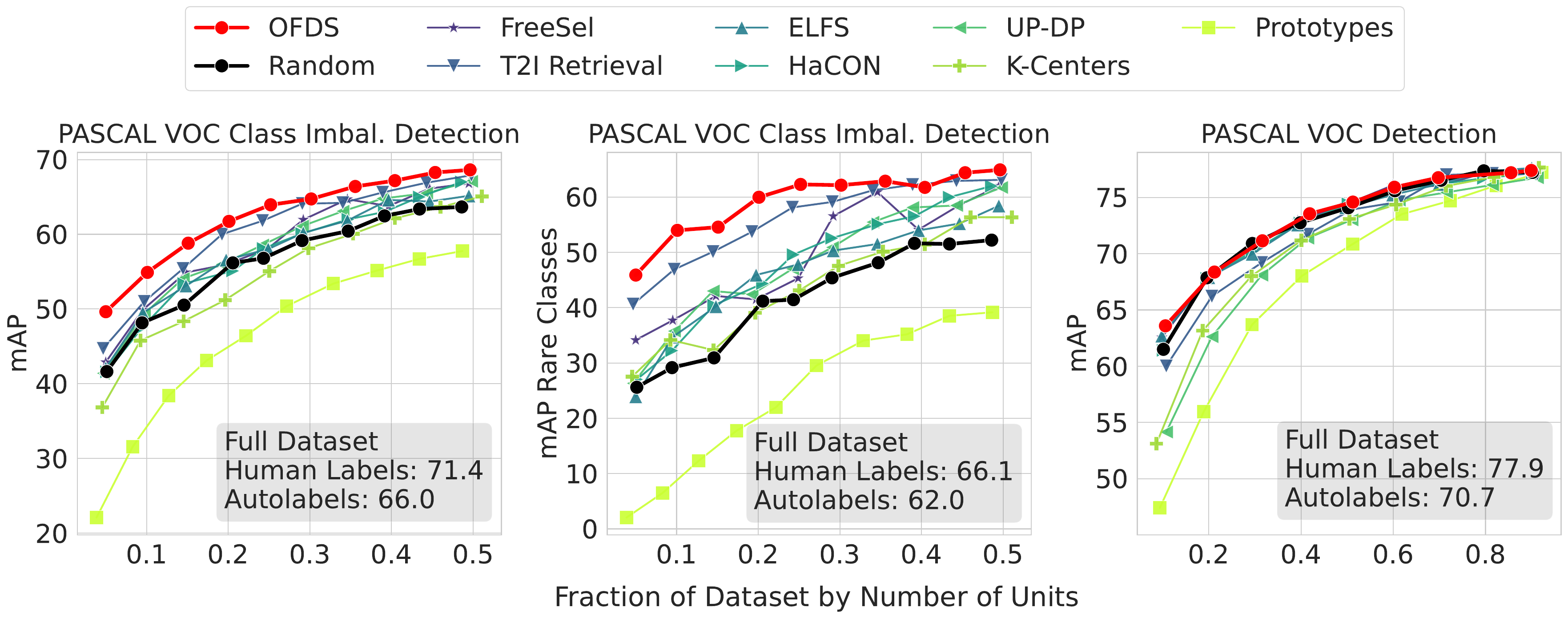}\\
 \includegraphics[width=.99\columnwidth]{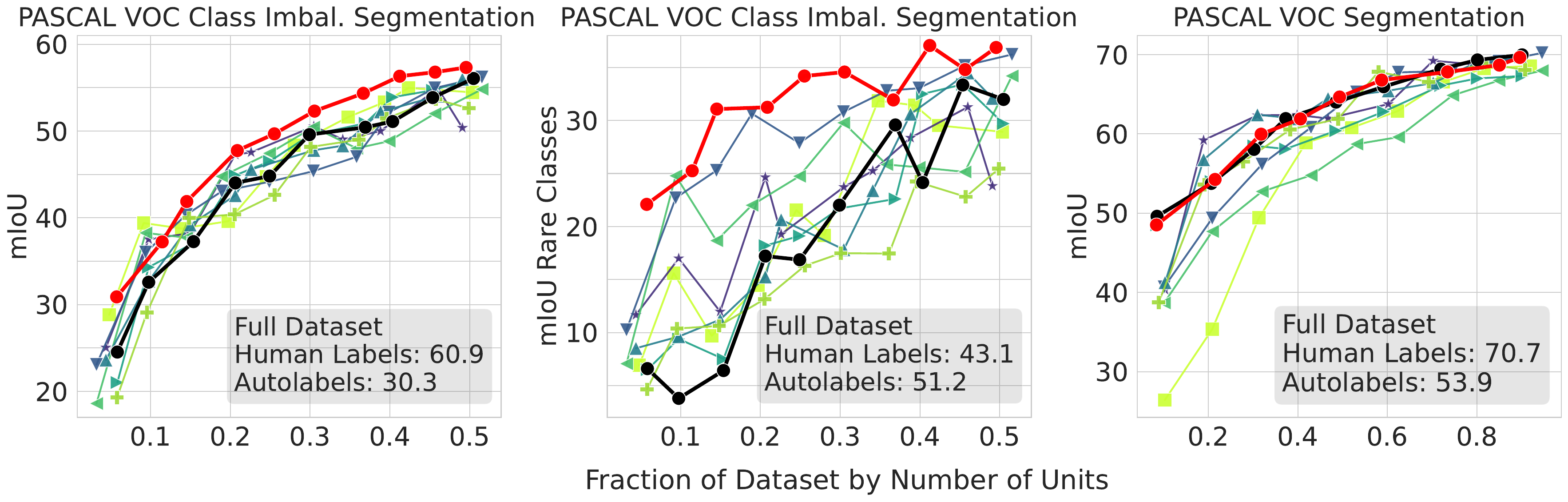}
 \vspace*{-.15cm}
 \caption{\textbf{Data Selection on PASCAL VOC.} \mhn{For object detection we use a FasterRCNN with ResNet-18 backbone and  for semantic segmentation a Segmenter with ViT-T backbone where the decoder part of the models is trained from scratch.} OFDS consistently performs among the best methods on PASCAL VOC and outperforms them on PASCAL VOC with class imbalance (class imbal.). }
 \label{fig:VOC_results}
 \vspace{-.1cm}
\end{figure*}

\begin{figure*}
    \vspace*{-.01cm}
     \centering
    \includegraphics[width=.99\columnwidth]{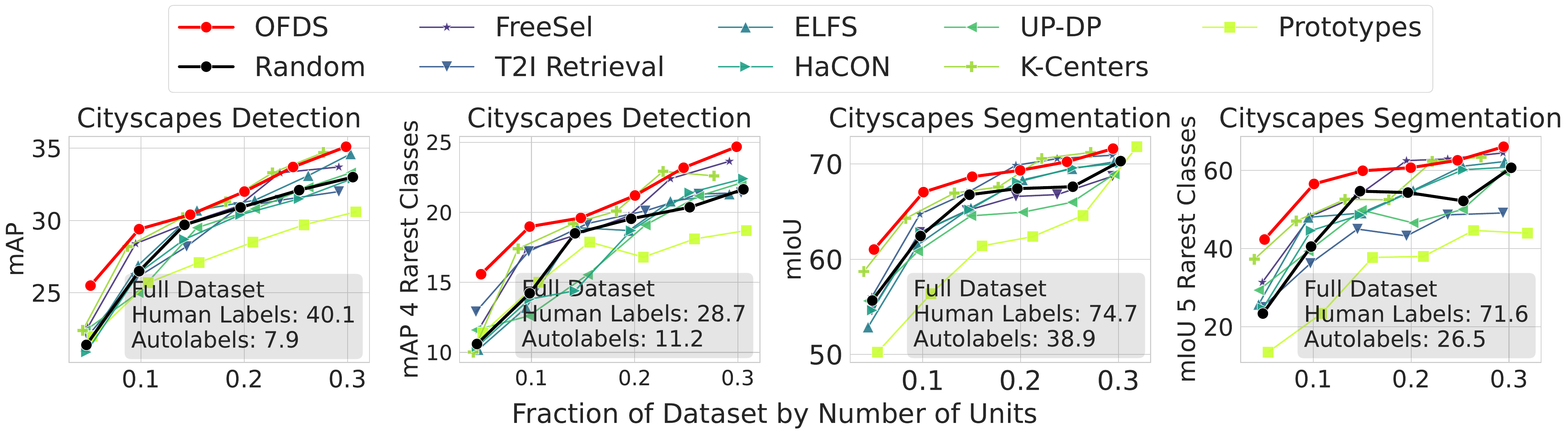}
    \vspace*{-.1cm}
     \caption{\textbf{Data Selection on Cityscapes.} The results are obtained with the same models as for Figure \ref{fig:VOC_results}. Training on \mhn{human annotations of subsets selected with OFDS yields the best results both for object detection and semantic segmentation. Stronger differences can be seen for rare classes.} }
     \label{fig:cityscapes_results}
     \vspace*{-0.4cm}
\end{figure*}

\subsection{Data Selection for Object-Level Prediction Tasks} \label{sec:data_selection_results}
In this section, we compare OFDS to 8 baselines in both class-imbalanced and balanced settings. We note that the imbalanced settings are more representative for real-world data selection. We train the decoder from scratch to investigate the influence of data selection alone and avoid confounding influences from pre-trained network weights.

\textbf{PASCAL VOC with Class Imbalance.} The results for object detection and semantic segmentation on the PASCAL VOC datasets with class imbalance are shown in Figure \ref{fig:VOC_results}  \lsn{left and middle column}. We observe that none of the existing baselines consistently outperforms random selection apart from text-to-image retrieval. In contrast, our method outperforms all baselines, including random selection and text-to-image retrieval for both object detection and semantic segmentation. Notably, the difference to the baselines is largest when assessing the performance on the six rare classes. While there exist post-hoc approaches to adjust training setups to the presence of rare classes \cite{long_tail1,long_tail2,long_tail4,long_tail5}, OFDS targets the class imbalance problem already at the level of data selection. 

\textbf{Full PASCAL VOC.} Figure \ref{fig:VOC_results} \lsn{(right column)} displays the results for the full PASCAL VOC dataset. We observe that random selection serves as a strong baseline, with no other selection method achieving substantial improvements over it. \lsn{This can be attributed to the fact that the class distribution of the full dataset was manually balanced (see App. \ref{sec:app_subsets}, \cite{pascal}), such that random subsets can be expected to achieve good coverage of all target classes.} FreeSel, HaCON, text-to-image retrieval, and OFDS perform on par with random selection, while the prototype-based approach consistently yields the lowest performance. UP-DP and K-Centers underperform compared to random selection. \lsn{A potential reason for this observation is that} these methods rely on image-level representations and were originally developed for image classification while object detection and  segmentation are multi-label tasks.

\textbf{Cityscapes.} The results for object detection and semantic segmentation on Cityscapes are shown in Figure \ref{fig:cityscapes_results}. \mhn{Cityscapes naturally has a more imbalanced class distribution than PASCAL VOC (see App. \ref{sec:app_subsets}).} 
\lsn{We observe that OFDS reliably performs among the best methods.}
\lsn{The performance of OFDS} is especially notable when evaluating on the rarest classes which highlights \lsn{its} ability to select effective instances from rare classes. Text-to-image retrieval, which is the strongest baseline on PASCAL VOC, performs substantially worse on Cityscapes in comparison. This can be attributed to the fact that the Cityscapes dataset contains complex scenes with more objects per image. A more detailed discussion on the disadvantages of text-to-image retrieval in this setting can be found in App. \ref{sec:retrieval}.

\begin{wrapfigure}{r}{0.5\textwidth}
     \centering
     \vspace{-.4cm}
    \includegraphics[width=.49\columnwidth]{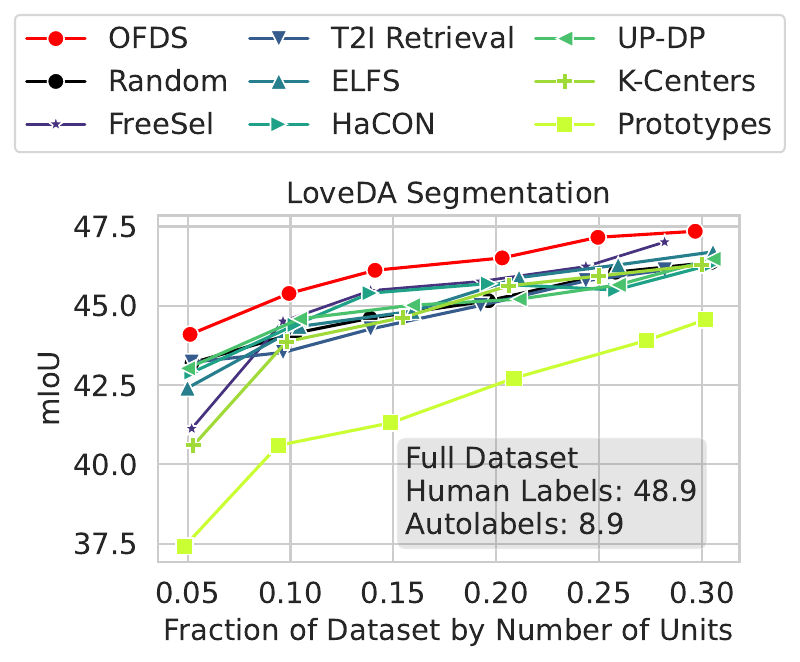}
    \vspace{-.4cm}
    \caption{\textbf{Data Selection on LoveDA.} The setup is the same as the segmentation results from Figure \ref{fig:VOC_results}. Even though the images in LoveDA come from the specialized domain of remote sensing, OFDS consistently outperforms all baselines.}
    \label{fig:loveda}
    \vspace{-.5cm}
\end{wrapfigure}

\textbf{LoveDA.} Figure \ref{fig:loveda} presents the results for semantic segmentation on the LoveDA dataset. \mhn{This dataset consists of images from the domain of remote sensing, which is more challenging, as foundation models generally perform worse in such specialized settings \cite{groundingDINO,groundingSAM}. Despite this, OFDS consistently achieves the best performance, albeit with a smaller margin over the baselines.}

\textbf{Comparison Across All Settings.} Conclusively, we highlight that OFDS outperforms or performs on par with the best baselines across \textit{all} experimental settings and models. OFDS performs best in both balanced and imbalanced class scenarios. This is crucial for practical applications where the presence of class imbalance may not be known in advance. Random selection remains a strong baseline in the class-balanced setting but is substantially surpassed by OFDS in the class imbalanced setting. FreeSel and K-Centers outperform random selection on Cityscapes but fail to reliably perform better than random selection on PASCAL VOC. In contrast, text-to-image retrieval is the strongest baseline on PASCAL VOC, but for the more complex Cityscapes and LoveDA datasets performs substantially worse than OFDS. The selection based on prototypes yields the worst performance across all settings. This underlines the finding by Sorscher et al. \cite{beyond} that dataset selection with the prototypical approach amplifies class imbalance.

\subsection{Can Open-World Foundation Models Eliminate the Need for Pixel-Level Human Annotations?}\label{sec:finetuning_results}

To analyze the model performance on downstream tasks when training with autolabels, we use Grounding DINO-T and Grounding SAM2-T to annotate the full datasets and train downstream models with the autolabels. Unlike for OFDS, the goal when generating autolabels is not to detect or segment objects with high precision but to balance precision and recall. Therefore, we calibrate Grounding DINO by selecting the threshold that yields the highest F1 Score on reference datasets. Details on the two calibration approaches for autolabeling and the difference to OFDS can be found in App. \ref{sec:app_cal}. The performance of downstream models trained using the resulting autolabels can be found in Figures \ref{fig:VOC_results}, \ref{fig:cityscapes_results} and \ref{fig:loveda}. For PASCAL VOC, we observe that the performances of models trained with autolabels are comparable to a 30$\%$ random subset with human labels for object detection and a 20$\%$ subset for semantic segmentation. For Cityscapes and LoveDA, using the full dataset with autolabels performs worse than even a 5$\%$ random subset with human labels. In summary, these findings indicate that for simpler datasets and very limited annotation budgets, autolabels can outperform training on human-annotated subsets. However, with increasing annotation budgets or more complex datasets, human annotations are indispensable. \\
\begin{wrapfigure}{r}{0.55\textwidth}
    \centering
    \vspace{-.1cm}
    \includegraphics[width=.54\columnwidth]{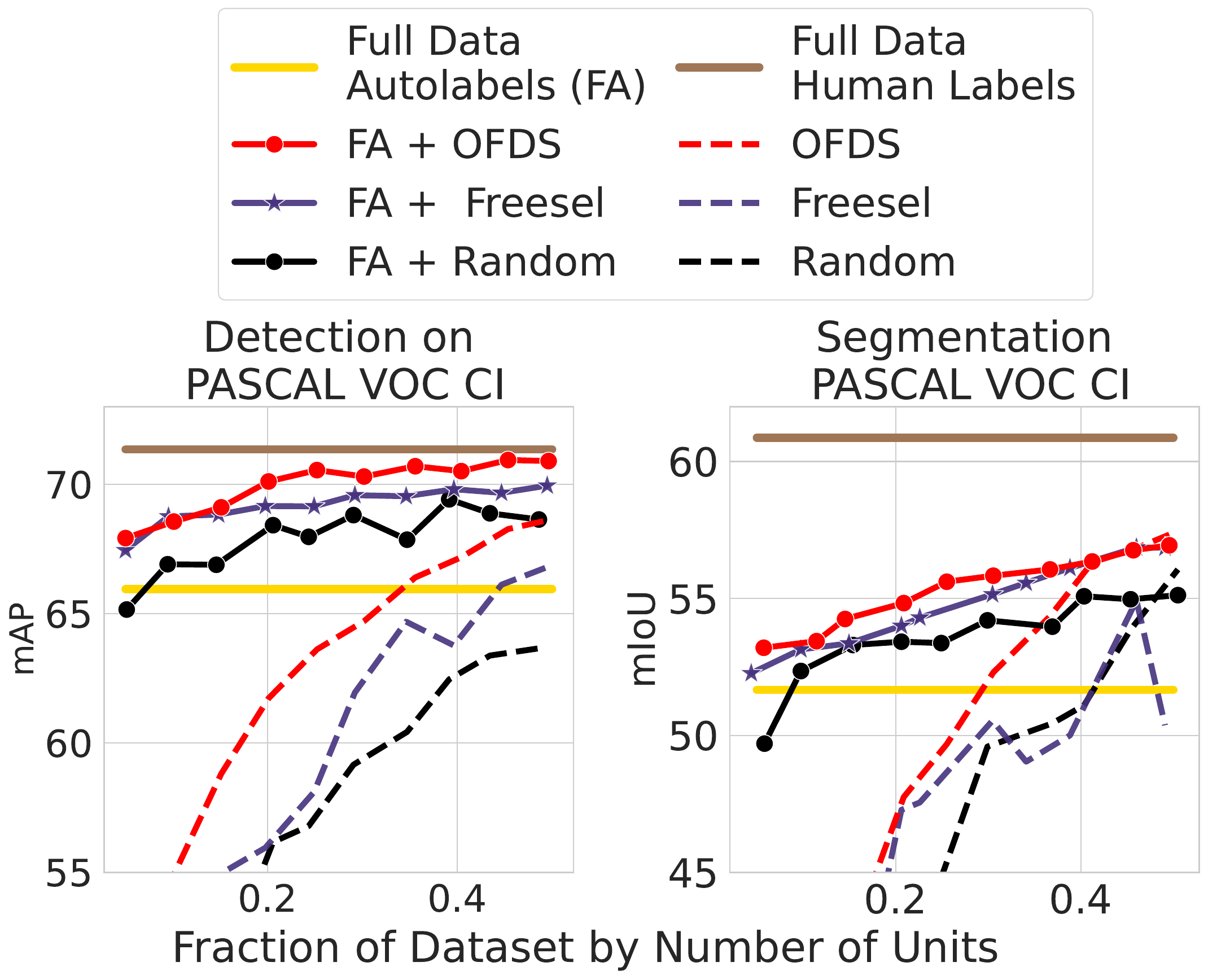}
    \vspace{.05cm}
    \caption{\textbf{Combining Autolabels With Data Selection Improves Downstream Performance under Constrained Annotation Budgets.} Dashed: models trained as in Sec. \ref{sec:data_selection_results}. Solid: fine-tuning the checkpoint pre-trained with autolabels on selected subsets with human labels. Results for Cityscapes in App. Figure \ref{fig:finetuning_cityscapes}.}
    \label{fig:dect_ft}
    \vspace{-.4cm}
\end{wrapfigure}
Nevertheless, the autolabeling process can in any case be carried out at little cost (see App. \ref{sec:app_cost}). Thus, we assess whether the performance under constrained annotation budgets can be improved by incorporating autolabels in addition to human-labeled subsets. Therefore, we first pre-train the models using autolabels on the entire dataset and then fine-tune on human-annotated subsets. In this setup, the purpose of the methods for data selection is to determine the subset used for fine-tuning.
We compare OFDS to random selection and FreeSel as the best and most consistent baseline from Section \ref{sec:data_selection_results}. The complete results on the PASCAL VOC datasets with class imbalance and Cityscapes are shown in Figures \ref{fig:dect_ft} and \ref{fig:finetuning_cityscapes}. We observe that fine-tuning with human-labeled images improves the performance over training purely with autolabels, even for the smallest annotation budget. The improvements on PASCAL VOC are larger in comparison to Cityscapes. This is a result of the stronger performance achieved by training with autolabels on PASCAL VOC. For both datasets, selecting the data for fine-tuning \mhn{with} OFDS leads to the best performance. \mhn{Thus}, data selection can be effectively combined with autolabeling to
\mhn{train models} under limited annotation budgets.



\subsection{Warm Starting Active Learning} 

\begin{wrapfigure}{r}{0.6\textwidth}
     \vspace{-.5cm}
     \centering
    \includegraphics[width=.59\columnwidth]{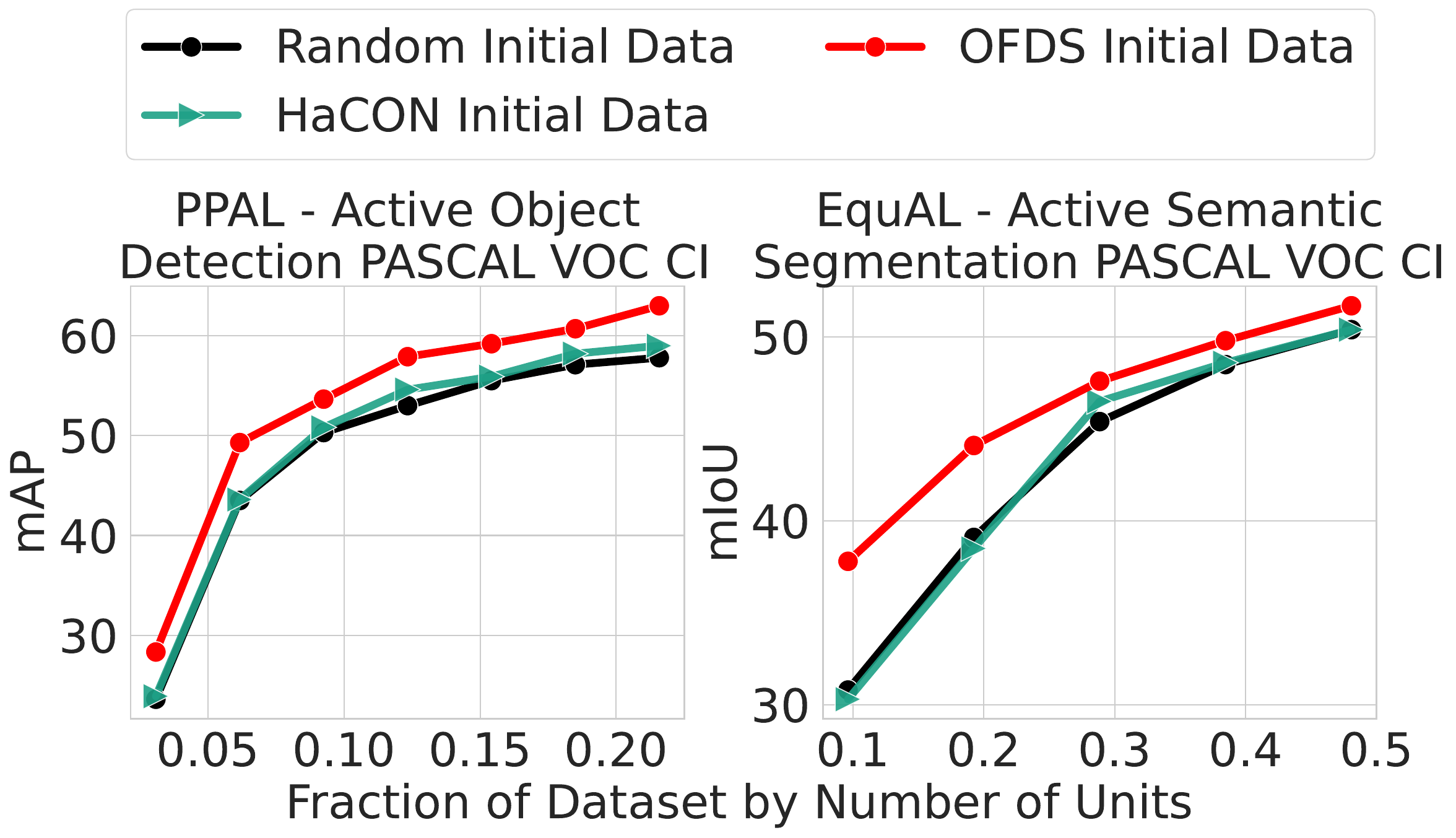}
    \vspace{-.2cm}
    \caption{\textbf{Initial Data Selection Through OFDS Improves Active Learning.} We train a ResNet-50 RetinaNet \cite{retinanet}for detection using PPAL \cite{ppal} and Wide-ResNet38 DeepLabv3+ \cite{deeplab} for semantic segmentation using EquAL. \mhn{The only difference is the selection of the initial data set.} 
    }
    \label{fig:al_main}
    \vspace{-.2cm}
\end{wrapfigure}

As discussed in Section \ref{sec:related_work}, the cold start problem refers to selecting an initial pool of labeled images for active learning which is typically done randomly. In Figure \ref{fig:al_main}, we replace this random initial dataset by
datasets selected through OFDS or HaCON, which is the only baseline specifically devised for the cold start problem. Exemplarily, we use the state-of-the-art \mhn{active learning} frameworks PPAL \cite{ppal} for object detection  and EquAL \cite{equal} for semantic segmentation which performed best in a benchmarking study \cite{al_segmentation} and apply both methods on PASCAL VOC with class imbalance. Further experiments that validate our observation with additional active learning methods and the Cityscapes dataset can be found in Section \ref{sec:active_learning}. We highlight that selecting the initial data through OFDS improves the performance of the model during the entire active learning process and yields better results than a random selection and HaCON. 

\vspace{-.2cm}
\section{Conclusion}
\vspace{-.2cm}
In this work, we address the practically important problem of data selection: choosing which images to annotate from an unlabeled data pool, given a set of target class names and a constrained annotation budget. We propose OFDS, a method that performs data selection at object-level to ensure semantic coverage of all target classes. Through extensive experiments on object detection and segmentation, we demonstrate that our method consistently outperforms existing baselines
, particularly in class-imbalanced settings.  We highlight that data selection remains an important problem since training purely with autolabels from foundation models yields competitive results only on simple datasets with low annotation budgets. Nonetheless, pre-training on auto-labels across the full dataset before fine-tuning on a human-labeled subset selected by OFDS yields the best performance. \mhn{Lastly, using OFDS for the selection of the initial labeled dataset in active learning yields consistent improvements.}

\textbf{Limitations.} OFDS depends on features generated by the object proposer and thereby inherits its biases and limitations, which may reduce its effectiveness in specialized target domains. However, our results on the LoveDA indicate that even in such settings, OFDS can yield improvements over the baselines or random selection albeit with a smaller margin.
Furthermore, achieving a balanced class distribution can be challenging even with OFDS due to class co-occurrences on image level.

\bibliographystyle{plain} 
\bibliography{main_bib}

\newpage

\newpage
\clearpage
\setcounter{page}{1}
\setcounter{section}{0}
\appendix
\onecolumn
\section*{Appendix}
\noindent We start with an overview of the content of the Appendix:
\begin{itemize}
    \item  In Section \ref{sec:app_model_ablations}, we perform ablations with different models for the downstream tasks.
    \item To demonstrate the advantage of using object-level instead of image-level features for the data selection, we compare to an additional image-focused baseline in Section \ref{sec:retrieval}.
    \item We provide further insights into the effect of clustering the object features for OFDS in Section \ref{sec:app_clustering}.
    \item In Section \ref{sec:active_learning}, we provide additional experiments on improving the cold start problem for active learning with OFDS.
    \item In Section \ref{sec:app_class_balance_scores}, we report class balance scores for the selected subsets to validate the effectiveness of OFDS in selecting subsets with improved class balance.
    \item Since the random baseline and FreeSel are based on a probabilistic selection process, we repeat the data selection on PASCAL VOC with class imbalance and assess the extent of the resulting fluctuations in Section \ref{sec:app_repeated}.
    \item The influence of using a stronger object proposer is discussed in Section \ref{sec:ablation_object_proposer}.
    \item In Section \ref{sec:app_subsets}, we provide further details on the class distributions including the rare classes used for evaluation and the subsets used to calibrate the object proposer.
    \item In Section \ref{sec:app_cal}, we discuss the calibration of the foundation model for generating autolabels and object proposals.
    \item Further details on the implementation of OFDS and K-Centers selection are discussed in Section \ref{sec:app_implementation}.
    \item In Section \ref{sec:app_cost}, we provide a discussion on the computational cost of OFDS and generating autolabels.
    \item The complete results for combining autolabels with data selection are shown in Section \ref{sec:app_finetuning}.
    \item The results for instance segmentation on Cityscapes are discussed in Section \ref{sec:app_instance}.
    \item In Section \ref{sec:baselines_summary}, we provide a summary of all baselines.
    \item In Section \ref{sec:grounding_dino}, we fine-tune Grounding Dino on selected subsets.
    \item In Section \ref{sec:app_hyperparameter}, we provide the hyperparameter configurations used for all training runs.
    \item The full numerical results for our main experiments are reported in Section \ref{sec:numerical_results}.
\end{itemize}

\section{Model Ablations for Downstream Tasks} \label{sec:app_model_ablations} \vspace{-.1cm}
\begin{figure}[bp]
     \centering
    \includegraphics[width=.7\columnwidth]{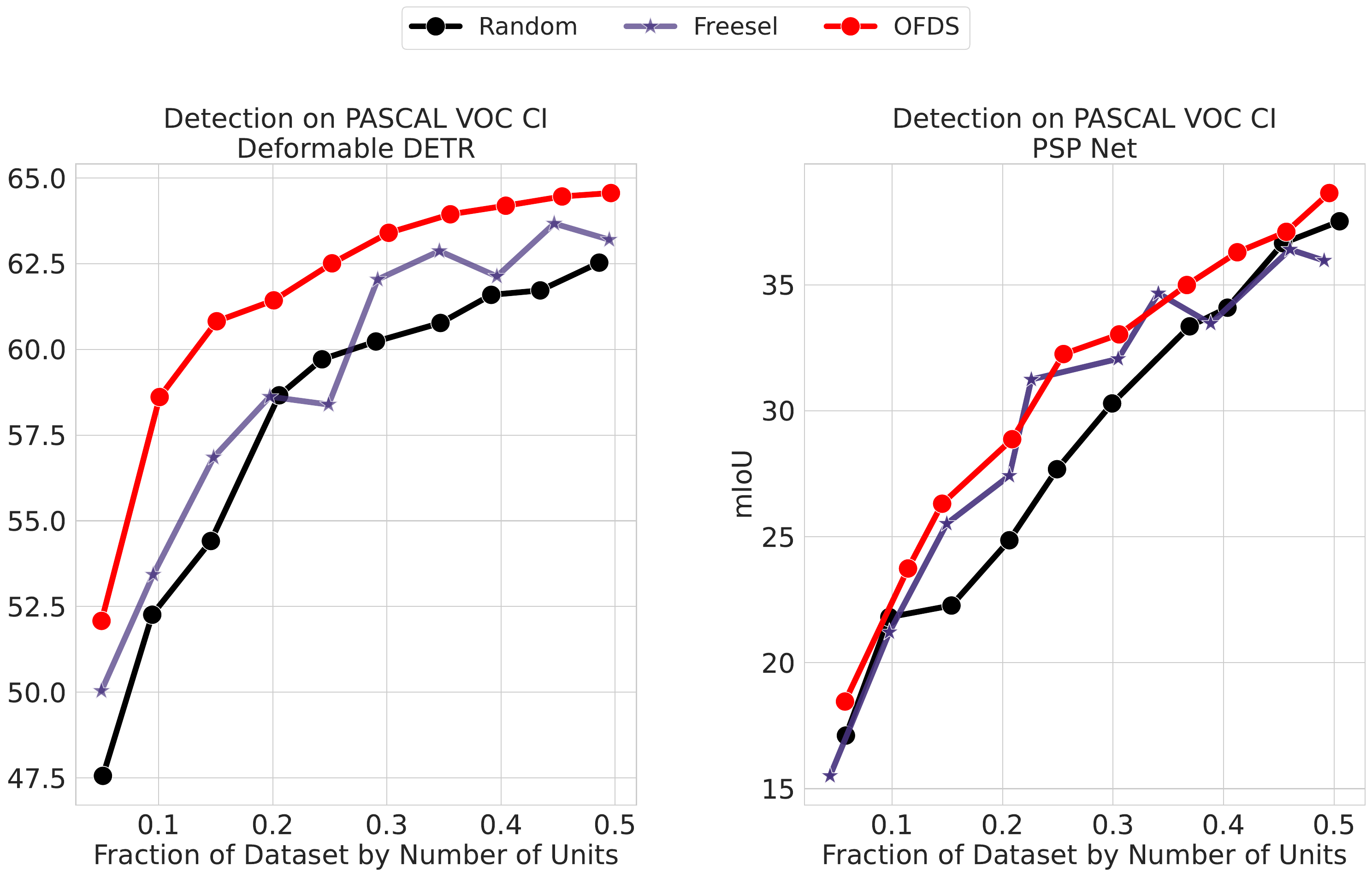}
    \caption{\textbf{Ablations with Different Models for the Downstream Tasks.} OFDS also leads to the best results for training a Deformable DETR for object detection and a PSP Net for semantic segmentation.}
    \label{fig:model_ablations}
\end{figure}
To highlight that the performance advantages of selecting data through OFDS are independent of the model chosen for the downstream tasks, we perform ablations using a Deformable DETR \cite{deformable} for object detection and a PSP Net \cite{psp} for semantic segmentation. Both models are based on ResNet-18 backbones which were pre-trained on ImageNet. As in Section \ref{sec:data_selection_results}, the decoders are trained from scratch to assess only the influence of data selection without any confounding effects from autolabels. The results are shown in Figure \ref{fig:model_ablations}. We observe that selecting the data through OFDS consistently results in the best performances which confirms our previous findings.
\section{Image-Focused vs. Object-Focused Features} \label{sec:retrieval}
\begin{figure*}
     \centering 
    \includegraphics[width=.7\columnwidth]{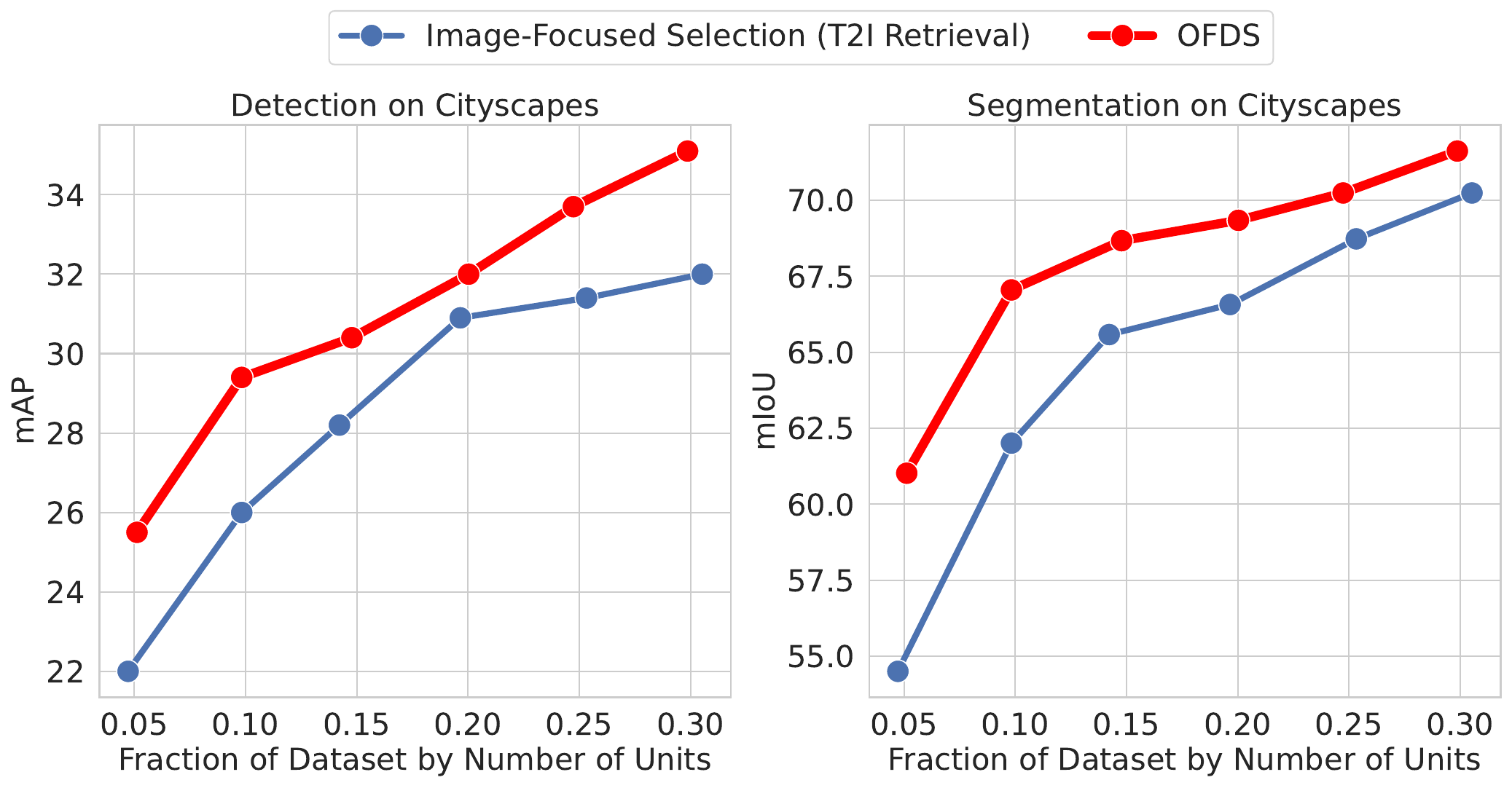} 
    \caption{\textbf{OFDS Outperforms Image-Focused Baseline.} The image-focused baseline is based on CLIP retrieval with an evenly split budget between all classes. The model and training hyperparameters are the same as in Figure \ref{fig:cityscapes_results}. }
    \label{fig:retrieval}
\end{figure*}

\begin{figure}[bp]
    \centering
    \includegraphics[width=.3\columnwidth]{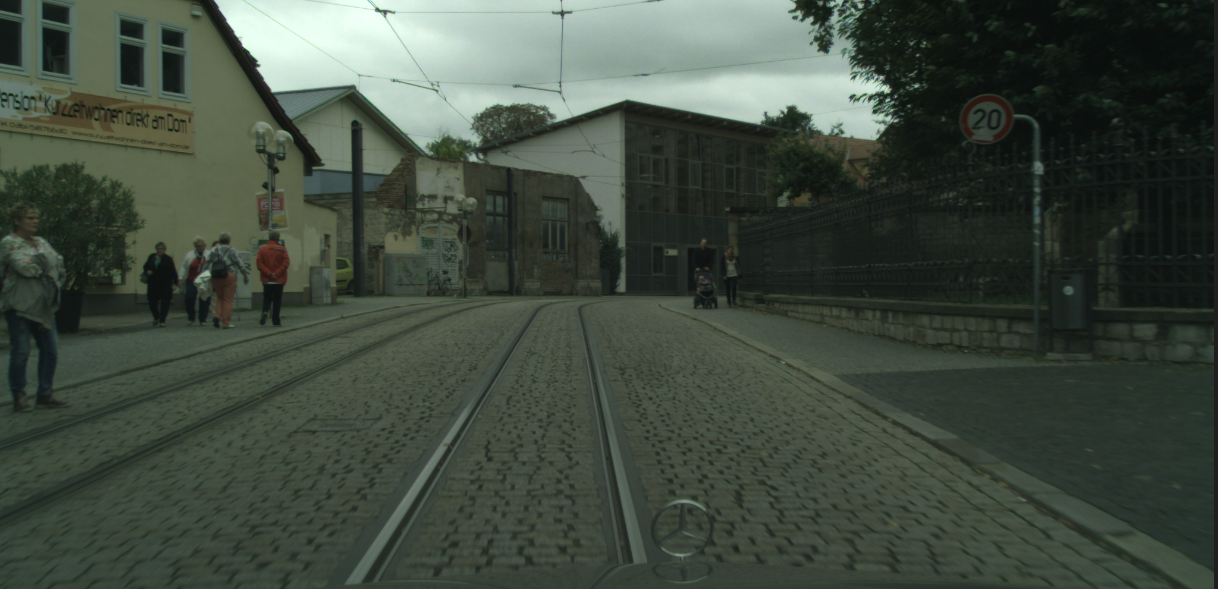}
    \hspace{.4cm}
    \includegraphics[width=.3\columnwidth]{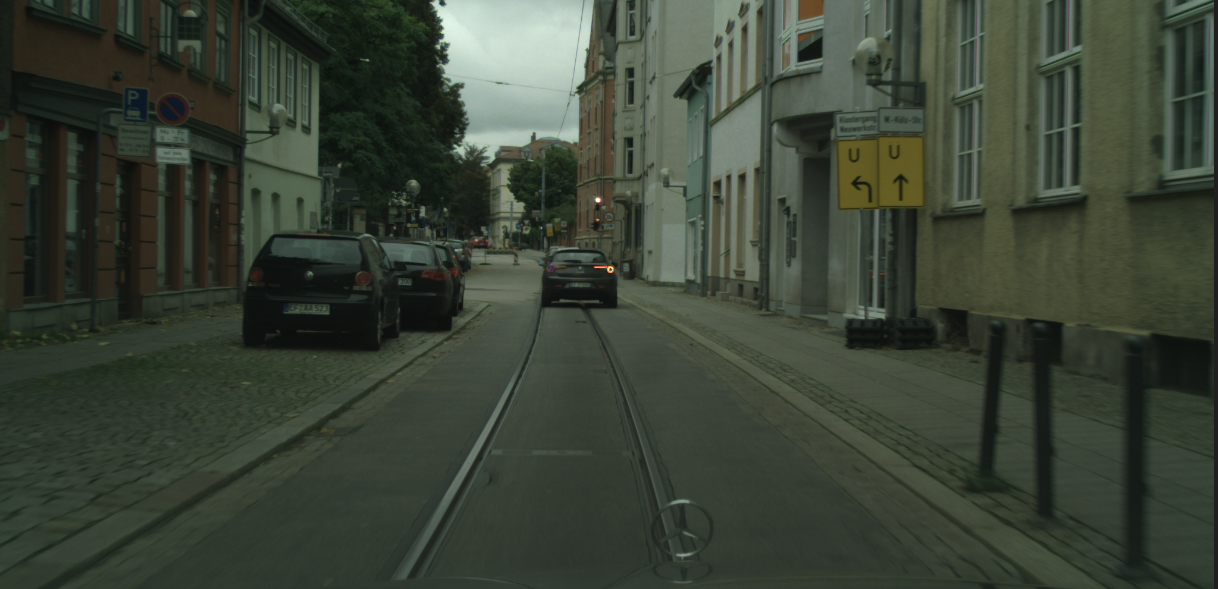}
    \hspace{.4cm}
    \includegraphics[width=.3\columnwidth]{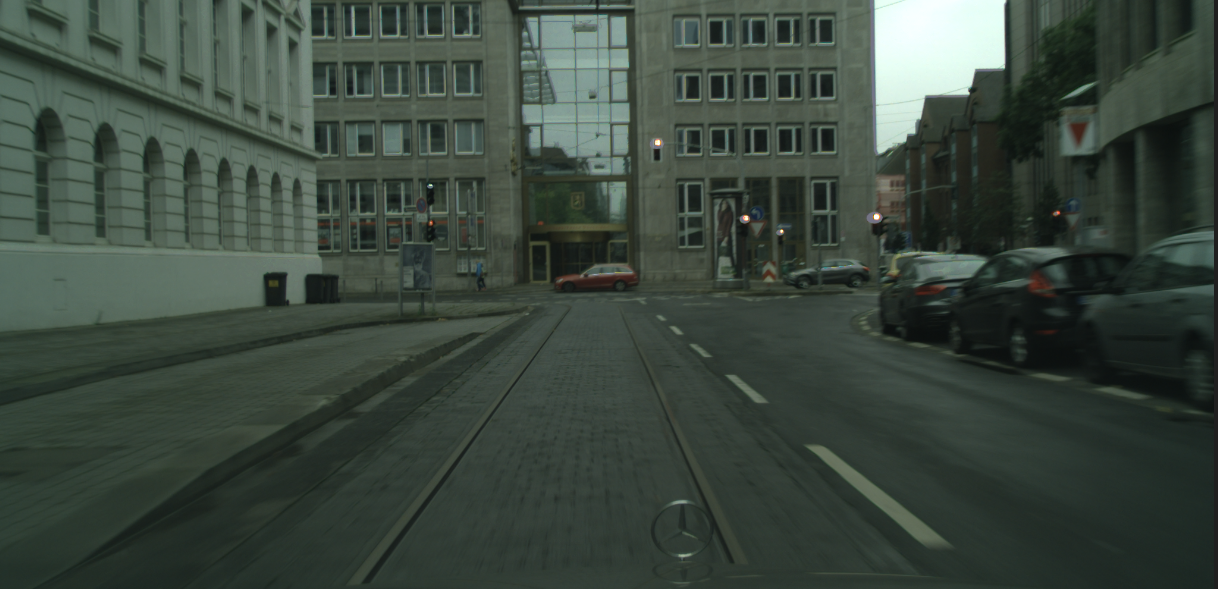}
    \caption{\textbf{Images with High Text-to-Image Similarity for Class $"$Train$"$ in Cityscapes.} We use a SigLip 2 ViT-B/16 CLIP model to perform retrieval on Cityscapes. The depicted images are among the top 25 images with the highest text-to-image similarity for the class $"$train$"$ of which there are 171 units in the data in total. These images contain streets with streetcar tracks but no actual trains. This highlights the downside of an image-focused approach where the image features can be dominated by objects that are not actually from the target class.  }
    \label{fig:retrieval_example}
\end{figure}

In order to motivate the use of object-level features in OFDS, we compare to an additional baseline which uses image-level features. Therefore, we perform test-to-image retrieval on the unlabeled training datasets with a SigLip 2 ViT-B/16 model \cite{siglip2} with the highest available resolutions. We evenly distribute the annotation budget between all classes and retrieve the images with the highest text-to-image similarity using the prompts $"$a photo of a $\{$classname$\}$$"$. The major drawback of such an approach is that image-level features can be dominated by large or frequent objects and can be confounded by objects outside of the target classes. In Figure \ref{fig:retrieval_example}, we highlight the three images with the highest similarity to the class $"$train$"$ in the Cityscapes dataset. None of the images actually contains a train (even though there are several images with clearly visible trains in the dataset) but only streets with streetcar tracks. In Figure \ref{fig:retrieval}, we compare the model performances when training on subsets selected through retrieval in comparison to OFDS on the Cityscapes dataset. We observe a clear improvement of OFDS over the image-focused baseline for both object detection and semantic segmentation.

\section{Influence of Clustering Object Features} 
\label{sec:app_clustering}
In this section, we discuss the impact of clustering object features in OFDS in greater detail. As outlined in Section \ref{sec:feature_extraction}, the purpose of clustering the features and selecting individual objects close to the cluster centers is to obtain a density-based covering of the semantic feature space of the individual classes and ensure intra-class diversity. In Section \ref{sec:app_clustering_performance}, we compare the clustering-based object selection in OFDS to an ablated variant that uses random selection per class without clustering. In Section \ref{sec:app_clustering_duplicates}, we illustrate further examples of semantic object groups identified through the clustering.

\vspace{-.3cm}
\subsection{Influence on the Performance} \label{sec:app_clustering_performance}
\vspace{-.1cm}
\begin{figure*}
     \centering 
     \includegraphics[width=.7\columnwidth]{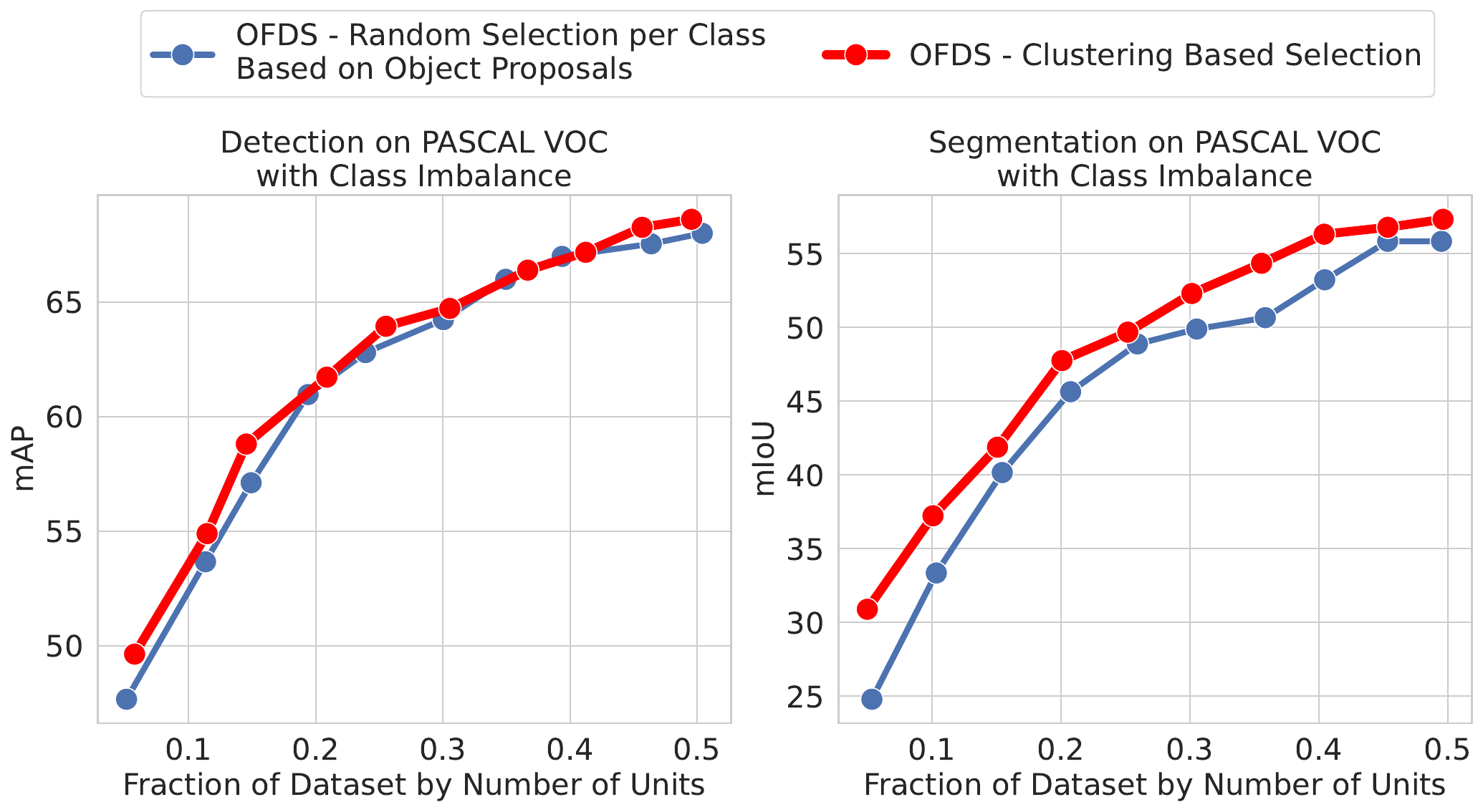}
    \includegraphics[width=.7\columnwidth]{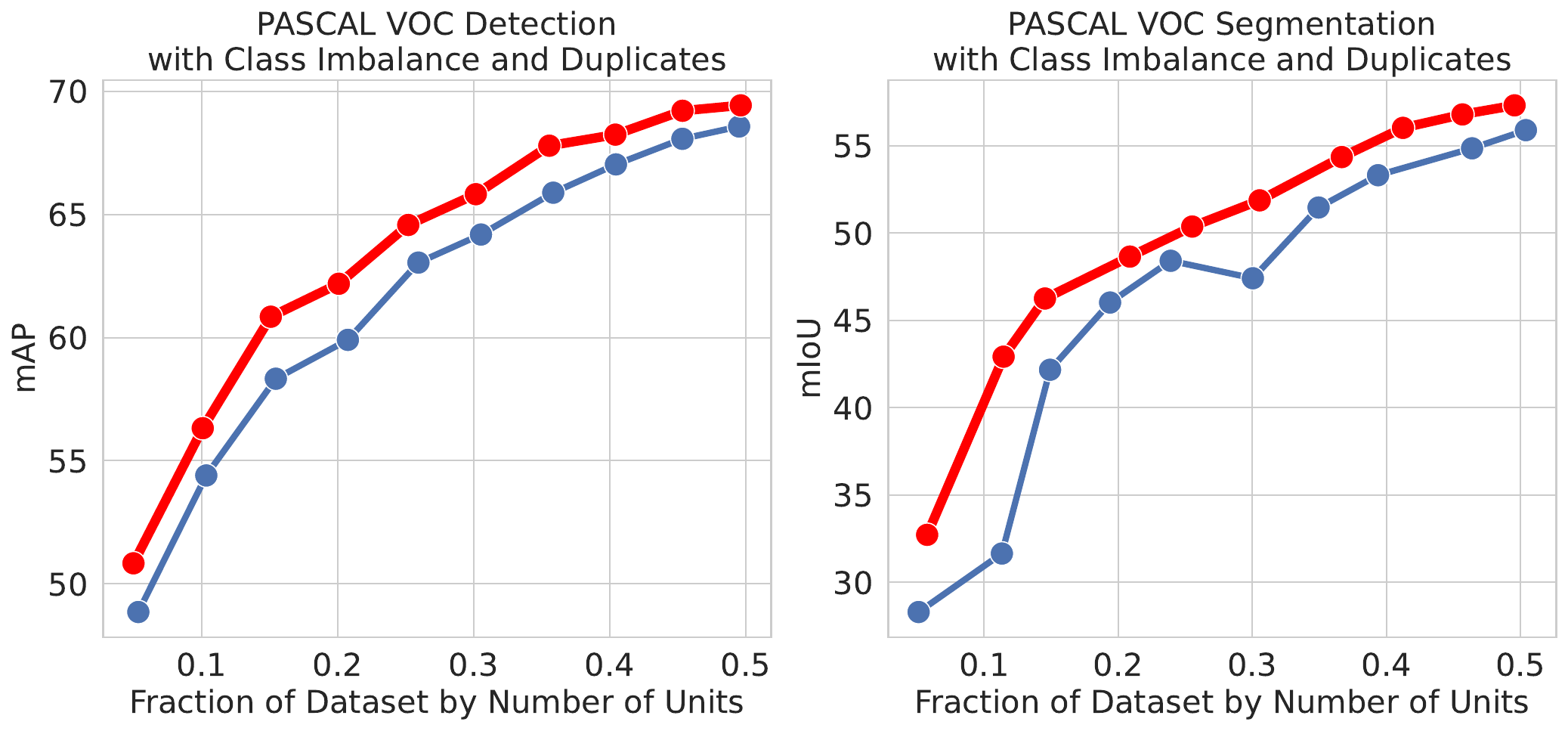} 
    
    \caption{\textbf{Influence of Clustering Object Features in OFDS.} We perform an ablation of OFDS where in Step 7 of Algorithm \ref{alg:ofds} the objects are chosen randomly per class from the object proposals instead of using the clustering-based selection. We compare these two variants of OFDS on the PASCAL VOC dataset with class imbalance. As an additional setting, we add 20$\%$ random duplicate images to the dataset. This is motivated by the fact that for example unlabeled web crawled datasets are known to feature a substantial amount of duplicates \cite{laion_dulicates}. In such case, it is particularly important to ensure that the objects in the selected subset are semantically diverse. We observe that the clustering-based object selection yields small but consistent performance improvements over the random selection of object proposals per class, in particular when duplicates are present. The results are obtained using the same model and training setup as in Section \ref{sec:data_selection_results}.}
    \label{fig:clustering_results_duplicates}
\end{figure*}

To assess the influence of the clustering of object features on the performance, we construct an ablation of OFDS where the objects from every class are chosen randomly from the object proposals instead of the clustering-based approach. More precisely, in Step 7 of Algorithm \ref{alg:ofds} we annotate images with $N_{C_l}$ randomly chosen objects from the current class. When using already annotated datasets like PASCAL VOC, the difference between these two variants of OFDS is diminished by the fact that the images and class objects in the datasets were selected and annotated by humans such that no or few very similar objects are contained. However, in datasets of e.g. webcrawled images such as LAION, up to 30$\%$ of images have been found to be exact or near duplicates \cite{semdedup,laion_dulicates}. As no dense human annotations are available for such datasets, we construct a comparable setting by adding 20$\%$ random duplicates to the PASCAL VOC datasets with class imbalance. The results for training downstream models for object detection and semantic segmentation are shown in Figure \ref{fig:clustering_results_duplicates} both with and without duplicates. We observe that the clustering steps leads to a small but consistent improvement in performance. Furthermore, the difference increases when training on the datasets with duplicates. Hence, we conclude that using OFDS with the object selection based on clustering of object features is favorable.
\vspace{-.2cm}
\subsection{Cluster Illustrations} 
\label{sec:app_clustering_duplicates}
\vspace{-.2cm}
\begin{figure*}[tp]
    \centering
    \includegraphics[width=0.95\linewidth]{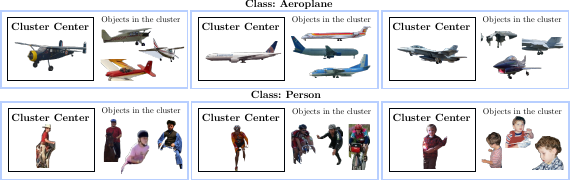}
    \includegraphics[width=0.95\linewidth]{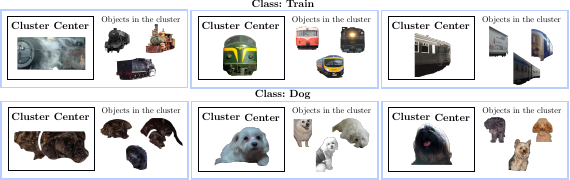}
    \vspace{-.3cm}
    \caption{\textbf{Illustration of Clusters for Classes from PASCAL VOC.} The clustering was performed with $N_{C_l}=30$ clusters per class. The resulting clusters exhibit the semantic consistency which we use to construct a uniform covering of the target classes.}
    \label{fig:illustration_clustering_additional}
    \vspace{-.8cm}
\end{figure*}

In Section \ref{sec:feature_extraction}, we motivate the use of unsupervised clustering to construct a density-based covering of the class semantics. In particular, the clustering is used to group semantically similar objects. Figure \ref{fig:illustration_clustering_additional} further illustrates this aspect with object clusters for three classes from PASCAL VOC.

\section{Further Details on the Cold Start Problem for Active Learning} \label{sec:active_learning}

\subsection{Experimental Setup}
\vspace{-.1cm}
For the experiments on active object detection using PPAL, we use a RetinaNet \cite{retinanet} with ResNet50 backbone and the same training hyperparameters as in the original publications. We start with a subset consisting of 2.5\% of the overall images and add 400 images (roughly 5\%) in every active learning round. For semantic segmentation we use the training setup from the benchmarking study \cite{al_segmentation}. The segmentation model is based on the DeepLabv3+ \cite{retinanet} architecture with a Wide-ResNet38 \cite{wideresnet} backbone. We start with a subset consisting of 10\% of the overall images and add (approximately) 10\% in every active learning round.  Note that Mittal et al. \cite{al_segmentation} use weaker augmentations than we used for our main experiments in Section \ref{sec:experiments}. For the active learning experiments, the budget is counted by the number of labeled images due to the conception of the used frameworks. 

\subsection{Combining Autolabels with OFDS for the Cold Start Problem}
\begin{figure}
     \centering
    \includegraphics[width=.4\columnwidth]{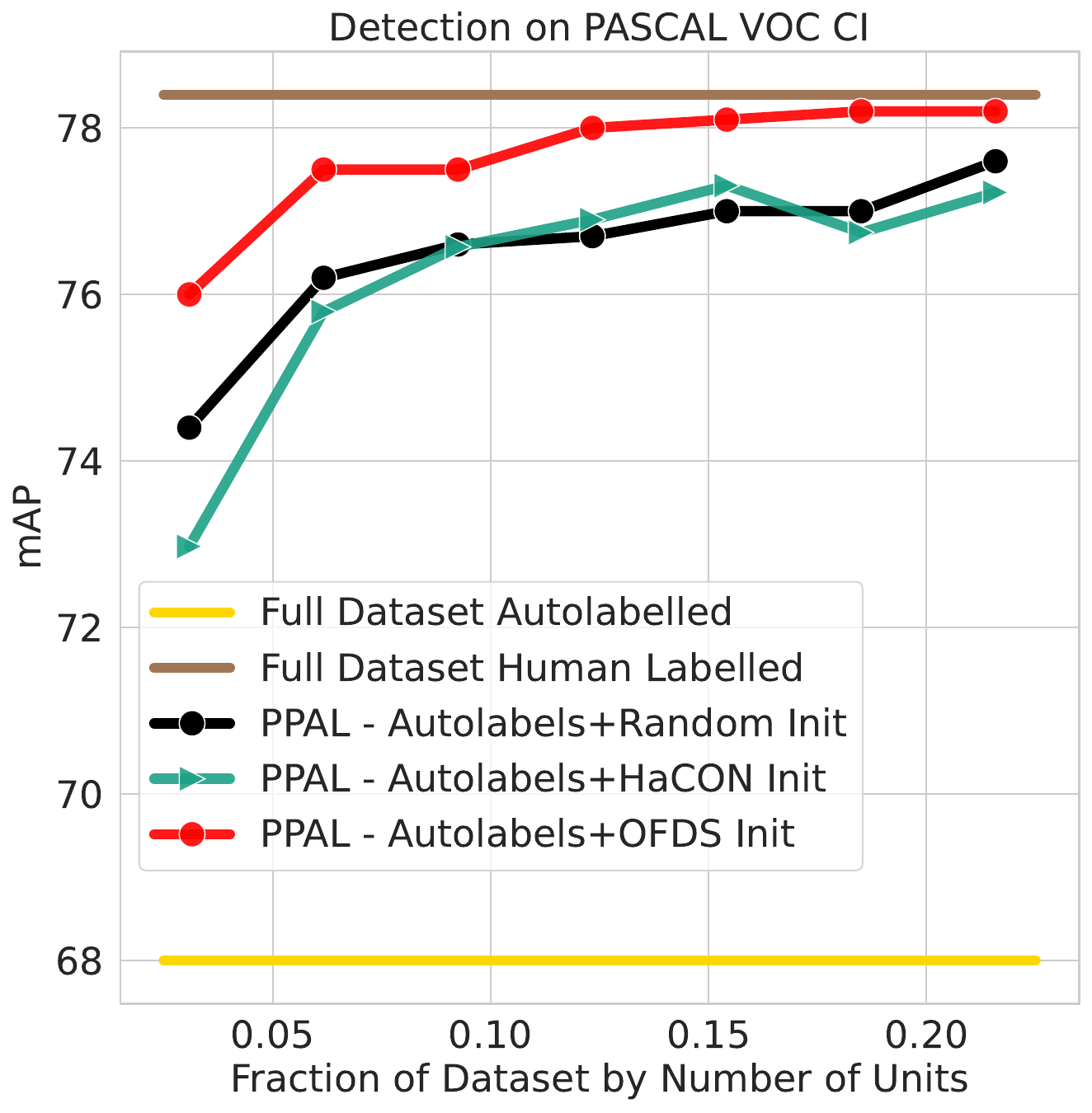}
    \vspace{-.2cm}
    \caption{\textbf{Leveraging Autolabels Together with OFDS further improves the Cold Start Problem.} The models are trained with the same setup as for Figure \ref{fig:al_main} but the model is first pre-trained with autolabels before fine-tuning it on the initial labeled data and subsequently performing active learning. Importantly, it is not possible to start active learning with PPAL only with autolabels as the selection of additional datapoints requires a labeled pool to compare to. The only difference between the three lines is the initial dataset. We observe that selecting the initial labeled dataset with OFDS yields the best results. }
    \label{fig:al_autolabels}
\end{figure}

In this section, we incorporate autolabels into the cold start problem for active learning. For methods like PPAL \cite{ppal} or \cite{active_learning_coreset}, the cold start for active learning cannot be started just with autolabels. This is due to the fact that these active learning methods base the selection of new datapoints to label on an already existing set of labeled images and still require an initial pool of labeled images. Thus, we start active learning methods using a model checkpoint trained on the entire dataset with autolabels together with an initial labeled subset. In Figure \ref{fig:al_autolabels} we perform active learning with PPAL for object detection on PASCAL VOC with class imbalance and initial datasets selected through OFDS, HaCON and random drawing. We observe that pre-training with autolabels substantially improves the performance over training the model from scratch in the initial round. Furthermore, selecting the initial dataset through OFDS improves the performance of the entire training in comparison to random selection.

\begin{figure}
     \centering
    \includegraphics[width=.7\columnwidth]{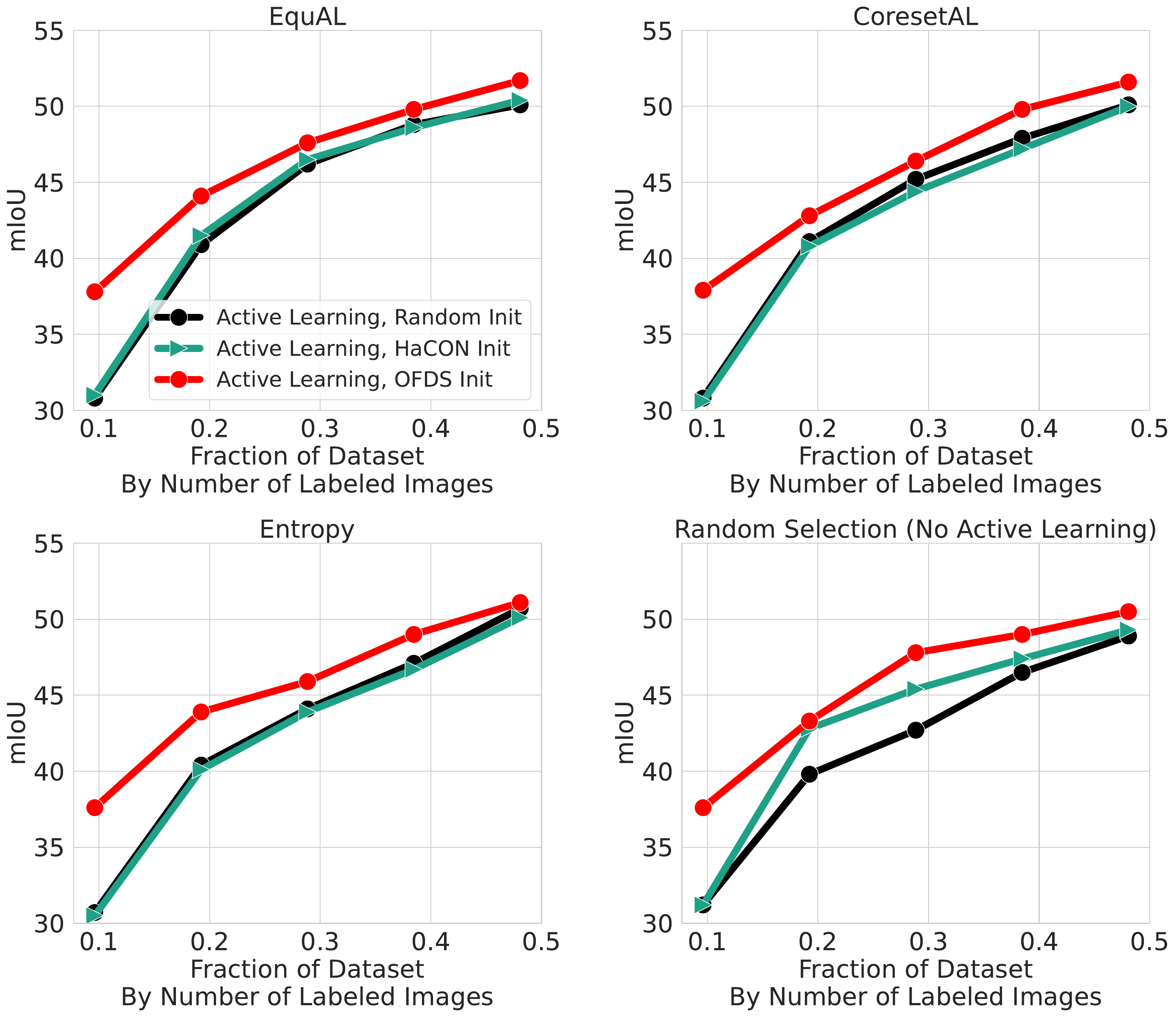}
    \vspace{-.2cm}
    \caption{\textbf{Different Initial Labeled Datasets for Active Semantic Segmentation on PASCAL VOC with Class Imbalance.} We train a DeepLavV3+ model with Wide-ResNet38 backbone on PASCAL VOC with class imbalance using the setup from Mittal et al. \cite{al_segmentation}. For every plot, the models were trained with the same active learning frameworks only with different initial datasets. Random selection refers to randomly selecting the additional images in every round. For all frameworks apart from random selection, OFDS improves the performance of the model during the active learning process.}
    \vspace{-.6cm}
    \label{fig:al_methods}
\end{figure}
\begin{figure}
     \centering

    \includegraphics[width=.7\columnwidth]{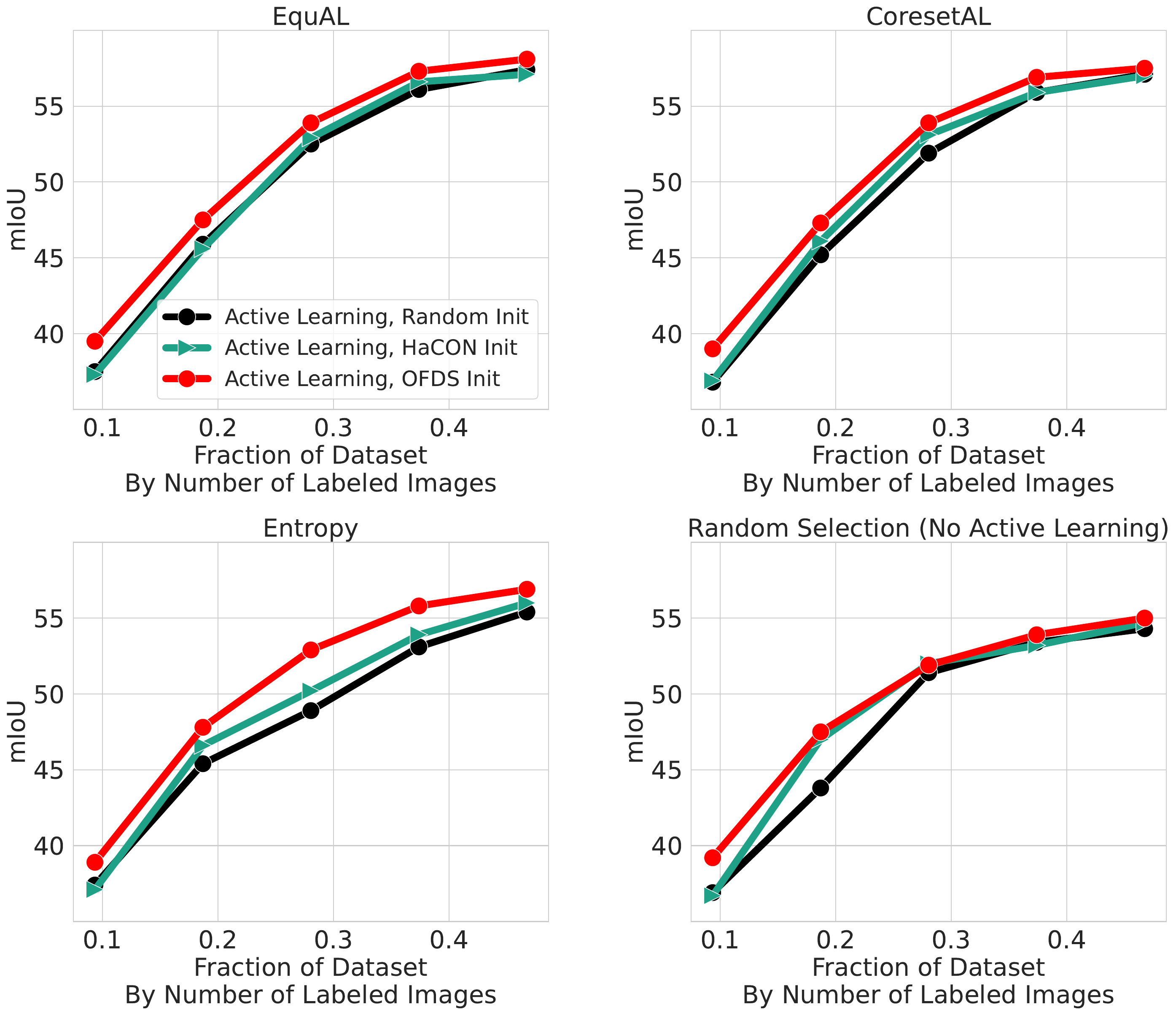}
    \vspace{-.2cm}
    \caption{\textbf{Different Initial Labeled Datasets for Active Semantic Segmentation on Cityscapes.} We use the same setup as in Figure \ref{fig:al_methods} to train models for semantic segmentation on Cityscapes. We again observe that for all frameworks apart from random selection, selecting the initial dataset through OFDS improves active learning.}
    \label{fig:al_methods_cs}
    \vspace{-.6cm}
\end{figure}

\subsection{Further Active Learning Methods}
In this section, we provide further results for cold start problem of different active learning methods. Due to the vast amount of methods for active learning, we restrict ourselves to the tasks of semantic segmentation and assess the cold start problem on two datasets using four baselines from a benchmarking study \cite{al_segmentation}. This includes the EquAL framework \cite{equal}, which achieved the most consistent results in the benchmarking study \cite{al_segmentation}, as well as entropy-based active learning \cite{entropy_al_1,entropy_al_2}, coreset \cite{active_learning_coreset} and the random baseline. Entropy and coreset are two of the most commonly used frameworks for active learning and remain strong baselines even in more recent works \cite{ppal}. Coreset for active learning differs from K-Centers for data selection from Section \ref{sec:feature_extraction}. Coreset for active learning, which we refer to as CoresetAL for clarity, utilizes the features of the downstream model being trained while K-Centers for data selection uses features from a pre-trained self-supervised model, in our case DINO. In Figures \ref{fig:al_methods} and \ref{fig:al_methods_cs}, we report the results for semantic segmentation on PASCAL VOC with class imbalance. We observe that for all frameworks apart from random selection, OFDS improves the performance of the model during the entire active learning process. This validates our observation from Section \ref{sec:experiments}.

\subsection{Direct Comparison of OFDS and Active Learning}
\begin{figure}
     \centering
    \includegraphics[width=.7\columnwidth]{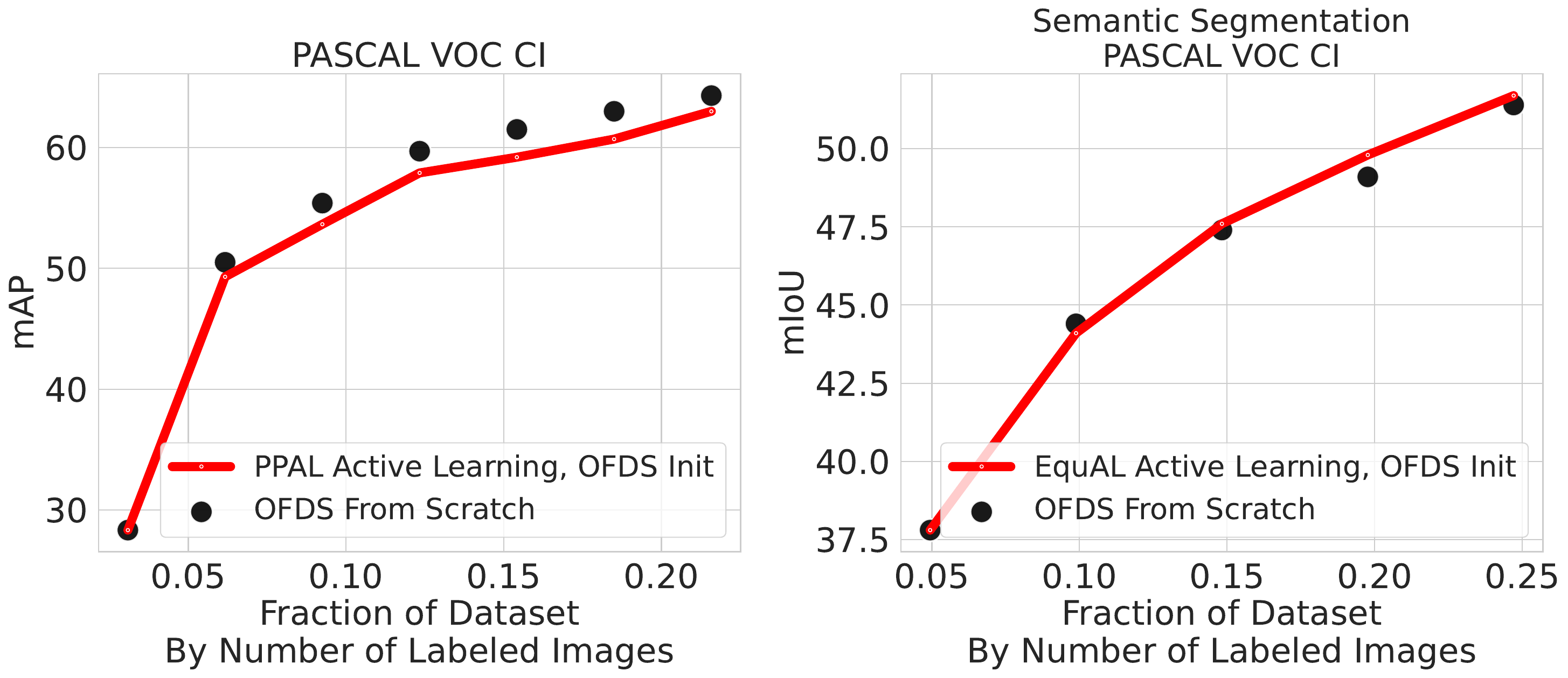}
    \caption{\textbf{Direct Comparison of OFDS and Active Learning.} We compare the models trained with the active learning frameworks from Figure \ref{fig:al_main}, where the dataset sizes are iteratively enlarged, to the same models trained from scratch on datasets selected through OFDS. We observe that training on the subsets selected through OFDS without any active learning achieves results that are on par to the active learning methods.}
    \label{fig:al_vs_ofds}
    \vspace{-.4cm}
\end{figure}

In Figure \ref{fig:al_vs_ofds}, we compare training a model with PPAL active learning for object detection and EquAL for semantic segmentation to a model trained from scratch on subsets selected with OFDS. We observe that on PASCAL VOC with class imbalance, training a model from scratch on data selected through OFDS yields results which are on par with the tested active learning methods. Importantly, we do not claim that training models on data selected through OFDS is generally on par with active learning as there is a vast amount of literature on active learning methods, many of which are highly optimized for specific tasks \cite{active_semseg}, model architectures \cite{al_transformer} or even datasets \cite{al_imagenet} to reach the best performance. Adequately comparing to this line of methods requires extensive experiments which go beyond the scope of this work. Instead, we highlight once more that data selection through OFDS is qualitatively different from active learning as it does not require an initial dataset and is both model and task agnostic. 

\begin{figure*}
     \centering
    \includegraphics[width=.7\columnwidth]{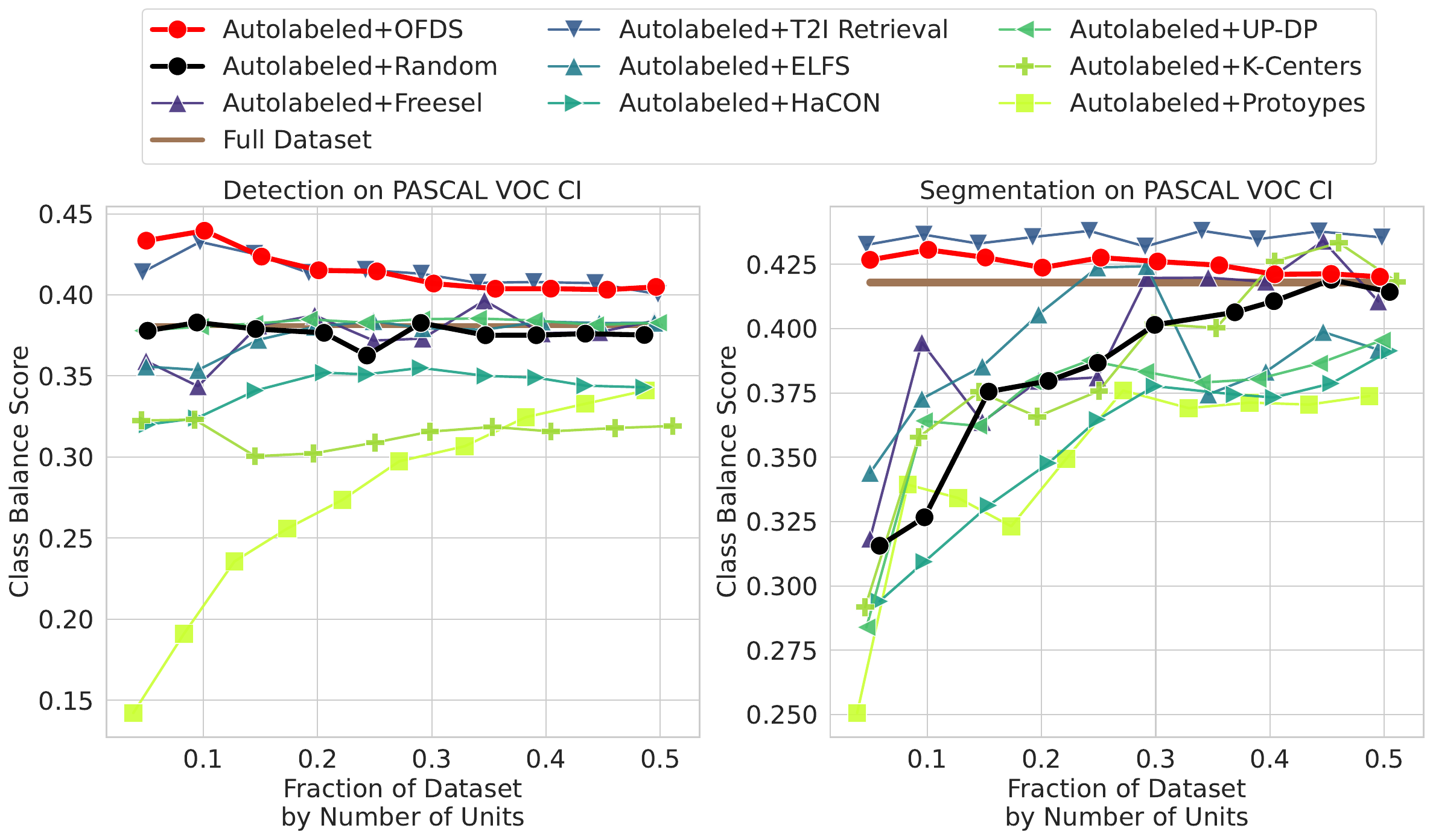}
    \includegraphics[width=.7\columnwidth]{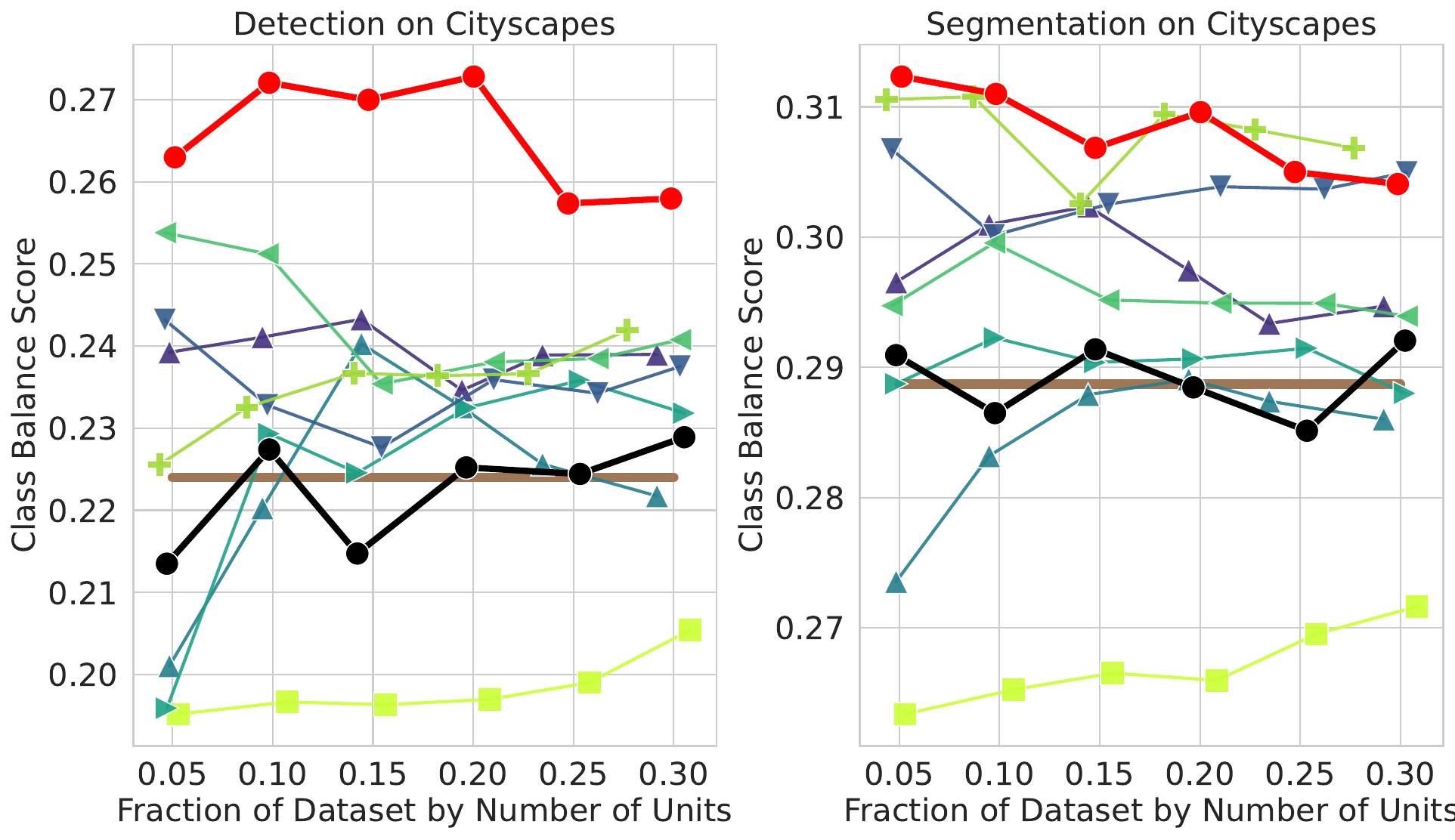}
    \includegraphics[width=.35\columnwidth]{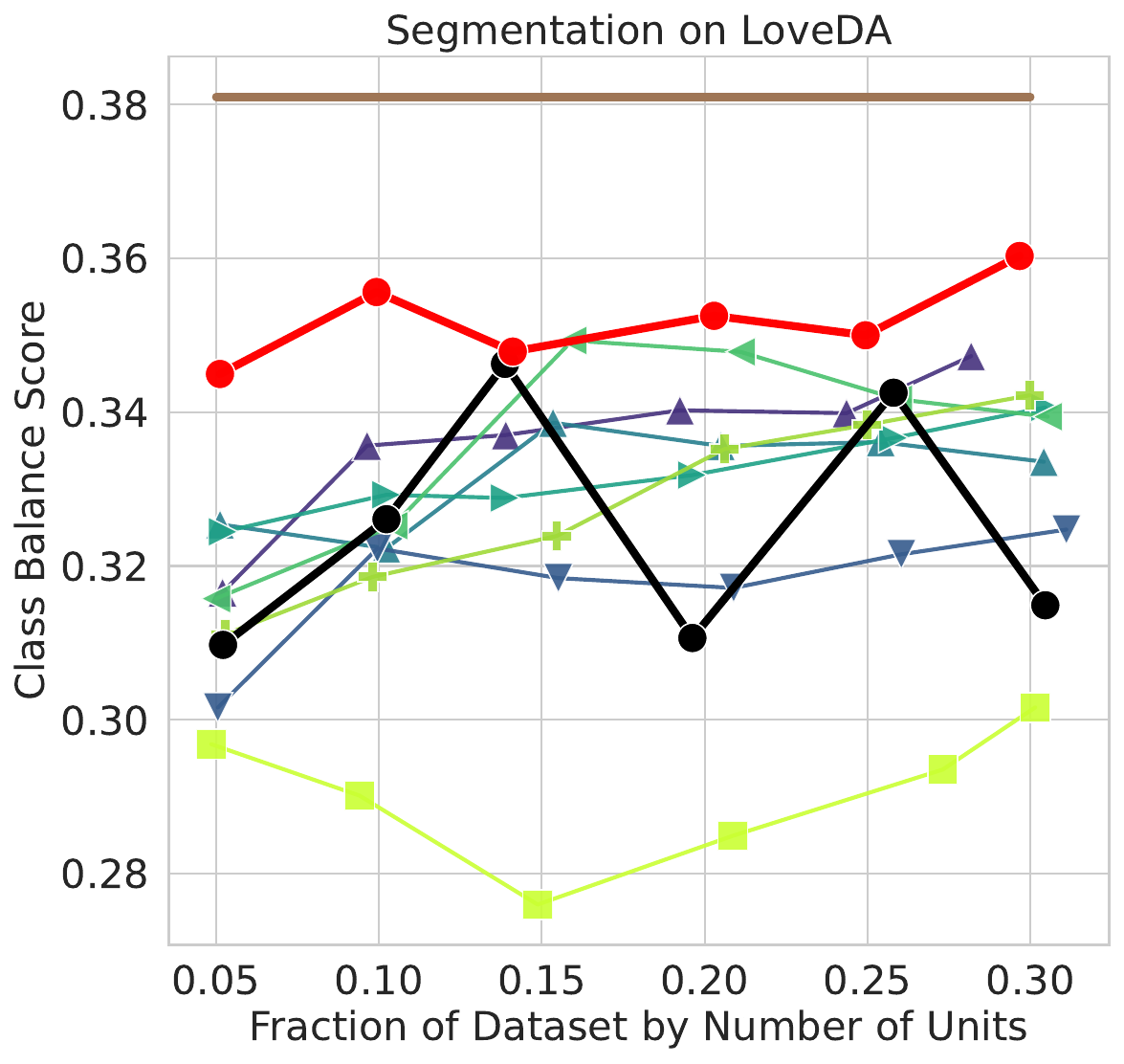}
    \caption{\textbf{OFDS Selects More Class-Balanced Subsets.} We compute the class balance score introduced by Sorscher et al. \cite{beyond} and find that the subsets selected by OFDS consistently feature higher class balance scores than the baselines. This indicates more evenly distributed class distributions.}
    \label{fig:app_class_balance_scores}
\end{figure*}

\section{Class Balance Scores} \label{sec:app_class_balance_scores}
To quantify how balanced the class distributions of the selected subsets are, we compute the class balance scores introduced by Sorscher et al. \cite{beyond}. The class balance score $b\in [0,1]$ is defined as the average balance between any two pairs of classes. It is determined by taking the expectation of drawing two random classes and computing the fraction between the number of objects in the smaller class in comparison to the larger class. A balance score of $1$ corresponds to evenly balanced classes and higher scores are generally better. In Figure \ref{fig:app_class_balance_scores}, we show the balance scores for subsets selected from PASCAL VOC with class imbalance, Cityscapes and LoveDA using the 8 baselines and OFDS. The subsets selected by OFDS consistently feature higher class balance scores compared then the baselines as well as the full datasets. Only on PASCAL VOC, using text-to-image retrieval achieves comparable scores. This can be explained by the fact that the number of objects per image on PASCAL VOC images is lower than on Cityscapes or LoveDA. Overall, the results demonstrates the effectiveness of OFDS in selecting subsets with improved class balance. Since some of the classes in the segmentation split of PASCAL VOC with class imbalance contain only very few objects, these classes are not represented in the small subsets selected through random drawing. As a result, the balance scores for random selection on smaller subsets are lower than the score of the full dataset.

\section{Repeated Dataset Selection with FreeSel and Random} \label{sec:app_repeated}

\begin{figure}[bp!]
     \centering
    \includegraphics[width=.9\columnwidth]{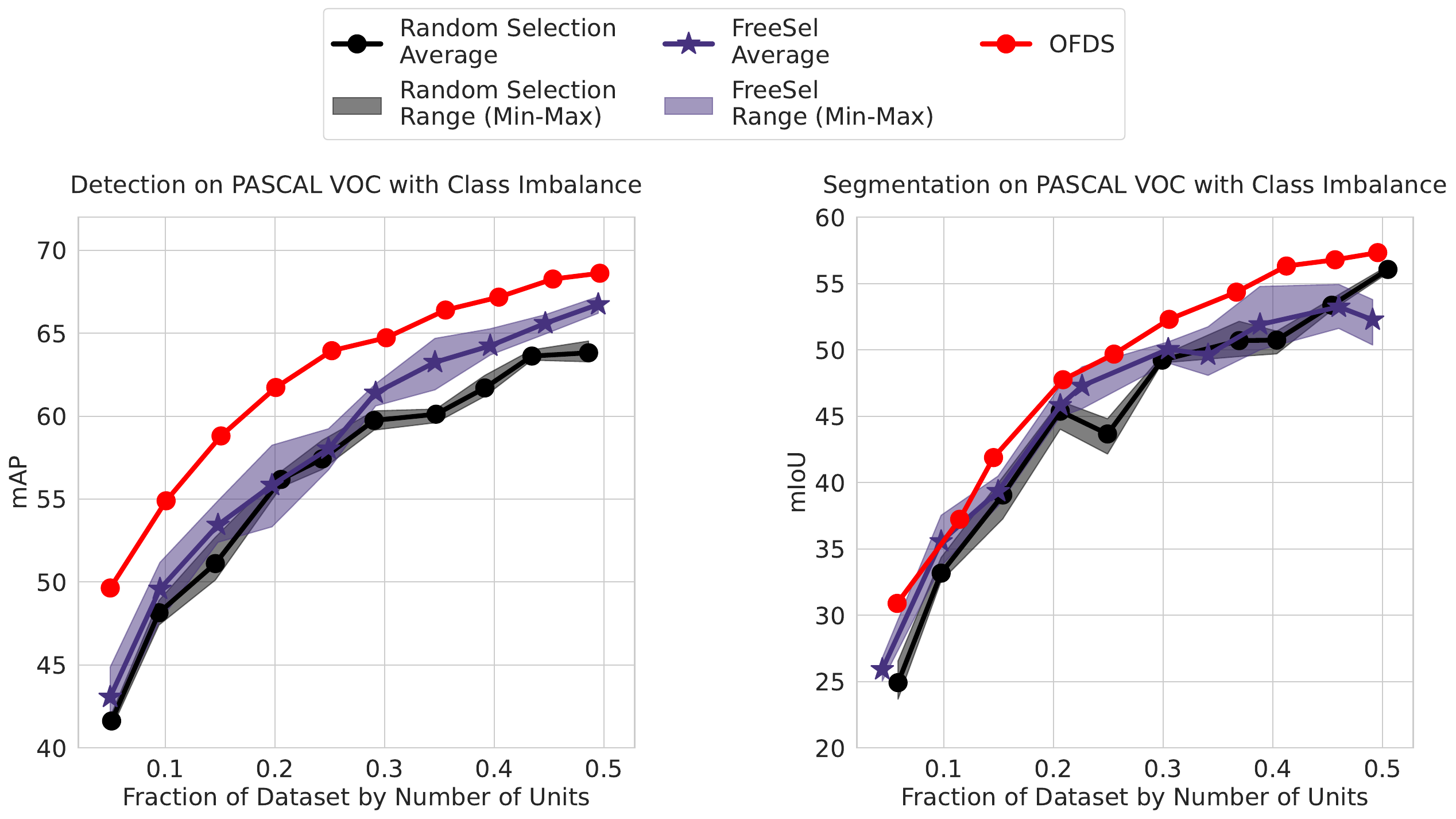}
    \vspace{-.2cm}
    \caption{\textbf{Repeated Data Selection For Baselines With Randomness.} Since the random baseline and FreeSel are based on probabilistic selection, we repeat the data selection process for object detection and semantic segmentation on PASCAL VOC with class imbalance three times. The solid lines represent the mean performance and the shaded areas indicate the range between the minimum and maximum performance for each subset size. The improvement achieved by OFDS over the baselines consistently exceeds the fluctuations resulting from the randomness in their selection processes. OFDS is included for reference but features a fully deterministic selection.}
    \label{fig:app_repeats}
    \vspace{1.8cm}
\end{figure}

Since both the random baseline and FreeSel rely on a probabilistic selection process, we repeat the data selection multiple times to assess the extent of the resulting fluctuations. Due to the high computational cost of repeating all experiments, we focus on object detection and semantic segmentation on PASCAL VOC with class imbalance and perform the selection three times for every subset size. The results are depicted in Figure \ref{fig:app_repeats}. The performance improvement achieved by OFDS over the baselines clearly surpasses the fluctuations caused by the randomness in FreeSel or random selection. Importantly, OFDS features a deterministic selection process, which is advantageous for practical applications where the data selection can only be performed once.

\section{Using a Stronger Object Proposer} \label{sec:ablation_object_proposer}

\begin{figure*}
     \centering
    \includegraphics[width=0.45\textwidth]{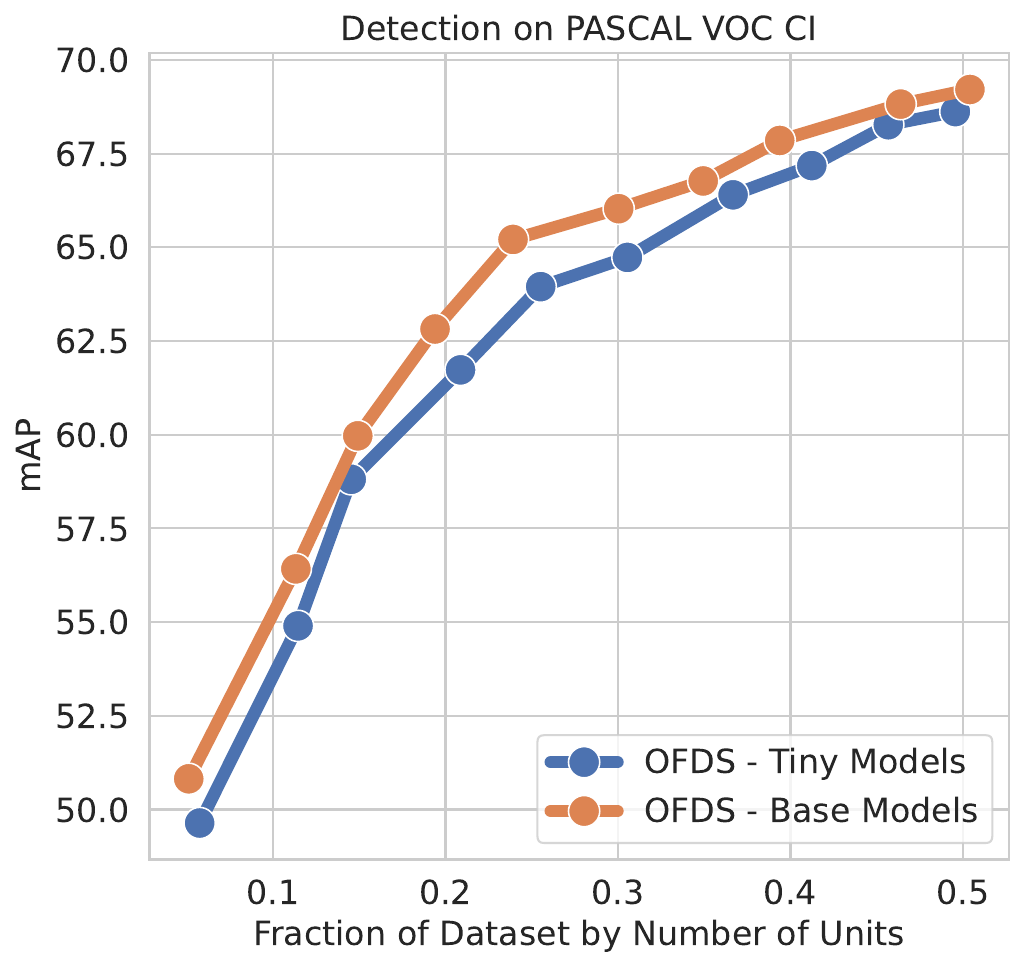}
    \vspace{-.3cm}
    \caption{\textbf{Using a Stronger Object Proposer Further Improves OFDS.} We exchange the Grounding SAM2 tiny model with the Grounding SAM2 base model which features a larger architecture and was trained on more data. In Section \ref{sec:experiments}, we employ Grounding SAM2 tiny to provide a fair comparison to the baselines which utilize DINO-T or DINO-S models and minimize the computational cost. }
    \label{fig:object_proposer_base_model}
\end{figure*}

In Section \ref{sec:experiments}, we employ Grounding SAM2 with the DINO-T backbone as the object proposer. This choice is motivated by two primary reasons. First, it ensures a fair comparison with baselines that utilize features from a DINO-T model. Second, the DINO-T backbone offers the highest throughput among the Grounding SAM2 models, minimizing the computational costs of extracting object features (further details on the computational cost of OFDS are discussed in Section \ref{sec:app_cost}). To investigate the effect of a stronger object proposer, we conduct an ablation study using object features from the Grounding SAM2 base models on the PASCAL VOC dataset with class imbalance. These models not only have more parameters but were also trained on larger datasets. The results for the downstream models trained on the selected subsets are presented in Figure \ref{fig:object_proposer_base_model}. Using a stronger object proposer yields small but consistent improvements for OFDS. We emphasize that, in its current state as used in Section \ref{sec:experiments}, OFDS consistently outperforms all baselines for object-level prediction tasks, even in specialized domains. Future enhancements in foundation models, which could yield stronger object proposers, have the potential to further improve the effectiveness of OFDS.

\section{Class Distributions and Rare Classes} \label{sec:app_subsets}
In this section, we provide details on the class distributions as well as the subsets selected for the evaluation on rare classes.
\subsection{Class Distributions}
\begin{figure*}
     \centering
    \includegraphics[width=0.48\textwidth]{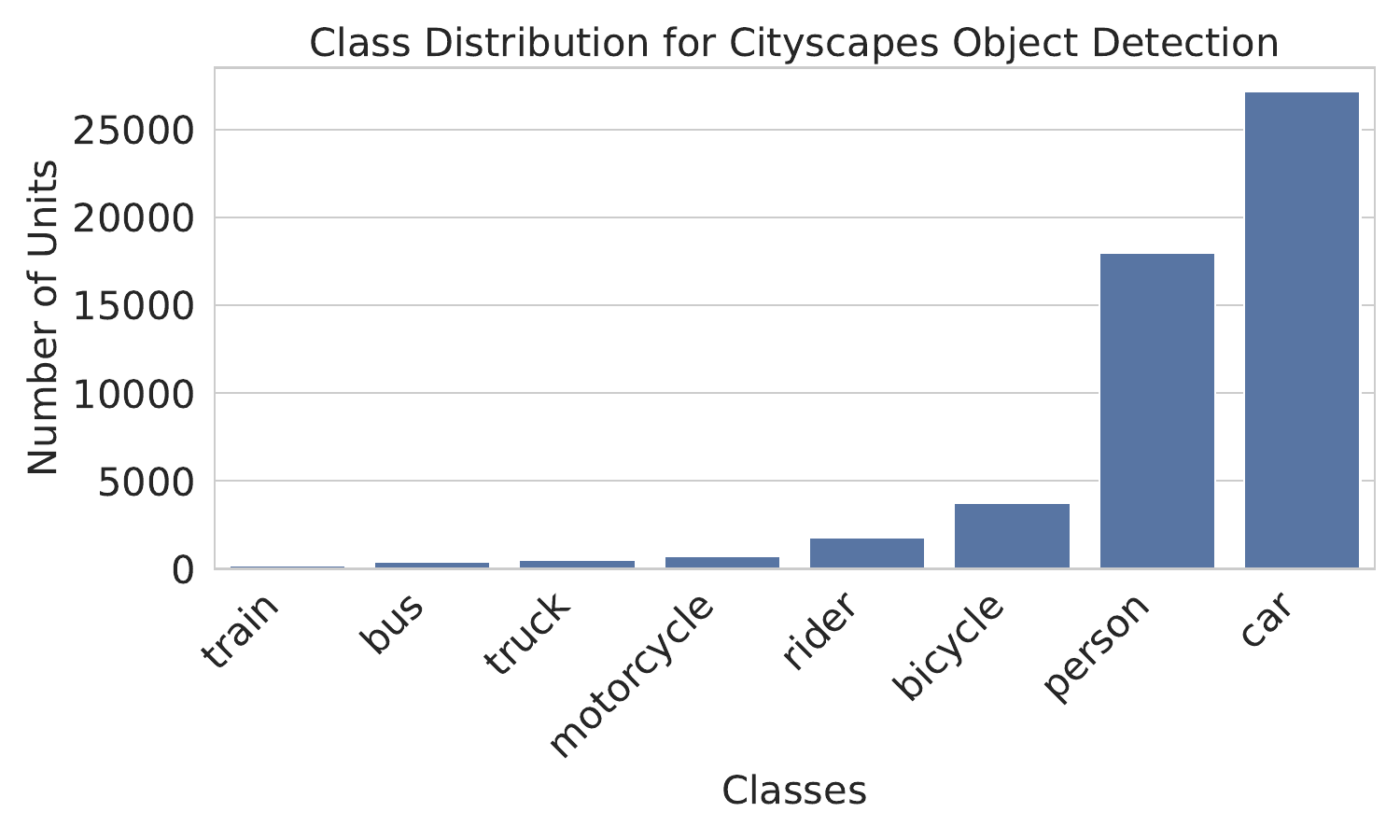}
    \includegraphics[width=0.48\textwidth]{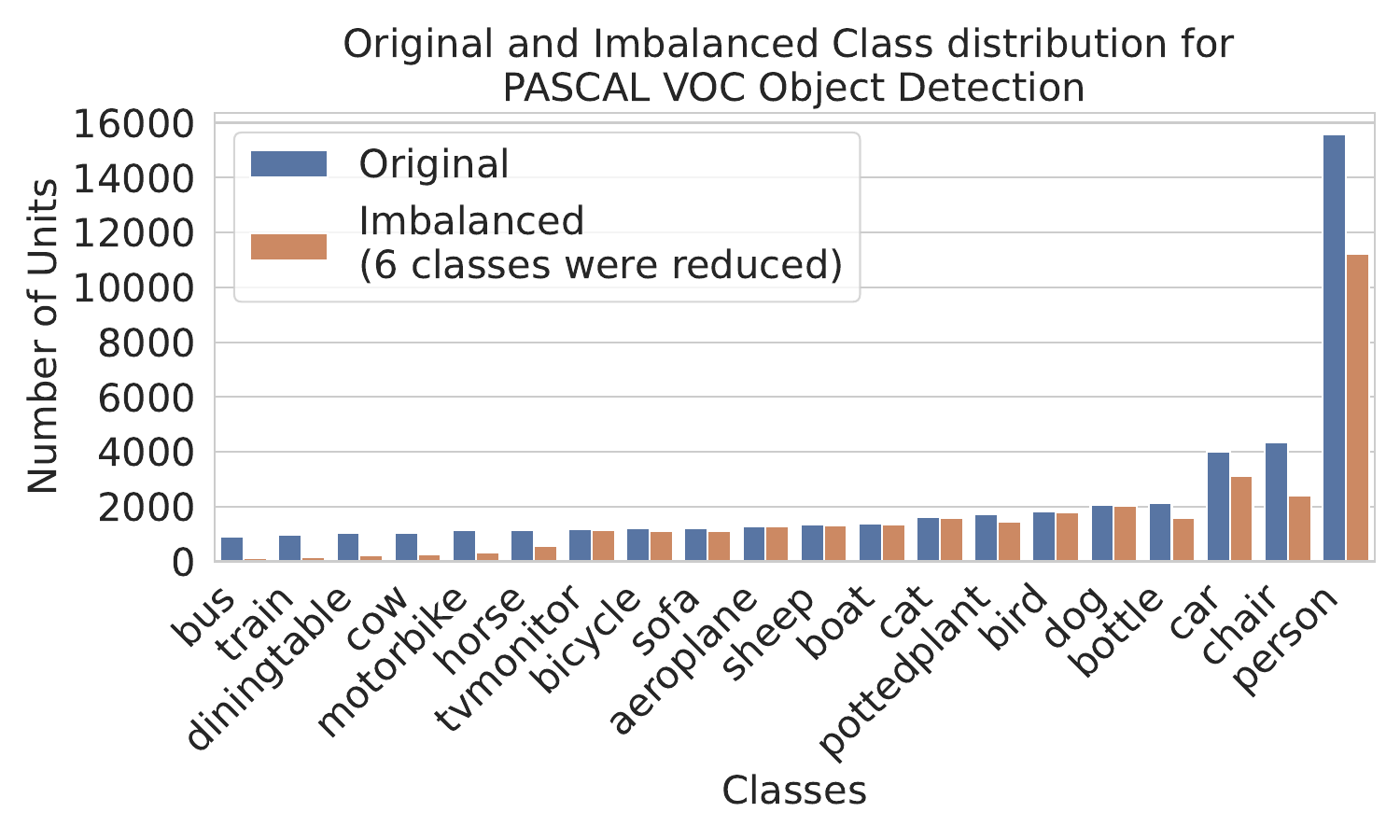}
    \vspace{-.3cm}
    \caption{\textbf{Class distributions for the Object Detection Datasets.}}
    \label{fig_class_distribution_object_detection}
\end{figure*}

\begin{figure*}
     \centering
    \includegraphics[width=0.48\textwidth]{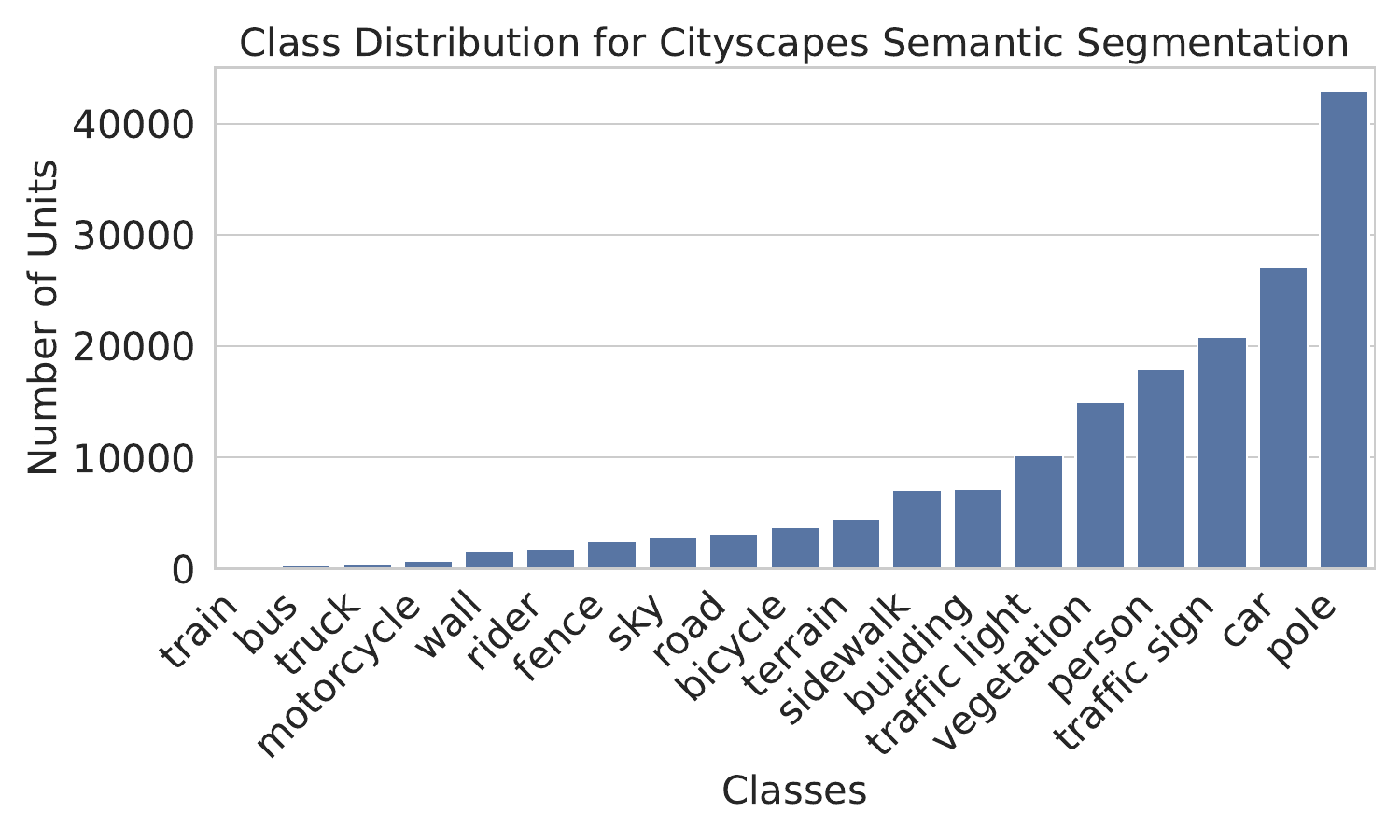}
    \includegraphics[width=0.48\textwidth]{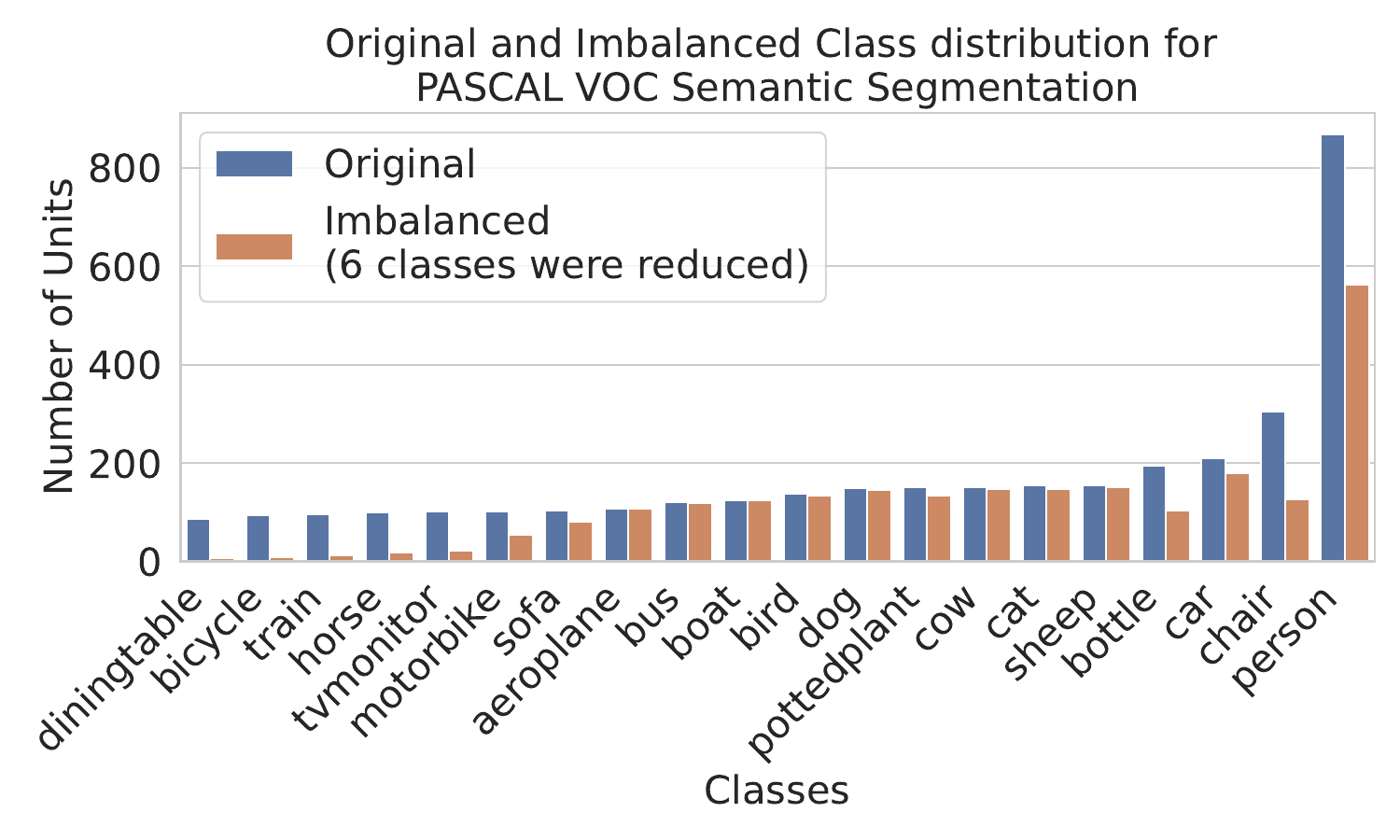}
    \vspace{-.3cm}
    \caption{\textbf{Class distributions for the Semantic Segmentation Datasets.}}
    \label{fig_class_distribution_segmentation}
\end{figure*}

\begin{figure*}
     \centering
    \includegraphics[width=0.6\textwidth]{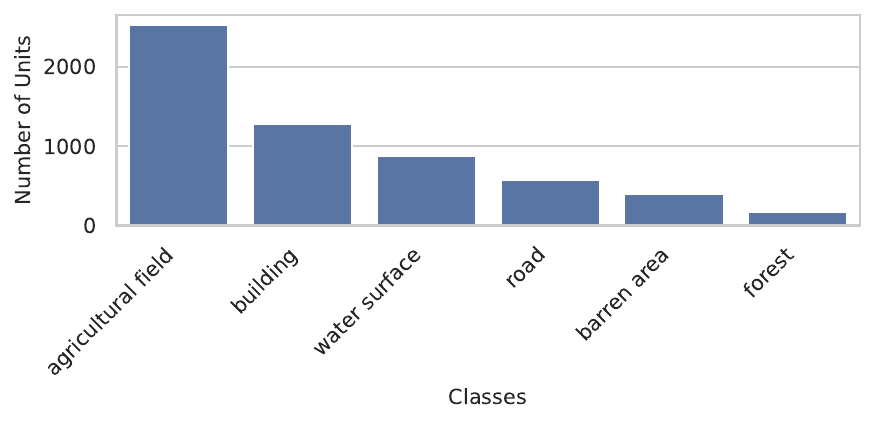}
    \vspace{-.3cm}
    \caption{\textbf{Class distributions for the LoveDA dataset.}}
    \label{fig:loveda_class_distr}
\end{figure*}

Figures \ref{fig_class_distribution_object_detection}, \ref{fig_class_distribution_segmentation}  and \ref{fig:loveda_class_distr} display the class distributions for the datasets used in Section \ref{sec:experiments}. The distributions for the Cityscapes and LoveDA are naturally imbalanced. The PASCAL VOC dataset features a more balanced class distributions. As unlabeled datasets in practice typically features imbalanced data distributions, we introduce two additional setting by pruning the six smallest classes of the PASCAL VOC datasets by 99$\%$, 95$\%$, 85$\%$, 80$\%$, 75$\%$ and 50$\%$. These datasets are referred to  as PASCAL VOC CI (class imbalanced). Note that for correctly computing the annotation budget for semantic segmentation, multiple masks on the same image need to be counted separately. This results in a different object distribution for LoveDA than stated in the original publication.

\subsection{Rare Classes}
For the evaluation on rare classes of PASCAL VOC with class imbalance, we consider the six classes that were pruned from the full dataset. On the Cityscapes dataset, we focus on the smallest four classes for object detection, representing half of the total classes. For semantic segmentation, we evaluate the performance on the five smallest classes, accounting for one fourth of the total classes. In both cases on Cityscapes, the rare classes jointly contain less than 3.5$\%$ of the overall number of objects.
\begin{itemize}
    \item \textbf{Rare Classes for PASCAL VOC Object Detection:} bus, train, diningtable, cow, motorbike, horse
    \item \textbf{Rare Classes for PASCAL VOC Semantic Segmentation:} diningtable, bicycle, train, horse, tvmonitor, motorbike
    \item \textbf{Rare Classes for Cityscapes Object Detection:} train, bus, truck
    \item \textbf{Rare Classes for Cityscapes Semantic Segmentation:} train, bus, truck, motorcycle, wall
\end{itemize}

\section{Calibrating the Foundation Model for Generating Autolabels and Object Proposals} \label{sec:app_cal}
\subsection{Calibration Data}\label{sec:app_cal_data}
As discussed in Section \ref{sec:experiments}, we utilize a subset of the MSCOCO \cite{MSCOCO} validation split to calibrate the object proposer. Therefore, we select only images containing objects from classes related to the target classes. Since MSCOCO does not feature the same classes as the target datasets, we manually identify and select these related classes:
\begin{itemize}
    \item \textbf{PASCAL VOC:} airplane, bicycle, bird, boat, bottle, bus, car, cat, chair, cow, diningtable, dog, horse, motorcycle, person, potted plant, sheep, couch, train, tvmonitor
    \item \textbf{Cityscapes:} car, bus, truck, motorcycle, bicycle, traffic light
\end{itemize}
Importantly, these classes are only used to calibrate the confidence threshold. For generating the object proposals and the autolabels, we use the actual target classes. As the LoveDA dataset features specialized classes from remote sensing for which no calibration data is available, we employ the same thresholds as for the Cityscapes dataset.

\subsection{Threshold Calibration of the Object Proposer}
\begin{figure}[tp]
     \centering
    \includegraphics[width=.7\columnwidth]{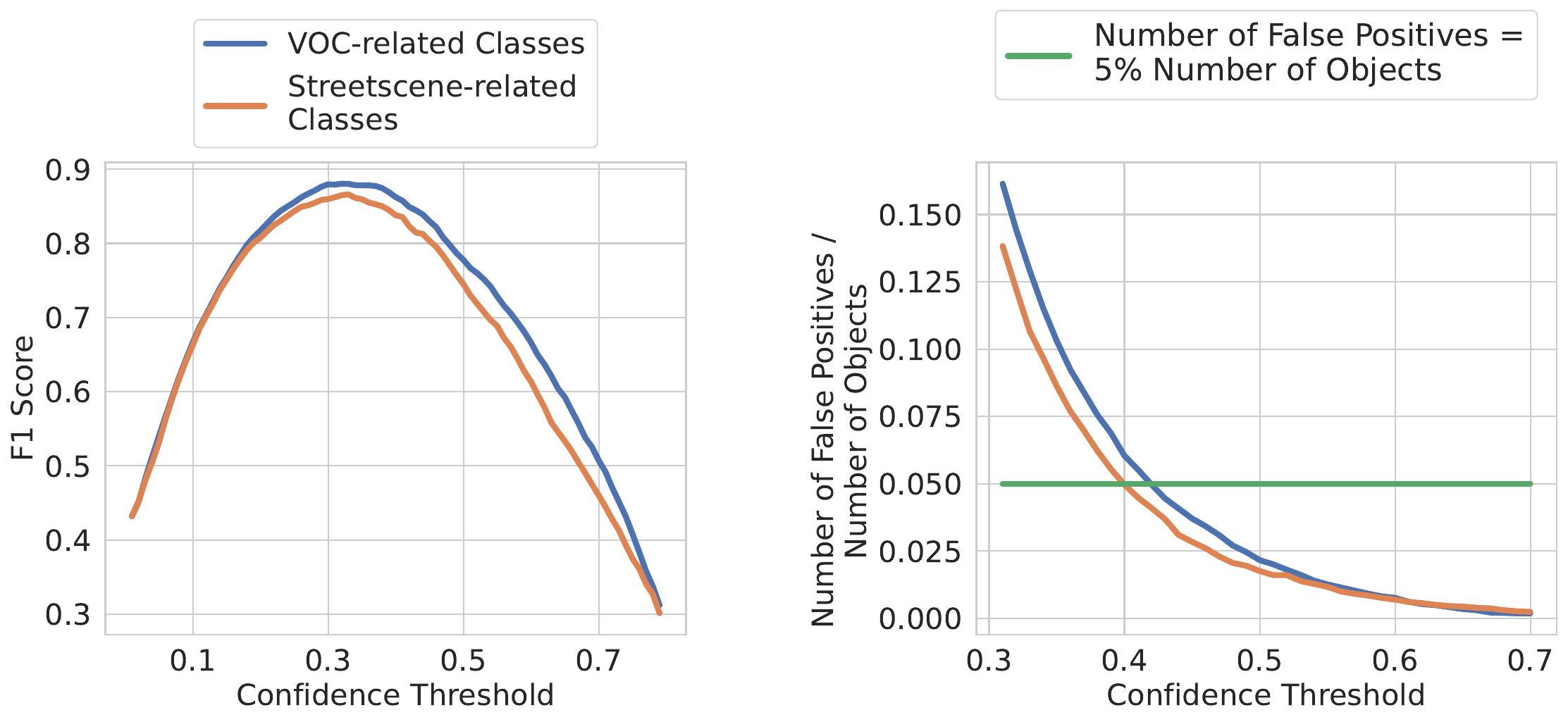}
    \caption{\textbf{Calibration of the Object Proposer} using the FPR to control the precision for OFDS and the F1 score for autolabeling. The calibration data is as described in Section \ref{sec:app_cal_data}}
    \label{fig:calibration}
    \vspace{-.4cm}
\end{figure}

In Figure \ref{fig:calibration}, we visualize the threshold calibration of the foundation model for autolabeling and for generating the object proposer. The threshold bounds the confidence with respect to the bounding boxes and the final class predictions are obtained by taking an argmax over the text scores. When generating autolabels, we use the threshold which achieves the highest F1 score on the calibration data. For generating object proposals, we use a threshold that yields 5$\%$ false positives on the calibration data to control the precision of the object proposals.

\section{Implementation Details} \label{sec:app_implementation}
In this section, we provide further details on the implementation of OFDS and K-Centers.

\textbf{OFDS} In step 5 of Algorithm \ref{alg:ofds}, the number of objects to select is set as $N_{C_l}=\frac{B-N(\mathcal{S})}{(M-l+1)N_O}$. Here, $N_{C_l}$ is determined from the leftover budget of annotation units $B-N(\mathcal{S})$ which is updated after every class. This budget is equally distributed between the $M-l+1$ remaining classes at every iteration of step 5. Therefore, we divide the leftover budget by the number of remaining classes to obtain the annotation budget per class. To obtain the number of objects to annotate from this annotation budget, we further divide by $N_O$ which is the expected number of annotations per selected object. \\
Since we select objects only from clusters that do not contain any annotated objects from previous steps, the number of selected objects does not necessarily correspond to the number of clusters. Instead, we initialize the number of clusters by $N_{C_l}$ and gradually increase it until we find $N_{C_l}$ clusters without any annotated objects. In practice, we achieve this by iteratively multiplying the number of clusters with 1.05 until enough clusters are present.\\
Furthermore, in step 1 of Algorithm \ref{alg:ofds} we only consider object proposals with bounding boxes smaller than 0.05$\%$ of the overall image area to filter out noisy proposals.

\textbf{K-Centers} When using the K-Centers algorithm to select subsets that are relatively large compared to the full dataset, the complexity becomes prohibitive due to the quadratic cost of computing the distances between all points in the selected subset and the non-selected subset. To overcome this problem, we use a batched version that considers batches of size 512 when selecting new points for the subset. Thereby, the complexity becomes independent of the size of the unlabeled image pool.

\textbf{Text-to-Image Retrieval} When performing text-to-image retrieval for data selection, we use the top-k images whose image embedding have the highest similarity to the text embedding of the prompt ``a photo of a $\{$classname$\}$''. This is done subsequently for every class. We iterate through the classes based on the alphabetical order.

\section{Computational Cost of Autolabeling and OFDS} \label{sec:app_cost}

\begin{figure}
     \centering
    \includegraphics[width=.4\columnwidth]{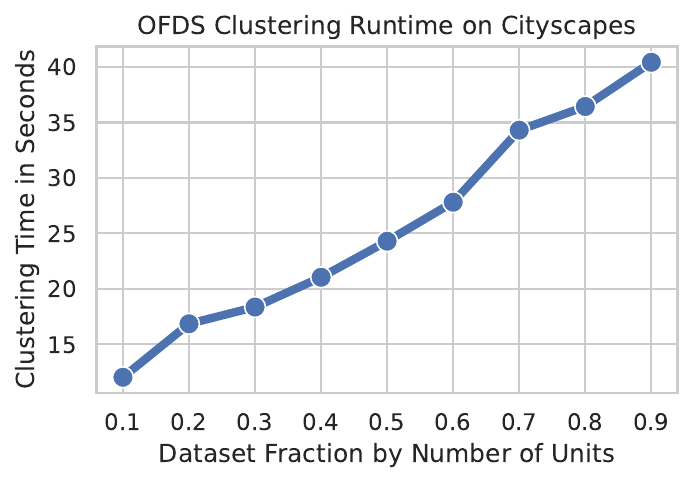}
    \caption{\textbf{Compute Time for the Data Selection in OFDS Without the Time to Generate the Object Proposals} The clustering and selection was performed on a Xeon Gold 6150 CPU.}
    \label{fig:timing}
\end{figure}

We use a modified pipeline of Grounding SAM 2 to compute the object proposals using a NVIDIA v100 GPU. We selected the DINO-T and SAM2-T models to achieve the highest throughput. In this setup, the throughput is $4.2$ images per second on average.  The clustering and selection process of OFDS takes only in the order of seconds on a Xeon Gold 6150 CPU. The times for the clustering steps on the Cityscapes dataset are shown in Figure \ref{fig:timing}. We emphasize that generating autolabels or object proposals and the selection process are thereby substantially less computationally expensive than the model trainings carried out in Section \ref{sec:experiments}. Each training took between 6 hours for object detection on PASCAL VOC to 17 and 21 hours for semantic segmentation on Cityscapes and PASCAL VOC using the same compute hardware as for the data selection. The trainings are performed using the mmdetection \cite{mmdetection} and mmsegmentation \cite{mmsegmentation} frameworks. In contrast, performing OFDS on the PASCAL VOC dataset took 50 minutes for object detection and 19 minutes for semantic segmentation including the generation of the object features.  \\
In the context of active learning, training a model for object detection on the PASCAL VOC CI dataset with PPAL took 12.3 GPU hours. When using OFDS to perform the initial data selection instead of random, this adds a computational overhead of 6.5$\%$. The active learning runs for semantic segmentation on the PASCAL VOC CI dataset took 19.1 GPU hours on average such that the computational overhead by using OFDS is only 1.7$\%$.


\section{Combining Dataset Selection with Autolabels} \label{sec:app_finetuning}

\subsection{Combining Autolabels with OFDS on Cityscapes}
\begin{figure}
     \centering
    \includegraphics[width=.75\columnwidth]{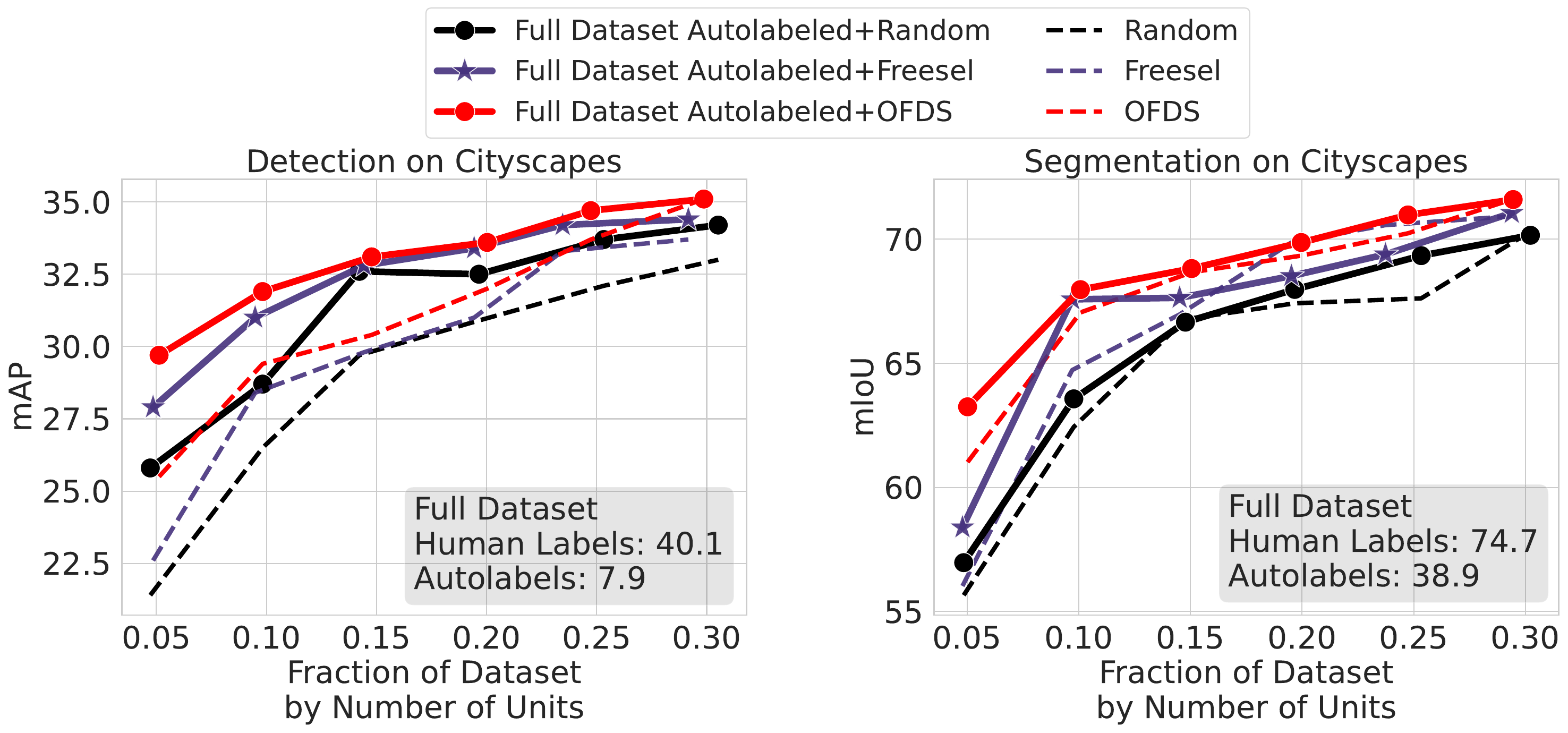}
    \caption{\textbf{Combining Autolabels With Data Selection on Cityscapes.} The training setup is the same as for Figure \ref{fig:dect_ft} but with hyperparameters adjusted for the Cityscapes dataset as stated in Section \ref{sec:app_hyperparameter}. The results validate our observation that fine-tuning with human-labeled improves the performance over training purely with autolabels and selecting the subset for labeling with OFDS yields the best performances.}
    \label{fig:finetuning_cityscapes}
    \vspace{-.2cm}
\end{figure}

To complement Figure \ref{fig:dect_ft}, we report the results of the models pre-trained with autolabels and then fine-tuned on selected subsets with human labels on the Cityscapes dataset in Figure \ref{fig:finetuning_cityscapes}. As for PASCAL VOC, we observe that fine-tuning with human-labeled images improves the performance over training purely with autolabels. The improvements achieved through pre-training on autolabels are smaller in comparison on PASCAL VOC as a results of the weaker performance of the models trained with autolabels.

\subsection{Complete Results on PASCAL VOC with Class Imbalance and Cityscapes}
\begin{figure*}
     \centering
    \includegraphics[width=.8\columnwidth]{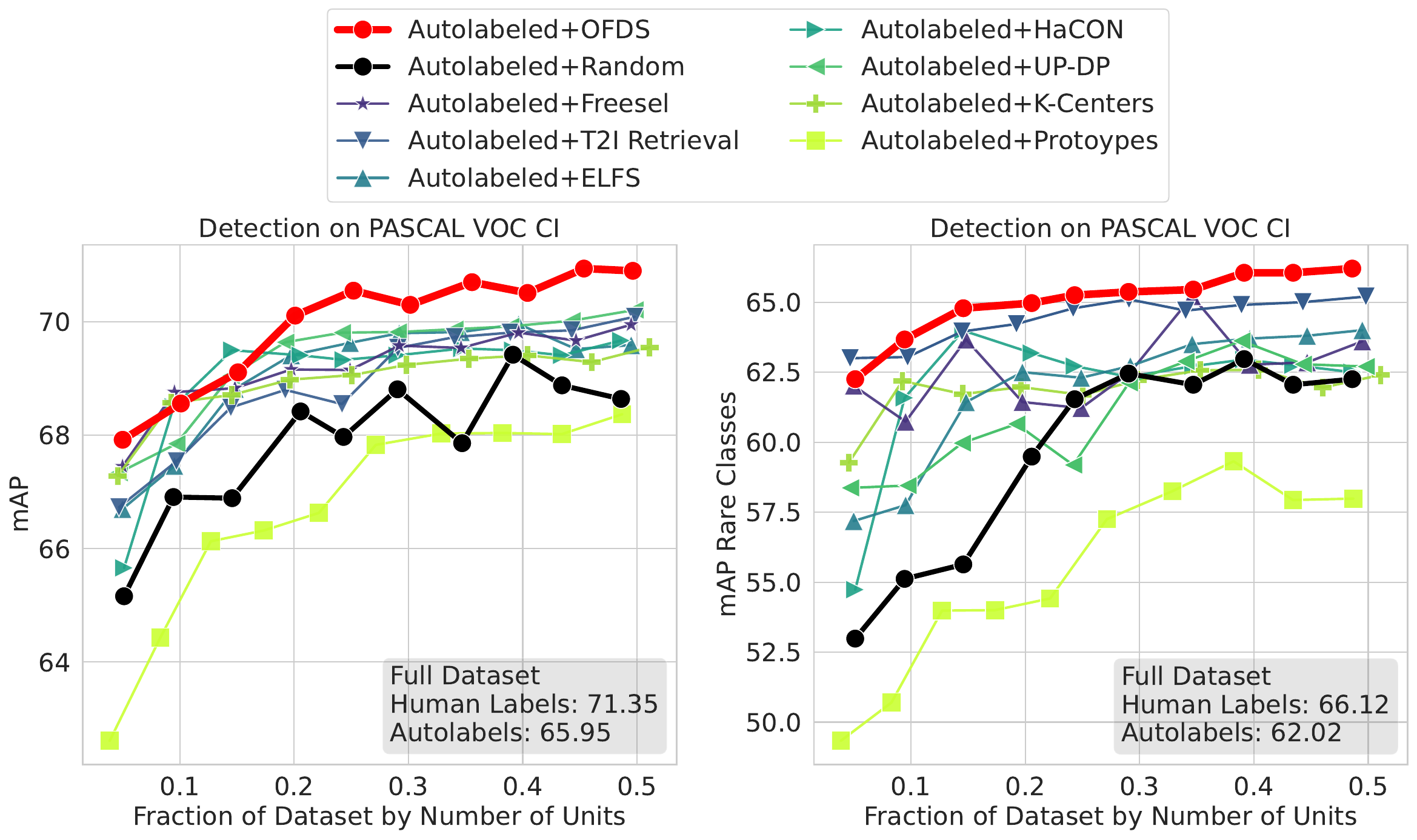}
    \includegraphics[width=.8\columnwidth]{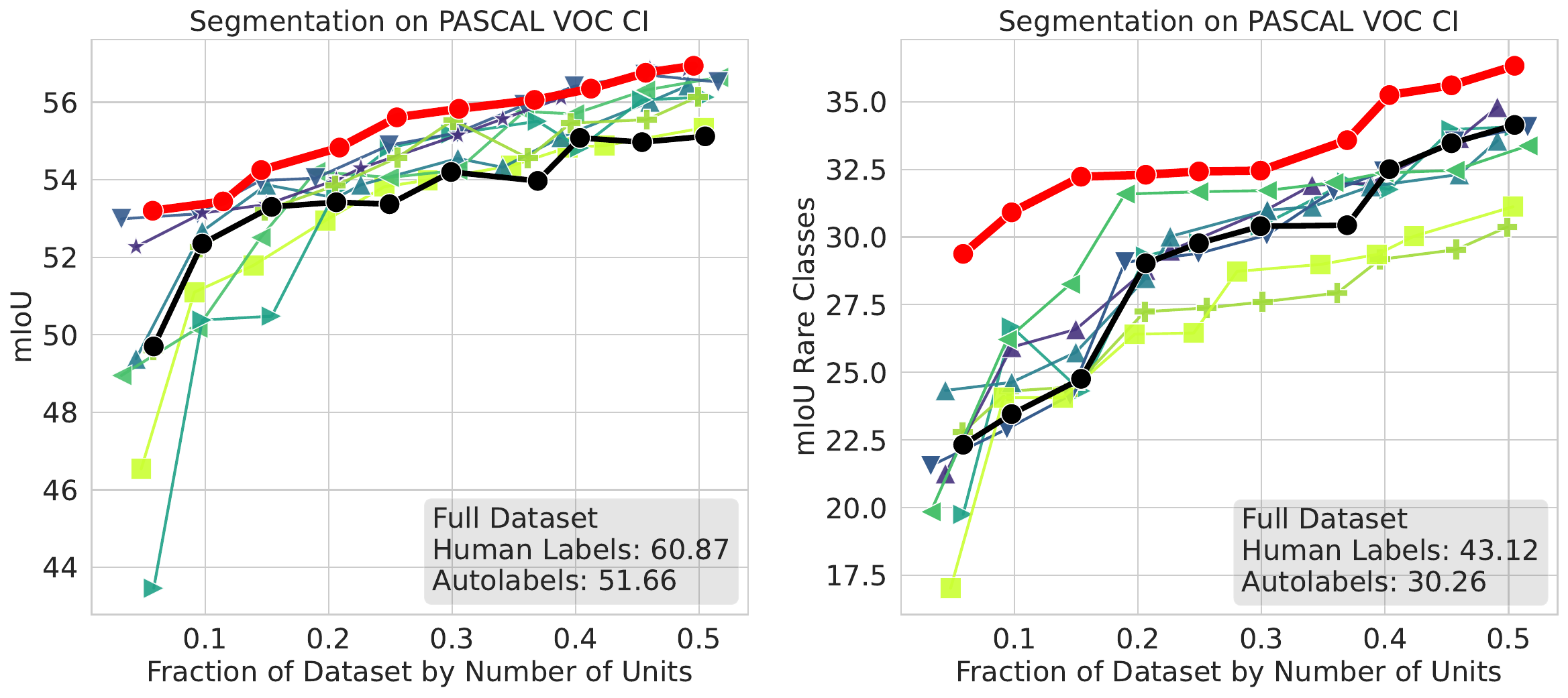}
    \caption{\textbf{Complete Results for Combining Autolabels with Data Selection on PASCAL VOC with Class Imbalance.}  The results correspond to a FasterRCNN with ResNet-18 backbone and a Segmenter with ViT-T backbone. The models were first pre-trained on the full dataset with autolabels and then fine-tuned on selected subsets with human labels. These subsets were selected by the six baselines or OFDS given the fixed annotation budgets indicated on the x-axis. }
    \label{fig:app_finetuning_voc_full}
    \vspace*{-0.4cm}
\end{figure*}

\begin{figure*}
     \centering
    \includegraphics[width=.8\columnwidth]{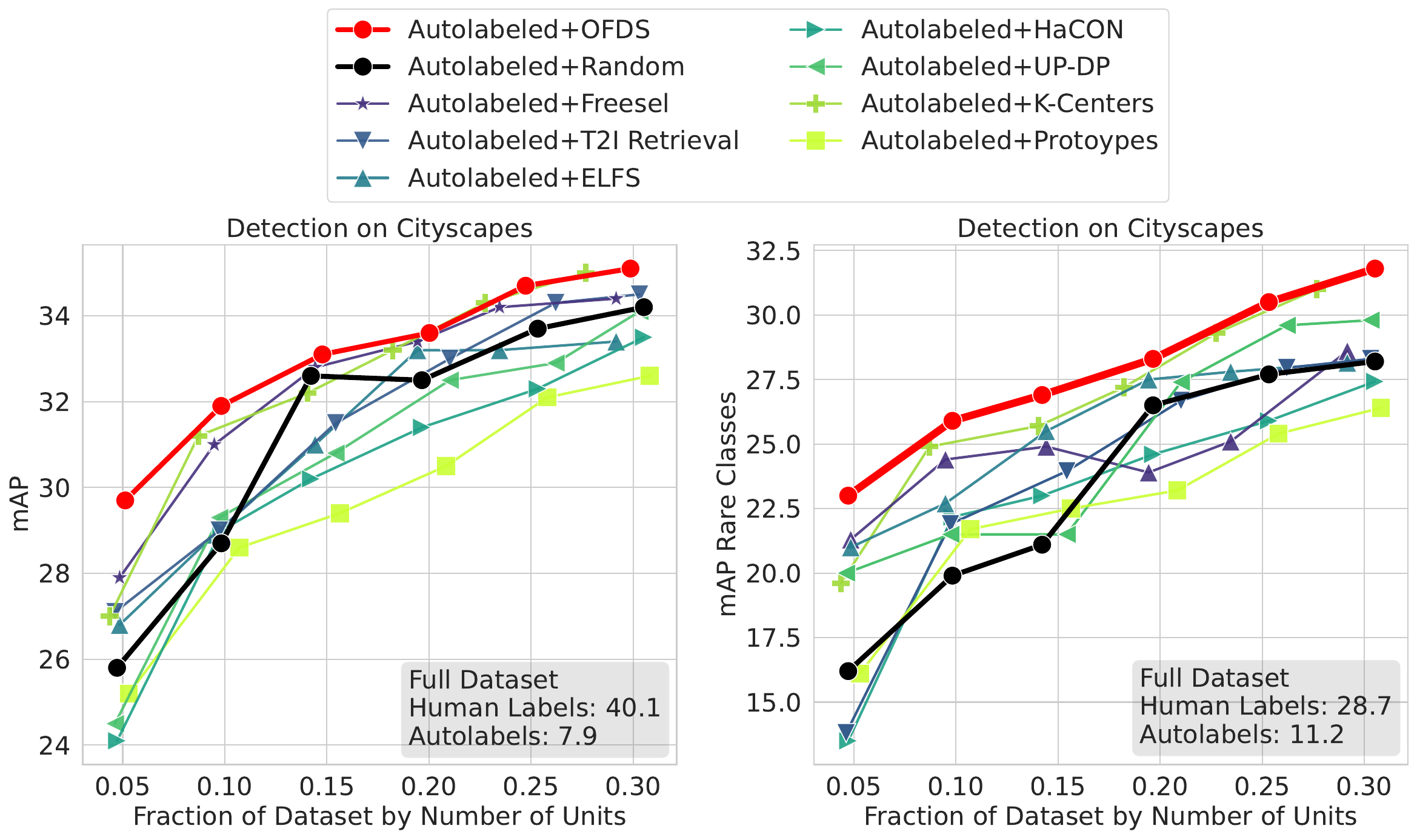}
    \includegraphics[width=.8\columnwidth]{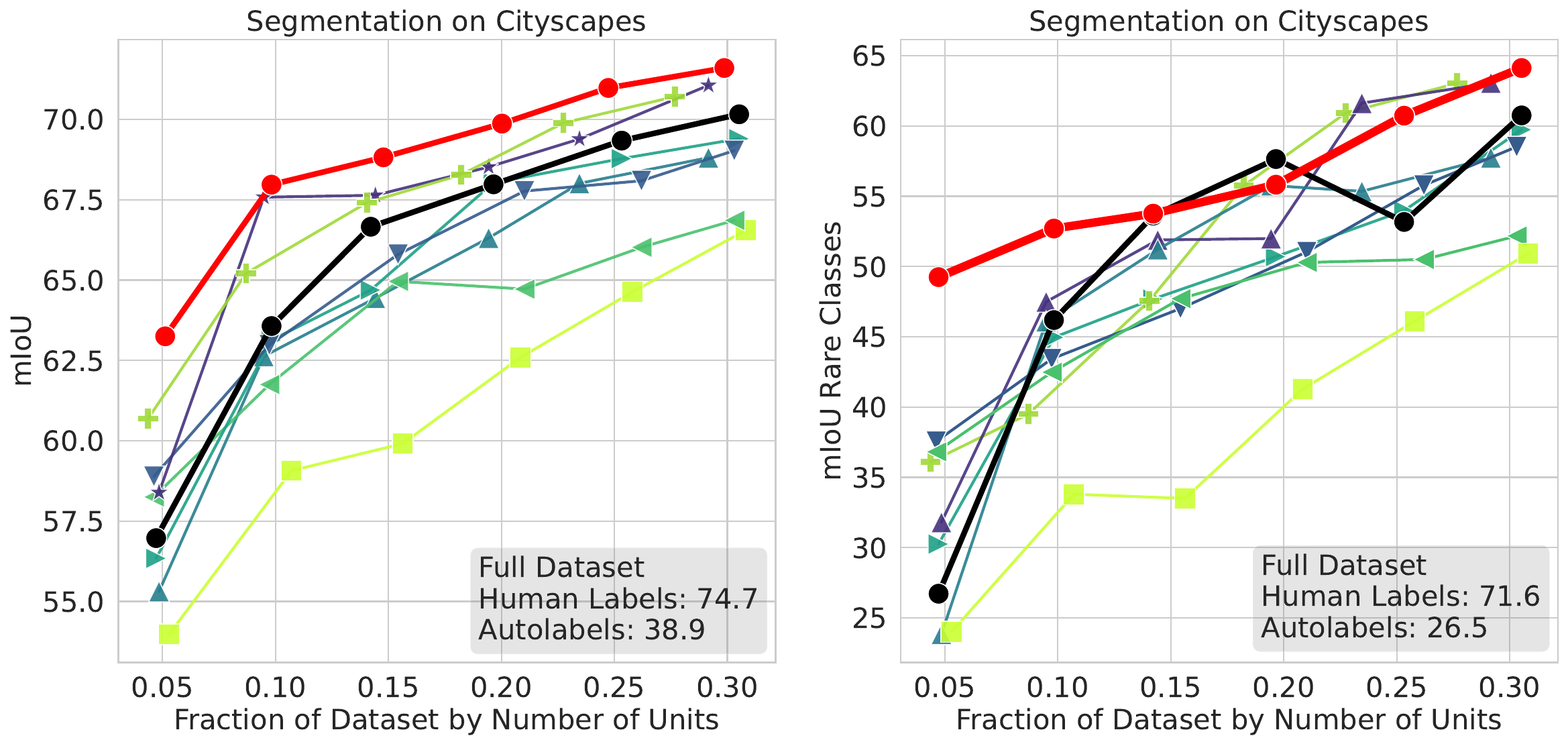}
    \caption{\textbf{Complete Results for Combining Autolabels with Data Selection on Cityscapes.}  The results were obtained using the same setup as for Figures \ref{fig:dect_ft} and \ref{fig:app_finetuning_voc_full}.}
    \label{fig:app_finetuning_cs_full}
\end{figure*}

The complete results including all six baselines for fine-tuning on selected datasets with human labels are shown in Figures \ref{fig:app_finetuning_voc_full} and \ref{fig:app_finetuning_cs_full}. As in Section \ref{sec:finetuning_results}, the pre-trained checkpoints are obtained through training with autolabels on the full datasets. The results confirm that selecting the data for fine-tuning through OFDS consistently leads to the best performances and yields improvements over all baselines.

\section{Results on Instance Segmentation }\label{sec:app_instance} 
\begin{figure}
     \centering
    \includegraphics[width=.45\columnwidth]{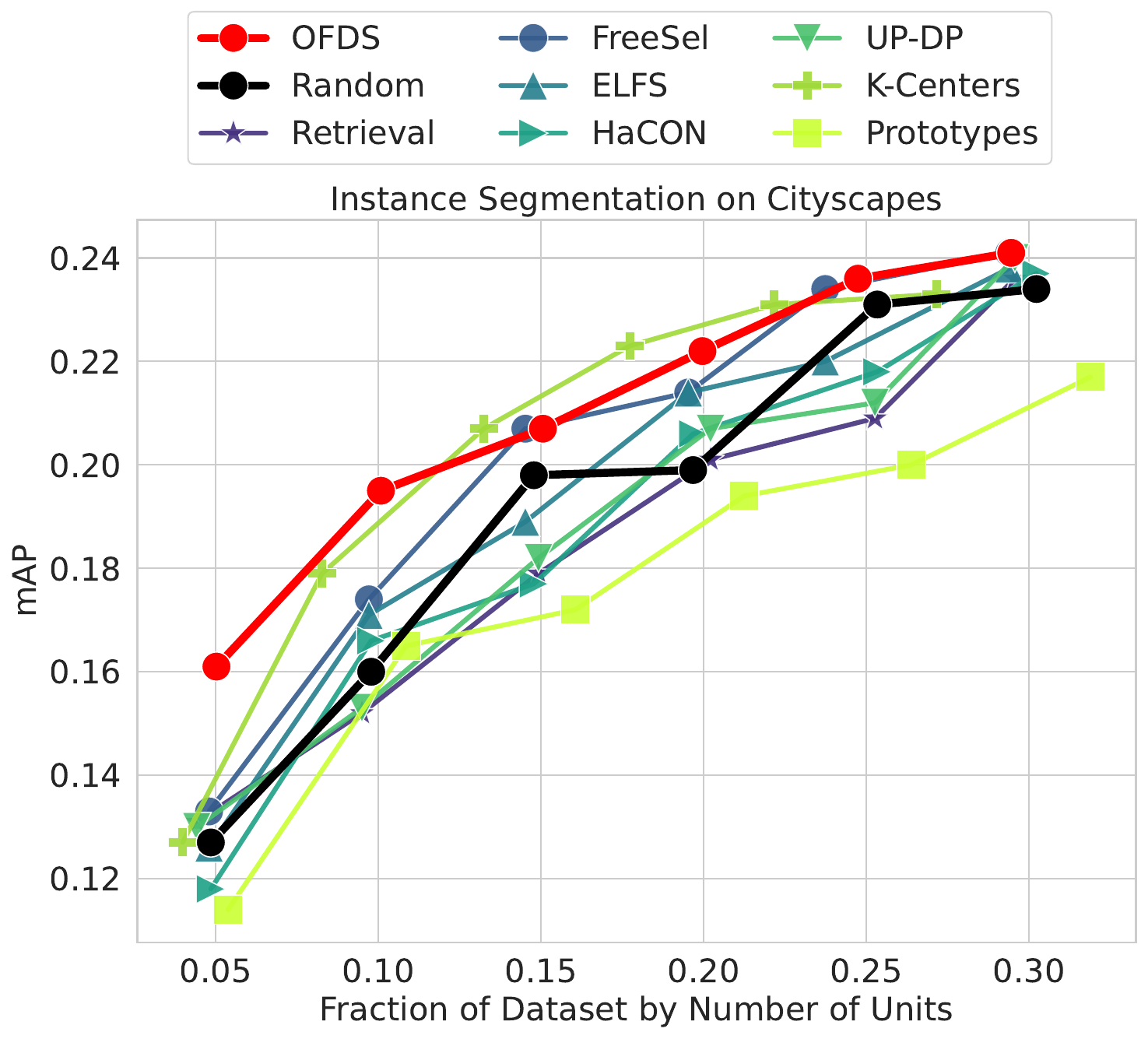}
    \caption{\textbf{Results for Instance Segmentation on Cityscapes.} The downstream models trained for the tasks is a MaskRCNN with ResNet-18 backbone. OFDS consistently outperforms all baselines also for this task.}
    \label{fig:instance}
\end{figure}

In Figure \ref{fig:instance}, we show the results when training a MaskRCNN model with ResNet-18 backbones for instance segmentation on the selected subsets of Cityscapes. We observe that also for this task, OFDS consistently outperforms all baselines.

\section{Summary of Baselines} \label{sec:baselines_summary}

In this section, we provide a more detailed summary of all baseline methods which we compare OFDS to:

\underline{FreeSel:} Xie et al. \cite{freesel} introduced FreeSel as a method for data selection based on a single pass of the unlabeled dataset through DINO. The selection is based on local semantic features of images. Contrary to our approach, the number of features per image is a fixed hyperparameter independent of the number of objects on a specific image.\\

\underline{UP-DP:} UP-DP \cite{updp} performs data selection based on  unsupervised prompt learning using vision-language models, in particular BLIP-2 \cite{blip2}. Contrary to our method, UP-DP requires training. We compare to the UP-DP selection based on probabilities predicted by cluster-level head which exhibited the best performance in the original publication.\\

\underline{Prototypes:} The current state-of-the art approach for unsupervised data selection for image classification was presented by Sorscher et al. \cite{beyond} for ImageNet \cite{imagenet} and has since been scaled to webdatasets \cite{laion_pruning}. The method consists of two steps. First, extract image features from an pre-trained model and perform $k$-means clustering using these features where $k$ equals the number of classes. Subsequently, the datapoints closest to the cluster centers are selected. We compare to this protoype selection using image features from a DINO-T \cite{dino} model.\\

\underline{K-Centers:} By the K-Centers algorithm, we refer to greed K-Centers by Sener et al. \cite{active_learning_coreset}. The selection requires image features and starts with a random image as initial subset. The images with the greatest distance to the subset are then incrementally added to the subset. We compare to K-Centers selection using features from DINO-T and the $\mathcal{L}_2$ in the feature space to determine the newly added points.\\%

\underline{HaCON:} Chen et al. \cite{hacon} developed HaCON to address the cold start problem for active learning. Yet, the method can be applied to general data selection. HaCON clusters the features of a self-supervised model and selects the samples which are located at the cluster boundaries.\newpage

\underline{ELFS} Zheng et al. \cite{elfs} introduced ELFS as a coreset method that uses pseudolabels instead of supervised labels. The method uses features from DINO to perform deep clustering. The resulting cluster labels are employed to compute diffuculty scores which are the basis for the selection. \\

\underline{Text-to-image retrieval} Selecting subsets through text-to-image retrieval requires a pretrained vision-language model for which we use a SigLip 2 ViT-B/16 model with the highest available resolution \cite{siglip2}. For every class, the top-k images with the highest similarity to the prompt ``a photo of a $\{$classname$\}$'' are selected. k is selected such that the available annotation budget is split evenly between all classes.\\

\underline{Random Selection} The simplest baseline is the selection of a random subset from the unlabeled images.

\section{Fine-Tuning Grounding DINO} \label{sec:grounding_dino}
The main purpose of OFDS is to perform single-pass data selection for training compact downstream models for specific target tasks. As an ablation, we fine-tune the GroundingDINO-T model on selected subsets of the PASCAL VOC dataset with class imbalance. We train for 80000 steps with a base learning rate of $10^{-4}$ and cosine learning rate scheduling using the AdamW \cite{adamw} optimizer. In Table \ref{tab:res:finetuning_gd}, we compare OFDS to FreeSel and random as the best baselines from Section \ref{sec:experiments}. Also in this setting, OFDS consistently yields the best results. However, we highlight that this is not its target application. For OFDS, we specifically target the training of small downstream models (which can run in resource-constrained environments) for dedicated target tasks instead of open-world foundation models.

\begin{table}
    \centering
    \resizebox{.4\columnwidth}{!}{
    \begin{tabular}{|l|c c c c |}
        \hline
        & \multicolumn{4}{c|}{Target Subset Size} \\
        Method & 5\% & 15\% & 25\% & 35\% \\
        \hline
        Random & 80.8 & 81.6 & 82.7 & 83.5 \\
        \textcolor{lightgray}{Subset Size}& \textcolor{lightgray}{5.15\%} & \textcolor{lightgray}{14.58\%} & \textcolor{lightgray}{24.32\%} & \textcolor{lightgray}{34.70\%} \\
        FreeSel & 83.6 & 84.6 & 84.9 & 85.1\\
        \textcolor{lightgray}{Subset Size}& \textcolor{lightgray}{4.99\%} & \textcolor{lightgray}{14.82\%} & \textcolor{lightgray}{24.89\%} & \textcolor{lightgray}{34.60\%} \\
        OFDS & \textbf{84.9} & \textbf{85.8} & \textbf{86.3} & \textbf{86.9} \\ 
        \textcolor{lightgray}{Subset Size}& \textcolor{lightgray}{5.00\%} & \textcolor{lightgray}{15.09\%} & \textcolor{lightgray}{25.19\%} & \textcolor{lightgray}{34.75\%} \\\hline
    \end{tabular}%
    }
    \vspace{0.2cm}
    \caption{\textbf{Fine-Tuning Grounding DINO-T for Object Detection (mAP) on PASCAL VOC with Class Imbalance.} We observe that OFDS consistently outperforms random selection and FreeSel. However, we highlight that this is not the target application of OFDS as it is designed to select data for training compact models for specific downstream tasks not explicitly for fine-tuning open-world foundation models.}
    \vspace{-0.2cm}
    \label{tab:res:finetuning_gd}
\end{table}

\section{Training Hyperparameters} \label{sec:app_hyperparameter}
The hyperparameter configurations for all models trained in this work can be found in Table \ref{tab:det_hyperparam} for object detection and in Table \ref{tab:seg_hyperparam} for semantic segmentation respectively. Due to the smaller resolution and subset size for the PASCAL VOC segmentation split, we trained with higher weight decay and learning rate in comparison to Cityscapes. Loss functions including loss weights are taken as in the original works that presented the model architectures.\\
\begin{table*}[h]
\centering
\begin{tabular}{l|c|c}
Hyperparameters & Faster-RCNN &  Deformable DETR \\ \midrule
Backbone & ResNet-18 & ResNet-18  \\
Optimizer & AdamW \cite{adamw} & AdamW \cite{adamw} \\
Optimizer Parameters & $\epsilon =$10e-8, $\beta \in$(0.9, 0.999) & $\epsilon =$10e-8, $\beta \in$(0.9, 0.999)   \\
Base lr & 1e-4 &  1e-4(VOC), backbone scaled by factor 0.1 \\
Weight decay & 5e-4, 3e-1 (VOC Fine-Tuning) &  1e-4 (VOC)  \\
Optimizer Steps & 80k, 60k (Fine-Tuning)  & 80k  \\
Batchsize & 8 & 8 \\
Lr schedule & Cosine Annealing & Cosine Annealing \\
Warmup steps & 1k & 1k \\
Warmup configuration & Linear Warmup, Factor 0.1 &  Linear Warmup, Factor 0.1  \\
Augmentations & PhotoMetric Distortion, Random Crop, & PhotoMetric Distortion, Random Crop,  \\
& Random Flip & Random Flip
\end{tabular}
\caption{\textbf{Hyperparameters for Object Detection.} We provide the hyperparameter configurations for all object detection models trained in this work. The specified values correspond to the setup described in Section \ref{sec:data_selection_results}. Unless explicitly stated otherwise, the same configurations are used for fine-tuning in Section \ref{sec:finetuning_results}.}
\label{tab:det_hyperparam}
\end{table*}

\begin{table*}[h]
\centering
\begin{tabular}{l|c|c}
Hyperparameters & Segmenter &  PSPNet \\ \midrule
Backbone & ViT-T & ResNet-18  \\
Optimizer & AdamW \cite{adamw} & SGD \\
Optimizer Parameters & $\epsilon =$10e-8, $\beta \in$(0.9, 0.999) & momentum 0.9\\
Base lr & 1e-5 (VOC, LoveDA),  1e-4 (Cityscapes) & 1e-2 (VOC) \\
& 1e-6 (VOC Fine-Tuning) & \\
Weight decay & 1e-2 (VOC, LoveDA), 5e-4 (Cityscapes) &  5e-4 (VOC)  \\ 
 & 1e-1 (VOC Fine-Tuning) & \\ 
Optimizer Steps & 80k, 60k (Fine-Tuning)  & 80k \\
Batchsize & 4 & 4 \\
Lr schedule & Cosine Annealing & Cosine Annealing \\
Warmup steps & 1k & 0 \\
Warmup configuration & Linear Warmup, Factor 0.1 &    \\
Augmentations & PhotoMetric Distortion, Random Crop, & PhotoMetric Distortion, Random Crop,  \\
& Random Flip & Random Flip
\end{tabular}
\caption{\textbf{Hyperparameters for Semantic Segmentation.} For all models trained for semantic segmentation in this work, we list the hyperparameters configurations. The stated values refer to the setup for Section \ref{sec:data_selection_results}. When not explicitly stated otherwise the configurations used for fine-tuning in Section \ref{sec:finetuning_results} are the same.}
\label{tab:seg_hyperparam}
\end{table*}

\section{Full numerical results} \label{sec:numerical_results} Tables \ref{tab:res_voc_det}, \ref{tab:res_voc_ic_det}, \ref{tab:res_voc_seg}, \ref{tab:res_voc_ic_seg}, \ref{tab:res_cs_det}, \ref{tab:res_cs_seg} and \ref{tab:res_loveda_seg} report the full numerical results for our main experiments.

\begin{table}
    \centering
    \resizebox{.9\columnwidth}{!}{
    \begin{tabular}{|l|c c c c c c c c c|}
        \hline
        & \multicolumn{9}{c|}{Target Subset Size} \\
        Method & 10\% & 20\% & 30\% & 40\% & 50\% & 60\% & 70\% & 80\% & 90\% \\
        \hline
        Random & 61.50 & 67.86 & 70.91 & 72.75 & 74.09 & 75.60 & 76.53 & \textbf{77.37} & 77.23 \\
        \textcolor{lightgray}{Subset size} & \textcolor{lightgray}{10.18\%} & \textcolor{lightgray}{19.62\%} & \textcolor{lightgray}{29.46\%} & \textcolor{lightgray}{39.82\%} & \textcolor{lightgray}{50.26\%} & \textcolor{lightgray}{60.35\%} & \textcolor{lightgray}{70.38\%} & \textcolor{lightgray}{79.62\%} & \textcolor{lightgray}{90.19\%} \\
        FreeSel & 62.56 & 67.93 & 70.88 & 72.94 & 74.37 & 76.08 & 76.01 & 77.24 & 77.47 \\
        \textcolor{lightgray}{Subset size}& \textcolor{lightgray}{9.83\%} & \textcolor{lightgray}{19.99\%} & \textcolor{lightgray}{29.43\%} & \textcolor{lightgray}{39.28\%} & \textcolor{lightgray}{49.71\%} & \textcolor{lightgray}{59.88\%} & \textcolor{lightgray}{69.91\%} & \textcolor{lightgray}{80.18\%} & \textcolor{lightgray}{90.01\%} \\
        K-Centers & 53.11 & 63.16 & 68.01 & 71.18 & 73.09 & 74.36 & 76.02 & 76.82 & 77.66 \\
        \textcolor{lightgray}{Subset size}& \textcolor{lightgray}{8.68\%} & \textcolor{lightgray}{18.70\%} & \textcolor{lightgray}{29.28\%} & \textcolor{lightgray}{40.05\%} & \textcolor{lightgray}{50.51\%} & \textcolor{lightgray}{60.60\%} & \textcolor{lightgray}{71.45\%} & \textcolor{lightgray}{81.84\%} & \textcolor{lightgray}{91.59\%} \\
        K-Means & 47.41 & 55.95 & 63.69 & 68.03 & 70.86 & 73.52 & 74.71 & 76.07 & 77.23 \\
        \textcolor{lightgray}{Subset size}& \textcolor{lightgray}{9.41\%} & \textcolor{lightgray}{18.91\%} & \textcolor{lightgray}{29.39\%} & \textcolor{lightgray}{40.20\%} & \textcolor{lightgray}{51.20\%} & \textcolor{lightgray}{61.88\%} & \textcolor{lightgray}{72.38\%} & \textcolor{lightgray}{82.31\%} & \textcolor{lightgray}{92.22\%} \\
        UP-DP & 54.15 & 62.62 & 68.09 & 71.38 & 73.01 & 74.76 & 75.45 & 76.12 & 76.76 \\
        \textcolor{lightgray}{Subset size}& \textcolor{lightgray}{10.80\%} & \textcolor{lightgray}{20.68\%} & \textcolor{lightgray}{31.48\%} & \textcolor{lightgray}{41.44\%} & \textcolor{lightgray}{51.00\%} & \textcolor{lightgray}{61.28\%} & \textcolor{lightgray}{71.52\%} & \textcolor{lightgray}{81.48\%} & \textcolor{lightgray}{91.20\%} \\
        HACON & 61.41 & 67.81 & 70.20 & 72.73 & 74.43 & 75.71 & 76.21 & 76.71 & 77.17 \\
        \textcolor{lightgray}{Subset size}& \textcolor{lightgray}{10.21\%} & \textcolor{lightgray}{19.57\%} & \textcolor{lightgray}{29.44\%} & \textcolor{lightgray}{39.78\%} & \textcolor{lightgray}{50.23\%} & \textcolor{lightgray}{60.35\%} & \textcolor{lightgray}{70.32\%} & \textcolor{lightgray}{79.55\%} & \textcolor{lightgray}{90.01\%} \\
        T2I Retrieval & 60.01 & 66.21 & 69.20 & 71.72 & 73.93 & 74.61 & 77.01 & 77.12 & 77.07 \\
        \textcolor{lightgray}{Subset size}& \textcolor{lightgray}{10.21\%} & \textcolor{lightgray}{19.58\%} & \textcolor{lightgray}{29.49\%} & \textcolor{lightgray}{39.67\%} & \textcolor{lightgray}{50.25\%} & \textcolor{lightgray}{60.32\%} & \textcolor{lightgray}{70.33\%} & \textcolor{lightgray}{79.62\%} & \textcolor{lightgray}{90.12\%} \\
        ELFS & 62.70 & 67.89 & 69.97 & 72.57 & 74.10 & 75.24 & 76.18 & 77.32 & 77.62 \\
        \textcolor{lightgray}{Subset size}& \textcolor{lightgray}{10.15\%} & \textcolor{lightgray}{19.68\%} & \textcolor{lightgray}{29.48\%} & \textcolor{lightgray}{39.80\%} & \textcolor{lightgray}{50.31\%} & \textcolor{lightgray}{60.19\%} & \textcolor{lightgray}{70.32\%} & \textcolor{lightgray}{79.66\%} & \textcolor{lightgray}{90.19\%} \\
        OFDS & \textbf{63.59} & \textbf{68.37} & \textbf{71.15} & \textbf{73.55} & \textbf{74.59} & \textbf{75.91} & \textbf{76.76} & 77.20 & \textbf{77.40} \\
        \textcolor{lightgray}{Subset size}& \textcolor{lightgray}{10.62\%} & \textcolor{lightgray}{21.24\%} & \textcolor{lightgray}{31.64\%} & \textcolor{lightgray}{41.88\%} & \textcolor{lightgray}{51.25\%} & \textcolor{lightgray}{60.25\%} & \textcolor{lightgray}{69.77\%} & \textcolor{lightgray}{85.52\%} & \textcolor{lightgray}{89.92\%} \\
        \hline
    \end{tabular}%
    }
    \vspace{0.2cm}
    \caption{\textbf{Object Detection Results (mAP) on PASCAL VOC}}
    \label{tab:res_voc_det}
\end{table}

\begin{table}
    \centering
    \resizebox{.9\columnwidth}{!}{
    \begin{tabular}{|l|c c c c c c c c c c|}
        \hline
        & \multicolumn{10}{c|}{Target Subset Size} \\
        Method & 5\% & 10\% & 15\% & 20\% & 25\% & 30\% & 35\% & 40\% & 45\% & 50\% \\
        \hline
        Random & 41.63 & 48.15 & 50.52 & 56.17 & 56.79 & 59.16 & 60.43 & 62.46 & 63.38 & 63.66 \\
        \textcolor{lightgray}{Subset size} & \textcolor{lightgray}{5.12\%} & \textcolor{lightgray}{9.44\%} & \textcolor{lightgray}{14.58\%} & \textcolor{lightgray}{20.56\%} & \textcolor{lightgray}{24.32\%} & \textcolor{lightgray}{29.04\%} & \textcolor{lightgray}{34.70\%} & \textcolor{lightgray}{39.14\%} & \textcolor{lightgray}{43.43\%} & \textcolor{lightgray}{48.60\%} \\
        FreeSel & 42.89 & 49.89 & 54.90 & 55.95 & 58.11 & 61.95 & 64.69 & 63.77 & 66.12 & 66.80 \\
        \textcolor{lightgray}{Subset size} & \textcolor{lightgray}{4.99\%} & \textcolor{lightgray}{9.53\%} & \textcolor{lightgray}{14.82\%} & \textcolor{lightgray}{19.74\%} & \textcolor{lightgray}{24.89\%} & \textcolor{lightgray}{29.19\%} & \textcolor{lightgray}{34.60\%} & \textcolor{lightgray}{39.64\%} & \textcolor{lightgray}{44.66\%} & \textcolor{lightgray}{49.48\%} \\
        K-Centers & 36.84 & 45.79 & 48.36 & 51.21 & 55.06 & 58.12  & 60.06 & 62.14 & 63.64 & 65.08 \\
        \textcolor{lightgray}{Subset size} & \textcolor{lightgray}{4.56\%} & \textcolor{lightgray}{9.25\%} & \textcolor{lightgray}{14.53\%} & \textcolor{lightgray}{19.62\%} & \textcolor{lightgray}{25.03\%} & \textcolor{lightgray}{29.83\%} & \textcolor{lightgray}{35.30\%} & \textcolor{lightgray}{40.42\%} & \textcolor{lightgray}{46.02\%} & \textcolor{lightgray}{51.09\%} \\
        Prototypes & 22.10 & 31.58 & 38.42 & 43.11 & 46.42 & 50.38 & 53.40 & 55.15 & 56.71 & 57.76 \\
        \textcolor{lightgray}{Subset size} & \textcolor{lightgray}{3.86\%} & \textcolor{lightgray}{8.29\%} & \textcolor{lightgray}{12.71\%} & \textcolor{lightgray}{17.35\%} & \textcolor{lightgray}{22.17\%} & \textcolor{lightgray}{27.14\%} & \textcolor{lightgray}{32.87\%} & \textcolor{lightgray}{38.22\%} & \textcolor{lightgray}{43.42\%} & \textcolor{lightgray}{48.70\%} \\
        UP-DP & 41.40 & 49.36 & 54.08 & 55.96 & 58.57 & 61.10 & 63.09 & 64.78 & 65.51 & 67.09 \\
        \textcolor{lightgray}{Subset size} & \textcolor{lightgray}{4.68\%} & \textcolor{lightgray}{9.71\%} & \textcolor{lightgray}{14.46\%} & \textcolor{lightgray}{19.23\%} & \textcolor{lightgray}{24.19\%} & \textcolor{lightgray}{29.07\%} & \textcolor{lightgray}{34.03\%} & \textcolor{lightgray}{38.91\%} & \textcolor{lightgray}{44.28\%} & \textcolor{lightgray}{49.79\%} \\
        HACON & 41.60 & 47.50 & 53.40 & 55.10 & 58.10 & 60.10 & 62.00 & 63.00 & 65.00 & 67.00 \\
        \textcolor{lightgray}{Subset size} & \textcolor{lightgray}{5.06\%} & \textcolor{lightgray}{9.49\%} & \textcolor{lightgray}{14.55\%} & \textcolor{lightgray}{20.51\%} & \textcolor{lightgray}{24.38\%} & \textcolor{lightgray}{29.01\%} & \textcolor{lightgray}{34.72\%} & \textcolor{lightgray}{39.13\%} & \textcolor{lightgray}{43.43\%} & \textcolor{lightgray}{48.65\%} \\
        T2I Retrieval & 44.70 & 51.00 & 55.40 & 60.00 & 61.80 & 64.10 & 64.20 & 65.60 & 66.90 & 67.90 \\
        \textcolor{lightgray}{Subset size} & \textcolor{lightgray}{5.15\%} & \textcolor{lightgray}{9.50\%} & \textcolor{lightgray}{14.62\%} & \textcolor{lightgray}{20.51\%} & \textcolor{lightgray}{24.28\%} & \textcolor{lightgray}{29.11\%} & \textcolor{lightgray}{34.73\%} & \textcolor{lightgray}{39.09\%} & \textcolor{lightgray}{43.39\%} & \textcolor{lightgray}{48.68\%} \\
        ELFS & 42.10 & 49.49 & 53.10 & 56.58 & 58.02 & 60.18 & 61.83 & 64.60 & 64.38 & 65.15 \\
        \textcolor{lightgray}{Subset size} & \textcolor{lightgray}{5.11\%} & \textcolor{lightgray}{9.49\%} & \textcolor{lightgray}{14.52\%} & \textcolor{lightgray}{20.53\%} & \textcolor{lightgray}{24.34\%} & \textcolor{lightgray}{29.06\%} & \textcolor{lightgray}{34.75\%} & \textcolor{lightgray}{39.15\%} & \textcolor{lightgray}{43.39\%} & \textcolor{lightgray}{48.68\%} \\
        OFDS & \textbf{49.64} & \textbf{54.90} & \textbf{58.81} & \textbf{61.73} & \textbf{63.95} & \textbf{64.73} & \textbf{66.40} & \textbf{67.18} & \textbf{68.27} & \textbf{68.62} \\
        \textcolor{lightgray}{Subset size} & \textcolor{lightgray}{5.00\%} & \textcolor{lightgray}{10.09\%} & \textcolor{lightgray}{15.09\%} & \textcolor{lightgray}{20.09\%} & \textcolor{lightgray}{25.19\%} & \textcolor{lightgray}{30.16\%} & \textcolor{lightgray}{35.56\%} & \textcolor{lightgray}{40.41\%} & \textcolor{lightgray}{45.35\%} & \textcolor{lightgray}{49.63\%} \\
        \hline
    \end{tabular}%
    }
    \vspace{0.2cm}
    \caption{\textbf{Object Detection Results (mAP) on PASCAL VOC with Class Imbalance}}
    \label{tab:res_voc_ic_det}
\end{table}

\begin{table}
    \centering
    \resizebox{.9\columnwidth}{!}{
    \begin{tabular}{|l|c c c c c c c c c|}
        \hline
        & \multicolumn{9}{c|}{Target Subset Size} \\
        Method & 10\% & 20\% & 30\% & 40\% & 50\% & 60\% & 70\% & 80\% & 90\% \\
        \hline
        Random & \textbf{49.60} & 53.75 & 58.05 & 61.93 & 64.01 & 65.95 & \textbf{68.18} & \textbf{69.33} & 69.97 \\
        \textcolor{lightgray}{Subset size} & \textcolor{lightgray}{8.52\%} & \textcolor{lightgray}{20.64\%} & \textcolor{lightgray}{30.19\%} & \textcolor{lightgray}{37.29\%} & \textcolor{lightgray}{48.52\%} & \textcolor{lightgray}{59.18\%} & \textcolor{lightgray}{71.95\%} & \textcolor{lightgray}{80.13\%} & \textcolor{lightgray}{90.11\%} \\
        FreeSel & 40.33 & \textbf{59.20} & 62.34 & \textbf{62.39} & 61.94 & 63.76 & 69.23 & 68.77 & 69.63 \\
        \textcolor{lightgray}{Subset size} & \textcolor{lightgray}{10.32\%} & \textcolor{lightgray}{18.93\%} & \textcolor{lightgray}{30.96\%} & \textcolor{lightgray}{39.82\%} & \textcolor{lightgray}{46.75\%} & \textcolor{lightgray}{60.21\%} & \textcolor{lightgray}{70.24\%} & \textcolor{lightgray}{79.36\%} & \textcolor{lightgray}{89.71\%} \\
        K-Centers & 38.73 & 53.58 & 56.49 & 60.54 & 61.89 & 67.83 & 66.57 & 68.93 & 68.08 \\
        \textcolor{lightgray}{Subset size} & \textcolor{lightgray}{8.89\%} & \textcolor{lightgray}{19.10\%} & \textcolor{lightgray}{27.79\%} & \textcolor{lightgray}{38.26\%} & \textcolor{lightgray}{48.97\%} & \textcolor{lightgray}{58.01\%} & \textcolor{lightgray}{69.21\%} & \textcolor{lightgray}{81.13\%} & \textcolor{lightgray}{90.76\%} \\
        Prototypes & 26.42 & 35.36 & 49.42 & 58.87 & 60.78 & 62.87 & 66.56 & 68.26 & 68.57 \\
        \textcolor{lightgray}{Subset size} & \textcolor{lightgray}{10.26\%} & \textcolor{lightgray}{20.87\%} & \textcolor{lightgray}{31.33\%} & \textcolor{lightgray}{41.79\%} & \textcolor{lightgray}{52.02\%} & \textcolor{lightgray}{62.29\%} & \textcolor{lightgray}{72.46\%} & \textcolor{lightgray}{81.61\%} & \textcolor{lightgray}{91.99\%} \\
        UP-DP & 38.67 & 47.69 & 52.73 & 54.77 & 58.71 & 59.60 & 64.86 & 66.73 & 67.98 \\
        \textcolor{lightgray}{Subset size} & \textcolor{lightgray}{10.12\%} & \textcolor{lightgray}{20.87\%} & \textcolor{lightgray}{31.98\%} & \textcolor{lightgray}{42.93\%} & \textcolor{lightgray}{53.14\%} & \textcolor{lightgray}{62.46\%} & \textcolor{lightgray}{74.57\%} & \textcolor{lightgray}{84.92\%} & \textcolor{lightgray}{94.56\%} \\
        HACON & 48.52 & 54.00 & 58.50 & 58.10 & 60.40 & 62.80 & 66.10 & 67.00 & 67.30 \\
        \textcolor{lightgray}{Subset size} & \textcolor{lightgray}{8.55\%} & \textcolor{lightgray}{20.59\%} & \textcolor{lightgray}{30.21\%} & \textcolor{lightgray}{37.40\%} & \textcolor{lightgray}{48.59\%} & \textcolor{lightgray}{59.21\%} & \textcolor{lightgray}{71.88\%} & \textcolor{lightgray}{80.09\%} & \textcolor{lightgray}{90.02\%} \\
        T2I Retrieval & 40.67 & 49.36 & 56.16 & 60.81 & 65.26 & \textbf{67.76} & 67.94 & 69.15 & \textbf{70.21} \\
        \textcolor{lightgray}{Subset size} & \textcolor{lightgray}{8.49\%} & \textcolor{lightgray}{20.66\%} & \textcolor{lightgray}{30.17\%} & \textcolor{lightgray}{37.25\%} & \textcolor{lightgray}{48.55\%} & \textcolor{lightgray}{59.22\%} & \textcolor{lightgray}{71.90\%} & \textcolor{lightgray}{80.15\%} & \textcolor{lightgray}{90.20\%} \\
        ELFS & 41.26 & 56.75 & \textbf{62.42} & 61.92 & 64.47 & 65.44 & 66.43 & 68.86 & 69.78 \\
        \textcolor{lightgray}{Subset size} & \textcolor{lightgray}{8.58\%} & \textcolor{lightgray}{20.60\%} & \textcolor{lightgray}{30.24\%} & \textcolor{lightgray}{37.23\%} & \textcolor{lightgray}{48.47\%} & \textcolor{lightgray}{59.17\%} & \textcolor{lightgray}{71.98\%} & \textcolor{lightgray}{80.08\%} & \textcolor{lightgray}{90.15\%} \\
        OFDS & 48.52 & 54.25 & 59.96 & 61.89 & \textbf{64.65} & 66.79 & 67.82 & 68.68 & 69.65 \\
        \textcolor{lightgray}{Subset size} & \textcolor{lightgray}{8.41\%} & \textcolor{lightgray}{21.49\%} & \textcolor{lightgray}{31.78\%} & \textcolor{lightgray}{40.56\%} & \textcolor{lightgray}{49.32\%} & \textcolor{lightgray}{58.72\%} & \textcolor{lightgray}{73.43\%} & \textcolor{lightgray}{85.03\%} & \textcolor{lightgray}{89.60\%} \\
        \hline
    \end{tabular}%
    }
    \vspace{0.2cm}
    \caption{\textbf{Semantic Segmentation Results (mIoU) on PASCAL VOC}}
    \label{tab:res_voc_seg}
\end{table}

\begin{table}
    \centering
    \resizebox{.9\columnwidth}{!}{
    \begin{tabular}{|l|c c c c c c c c c c|}
        \hline
        & \multicolumn{10}{c|}{Target Subset Size} \\
        Method & 5\% & 10\% & 15\% & 20\% & 25\% & 30\% & 35\% & 40\% & 45\% & 50\% \\
        \hline
        Random & 24.53 & 32.58 & 37.26 & 44.02 & 44.81 & 49.59 & 50.44 & 51.10 & 53.86 & 56.06 \\
        \textcolor{lightgray}{Subset size} & \textcolor{lightgray}{5.83\%} & \textcolor{lightgray}{9.76\%} & \textcolor{lightgray}{15.38\%} & \textcolor{lightgray}{20.62\%} & \textcolor{lightgray}{24.93\%} & \textcolor{lightgray}{29.91\%} & \textcolor{lightgray}{36.92\%} & \textcolor{lightgray}{40.35\%} & \textcolor{lightgray}{45.37\%} & \textcolor{lightgray}{50.49\%} \\
        FreeSel & 25.04 & 37.53 & 38.23 & 47.28 & 47.55 & 50.54 & 49.04 & 50.00 & 54.93 & 50.37 \\
        \textcolor{lightgray}{Subset size} & \textcolor{lightgray}{4.39\%} & \textcolor{lightgray}{9.76\%} & \textcolor{lightgray}{14.96\%} & \textcolor{lightgray}{20.62\%} & \textcolor{lightgray}{22.60\%} & \textcolor{lightgray}{30.46\%} & \textcolor{lightgray}{34.09\%} & \textcolor{lightgray}{38.83\%} & \textcolor{lightgray}{46.01\%} & \textcolor{lightgray}{49.09\%} \\
        K-Centers & 19.29 & 29.10 & 39.97 & 40.39 & 42.64 & 48.17 & 49.01 & 51.46 & 53.66 & 52.63 \\
        \textcolor{lightgray}{Subset size} & \textcolor{lightgray}{5.79\%} & \textcolor{lightgray}{9.55\%} & \textcolor{lightgray}{14.83\%} & \textcolor{lightgray}{20.53\%} & \textcolor{lightgray}{25.56\%} & \textcolor{lightgray}{30.08\%} & \textcolor{lightgray}{36.12\%} & \textcolor{lightgray}{39.59\%} & \textcolor{lightgray}{45.75\%} & \textcolor{lightgray}{49.89\%} \\
        Prototypes & 28.83 & \textbf{39.40} & 38.77 & 39.61 & 44.74 & 48.37 & 51.61 & 53.34 & 54.97 & 54.46 \\
        \textcolor{lightgray}{Subset size} & \textcolor{lightgray}{4.82\%} & \textcolor{lightgray}{9.13\%} & \textcolor{lightgray}{13.90\%} & \textcolor{lightgray}{19.73\%} & \textcolor{lightgray}{24.50\%} & \textcolor{lightgray}{28.01\%} & \textcolor{lightgray}{34.77\%} & \textcolor{lightgray}{39.33\%} & \textcolor{lightgray}{42.33\%} & \textcolor{lightgray}{50.36\%} \\
        UP-DP & 18.60 & 38.26 & 37.71 & 44.77 & 47.44 & 50.45 & 48.01 & 48.83 & 52.01 & 54.82 \\
        \textcolor{lightgray}{Subset size} & \textcolor{lightgray}{3.21\%} & \textcolor{lightgray}{9.38\%} & \textcolor{lightgray}{14.49\%} & \textcolor{lightgray}{18.93\%} & \textcolor{lightgray}{24.88\%} & \textcolor{lightgray}{30.42\%} & \textcolor{lightgray}{35.83\%} & \textcolor{lightgray}{39.88\%} & \textcolor{lightgray}{45.63\%} & \textcolor{lightgray}{51.54\%} \\
        HACON & 21.02 & 34.30 & 37.10 & 44.90 & 46.50 & 49.50 & 50.90 & 53.90 & 54.70 & 56.10 \\
        \textcolor{lightgray}{Subset size} & \textcolor{lightgray}{5.79\%} & \textcolor{lightgray}{9.76\%} & \textcolor{lightgray}{15.36\%} & \textcolor{lightgray}{20.61\%} & \textcolor{lightgray}{24.93\%} & \textcolor{lightgray}{29.99\%} & \textcolor{lightgray}{36.90\%} & \textcolor{lightgray}{40.35\%} & \textcolor{lightgray}{45.34\%} & \textcolor{lightgray}{50.44\%} \\
        T2I Retrieval & 23.08 & 36.02 & 40.36 & 43.06 & 44.15 & 45.38 & 47.03 & 52.15 & 54.96 & 56.21 \\
        \textcolor{lightgray}{Subset size} & \textcolor{lightgray}{5.88\%} & \textcolor{lightgray}{9.82\%} & \textcolor{lightgray}{15.34\%} & \textcolor{lightgray}{20.67\%} & \textcolor{lightgray}{24.92\%} & \textcolor{lightgray}{29.94\%} & \textcolor{lightgray}{36.91\%} & \textcolor{lightgray}{40.39\%} & \textcolor{lightgray}{45.38\%} & \textcolor{lightgray}{50.51\%} \\
        ELFS & 23.62 & 32.72 & 39.19 & 42.58 & 45.58 & 47.79 & 48.31 & 52.22 & 54.03 & 55.99 \\
        \textcolor{lightgray}{Subset size} & \textcolor{lightgray}{5.77\%} & \textcolor{lightgray}{9.76\%} & \textcolor{lightgray}{15.34\%} & \textcolor{lightgray}{20.65\%} & \textcolor{lightgray}{24.89\%} & \textcolor{lightgray}{29.87\%} & \textcolor{lightgray}{36.99\%} & \textcolor{lightgray}{40.35\%} & \textcolor{lightgray}{45.36\%} & \textcolor{lightgray}{50.53\%} \\
        OFDS & \textbf{30.89} & 37.23 & \textbf{41.88} & \textbf{47.75} & \textbf{49.68} & \textbf{52.30} & \textbf{54.35} & \textbf{56.32} & \textbf{56.79} & \textbf{57.33} \\
        \textcolor{lightgray}{Subset size} & \textcolor{lightgray}{5.75\%} & \textcolor{lightgray}{11.45\%} & \textcolor{lightgray}{14.53\%} & \textcolor{lightgray}{20.87\%} & \textcolor{lightgray}{25.52\%} & \textcolor{lightgray}{30.54\%} & \textcolor{lightgray}{36.67\%} & \textcolor{lightgray}{41.23\%} & \textcolor{lightgray}{45.67\%} & \textcolor{lightgray}{49.56\%} \\
        \hline
    \end{tabular}%
    }
    \vspace{0.2cm}
    \caption{\textbf{Semantic Segmentation Results (mIoU) on PASCAL VOC with Class Imbalance}}
    \label{tab:res_voc_ic_seg}
\end{table}

\begin{table}
    \centering
    \resizebox{.6\columnwidth}{!}{
    \begin{tabular}{|l|c c c c c c|}
        \hline
        & \multicolumn{6}{c|}{Target Subset Size} \\
        Method & 5\% & 10\% & 15\% & 20\% & 25\% & 30\% \\
        \hline
        Random & 21.40 & 26.50 & 29.70 & 30.90 & 32.10 & 33.00 \\
        \textcolor{lightgray}{Subset size} & \textcolor{lightgray}{4.72\%} & \textcolor{lightgray}{9.83\%} & \textcolor{lightgray}{14.22\%} & \textcolor{lightgray}{19.65\%} & \textcolor{lightgray}{25.33\%} & \textcolor{lightgray}{30.52\%} \\
        FreeSel & 22.60 & 28.40 & 29.80 & 31.00 & 33.30 & 33.70 \\
        & \textcolor{lightgray}{4.85\%} & \textcolor{lightgray}{9.49\%} & \textcolor{lightgray}{14.42\%} & \textcolor{lightgray}{19.44\%} & \textcolor{lightgray}{23.46\%} & \textcolor{lightgray}{29.17\%} \\
        K-Centers & 22.40 & 28.20 & 30.20 & 31.30 & 33.30 & 34.70 \\
        \textcolor{lightgray}{Subset size} & \textcolor{lightgray}{4.36\%} & \textcolor{lightgray}{8.70\%} & \textcolor{lightgray}{14.05\%} & \textcolor{lightgray}{18.22\%} & \textcolor{lightgray}{22.74\%} & \textcolor{lightgray}{27.68\%} \\
        Prototypes & 22.00 & 25.70 & 27.10 & 28.50 & 29.70 & 30.60 \\
        \textcolor{lightgray}{Subset size} & \textcolor{lightgray}{5.30\%} & \textcolor{lightgray}{10.71\%} & \textcolor{lightgray}{15.63\%} & \textcolor{lightgray}{20.84\%} & \textcolor{lightgray}{25.80\%} & \textcolor{lightgray}{30.81\%} \\
        UP-DP & 22.40 & 25.00 & 29.50 & 30.80 & 32.30 & 33.30 \\
        \textcolor{lightgray}{Subset size} & \textcolor{lightgray}{4.62\%} & \textcolor{lightgray}{9.73\%} & \textcolor{lightgray}{15.43\%} & \textcolor{lightgray}{21.02\%} & \textcolor{lightgray}{26.20\%} & \textcolor{lightgray}{30.31\%} \\
        HACON & 20.90 & 26.40 & 28.70 & 30.40 & 31.50 & 32.90 \\
        \textcolor{lightgray}{Subset size}  & \textcolor{lightgray}{4.72\%} & \textcolor{lightgray}{9.81\%} & \textcolor{lightgray}{14.18\%} & \textcolor{lightgray}{19.55\%} & \textcolor{lightgray}{25.29\%} & \textcolor{lightgray}{30.48\%} \\
        T2I Retrieval & 22.00 & 26.00 & 28.20 & 30.90 & 31.40 & 32.00 \\
        \textcolor{lightgray}{Subset size} & \textcolor{lightgray}{4.70\%} & \textcolor{lightgray}{9.77\%} & \textcolor{lightgray}{14.17\%} & \textcolor{lightgray}{19.61\%} & \textcolor{lightgray}{25.23\%} & \textcolor{lightgray}{30.40\%} \\
        ELFS & 21.30 & 26.90 & 30.70 & 31.40 & 33.10 & 34.60 \\
        \textcolor{lightgray}{Subset size} & \textcolor{lightgray}{4.79\%} & \textcolor{lightgray}{9.88\%} & \textcolor{lightgray}{14.25\%} & \textcolor{lightgray}{19.68\%} & \textcolor{lightgray}{25.35\%} & \textcolor{lightgray}{30.47\%} \\
        OFDS & \textbf{25.50} & \textbf{29.40} & \textbf{30.40} & \textbf{32.00} & \textbf{33.70} & \textbf{35.10} \\
        \textcolor{lightgray}{Subset size} & \textcolor{lightgray}{5.12\%} & \textcolor{lightgray}{9.83\%} & \textcolor{lightgray}{14.78\%} & \textcolor{lightgray}{20.03\%} & \textcolor{lightgray}{24.73\%} & \textcolor{lightgray}{29.87\%} \\
        \hline
    \end{tabular}%
    }
    \vspace{0.2cm}
    \caption{\textbf{Object Detection Results (mAP) on Cityscapes}}
    \label{tab:res_cs_det}
\end{table}

\begin{table}
    \centering
    \resizebox{.6\columnwidth}{!}{
    \begin{tabular}{|l|c c c c c c|}
        \hline
        & \multicolumn{6}{c|}{Target Subset Size} \\
        Method & 5\% & 10\% & 15\% & 20\% & 25\% & 30\% \\
        \hline
        Random & 55.66 & 62.44 & 66.78 & 67.42 & 67.62 & 70.31 \\
        \textcolor{lightgray}{Subset size} & \textcolor{lightgray}{4.85\%} & \textcolor{lightgray}{9.78\%} & \textcolor{lightgray}{14.78\%} & \textcolor{lightgray}{19.67\%} & \textcolor{lightgray}{25.33\%} & \textcolor{lightgray}{30.23\%} \\
        FreeSel & 56.03 & 64.73 & 66.97 & \textbf{69.87} & 70.57 & 70.93 \\
        \textcolor{lightgray}{Subset size} & \textcolor{lightgray}{4.79\%} & \textcolor{lightgray}{9.71\%} & \textcolor{lightgray}{14.52\%} & \textcolor{lightgray}{19.52\%} & \textcolor{lightgray}{23.74\%} & \textcolor{lightgray}{29.39\%} \\
        K-Centers & 58.71 & 64.29 & 66.96 & 67.58 & \textbf{70.58} & 71.22 \\
        \textcolor{lightgray}{Subset size} & \textcolor{lightgray}{3.98\%} & \textcolor{lightgray}{8.27\%} & \textcolor{lightgray}{13.24\%} & \textcolor{lightgray}{17.72\%} & \textcolor{lightgray}{22.16\%} & \textcolor{lightgray}{27.15\%} \\
        Prototypes & 50.24 & 56.37 & 61.40 & 62.38 & 64.60 & \textbf{71.82} \\
        \textcolor{lightgray}{Subset size} & \textcolor{lightgray}{5.38\%} & \textcolor{lightgray}{10.85\%} & \textcolor{lightgray}{16.05\%} & \textcolor{lightgray}{21.24\%} & \textcolor{lightgray}{26.38\%} & \textcolor{lightgray}{31.89\%} \\
        UP-DP & 55.62 & 60.83 & 64.56 & 64.94 & 65.99 & 68.93 \\
        \textcolor{lightgray}{Subset size} & \textcolor{lightgray}{4.41\%} & \textcolor{lightgray}{9.49\%} & \textcolor{lightgray}{14.92\%} & \textcolor{lightgray}{20.21\%} & \textcolor{lightgray}{25.26\%} & \textcolor{lightgray}{29.57\%} \\
        HACON & 54.64 & 62.82 & 65.20 & 68.10 & 69.60 & 70.10 \\
        \textcolor{lightgray}{Subset size} & \textcolor{lightgray}{4.73\%} & \textcolor{lightgray}{9.83\%} & \textcolor{lightgray}{14.85\%} & \textcolor{lightgray}{19.72\%} & \textcolor{lightgray}{25.40\%} & \textcolor{lightgray}{30.18\%} \\
        Retrieval & 54.61 & 62.80 & 65.16 & 66.65 & 66.74 & 68.74 \\
        \textcolor{lightgray}{Subset size} & \textcolor{lightgray}{4.94\%} & \textcolor{lightgray}{9.72\%} & \textcolor{lightgray}{14.75\%} & \textcolor{lightgray}{19.68\%} & \textcolor{lightgray}{25.37\%} & \textcolor{lightgray}{30.25\%} \\
        ELFS & 52.87 & 61.51 & 65.40 & 68.34 & 69.51 & 70.23 \\
        \textcolor{lightgray}{Subset size} & \textcolor{lightgray}{4.80\%} & \textcolor{lightgray}{9.70\%} & \textcolor{lightgray}{14.72\%} & \textcolor{lightgray}{19.62\%} & \textcolor{lightgray}{25.38\%} & \textcolor{lightgray}{30.26\%} \\
        OFDS & \textbf{61.02} & \textbf{67.05} & \textbf{68.67} & 69.34 & 70.24 & 71.62 \\
        \textcolor{lightgray}{Subset size} & \textcolor{lightgray}{5.02\%} & \textcolor{lightgray}{10.07\%} & \textcolor{lightgray}{15.06\%} & \textcolor{lightgray}{19.96\%} & \textcolor{lightgray}{24.74\%} & \textcolor{lightgray}{29.45\%} \\
        \hline
    \end{tabular}%
    }
    \vspace{0.2cm}
    \caption{\textbf{Semantic Segmentation Results (mIoU) on Cityscapes}}
    \label{tab:res_cs_seg}
\end{table}

\begin{table}
    \centering
    \resizebox{.6\columnwidth}{!}{
    \begin{tabular}{|l|c c c c c c|}
        \hline
        & \multicolumn{6}{c|}{Target Subset Size} \\
        Method & 5\% & 10\% & 15\% & 20\% & 25\% & 30\% \\
        \hline
        Random & 43.20 & 44.14 & 44.61 & 45.16 & 46.06 & 46.37 \\
        \textcolor{lightgray}{Subset size} & \textcolor{lightgray}{5.22\%} & \textcolor{lightgray}{10.23\%} & \textcolor{lightgray}{13.87\%} & \textcolor{lightgray}{19.62\%} & \textcolor{lightgray}{25.80\%} & \textcolor{lightgray}{30.45\%} \\
        FreeSel & 41.13 & 44.51 & 45.48 & 45.77 & 46.24 & 47.01 \\
        \textcolor{lightgray}{Subset size} & \textcolor{lightgray}{5.20\%} & \textcolor{lightgray}{9.64\%} & \textcolor{lightgray}{13.89\%} & \textcolor{lightgray}{19.24\%} & \textcolor{lightgray}{24.35\%} & \textcolor{lightgray}{28.18\%} \\
        K-Centers & 40.60 & 43.87 & 44.63 & 45.63 & 45.94 & 46.29 \\
        \textcolor{lightgray}{Subset size} & \textcolor{lightgray}{5.29\%} & \textcolor{lightgray}{9.81\%} & \textcolor{lightgray}{14.87\%} & \textcolor{lightgray}{20.86\%} & \textcolor{lightgray}{27.31\%} & \textcolor{lightgray}{30.14\%} \\
        Prototypes & 37.42 & 40.60 & 41.31 & 42.72 & 43.91 & 44.56 \\
        \textcolor{lightgray}{Subset size} & \textcolor{lightgray}{5.29\%} & \textcolor{lightgray}{9.81\%} & \textcolor{lightgray}{14.87\%} & \textcolor{lightgray}{20.86\%} & \textcolor{lightgray}{27.31\%} & \textcolor{lightgray}{30.14\%} \\
        ELFS & 42.41 & 44.35 & 44.83 & 45.89 & 46.29 & 46.70 \\
        \textcolor{lightgray}{Subset size} & \textcolor{lightgray}{5.12\%} & \textcolor{lightgray}{10.22\%} & \textcolor{lightgray}{15.34\%} & \textcolor{lightgray}{20.49\%} & \textcolor{lightgray}{25.40\%} & \textcolor{lightgray}{30.41\%} \\
        UP-DP & 43.03 & 44.59 & 45.02 & 45.21 & 45.66 & 46.48 \\
        \textcolor{lightgray}{Subset size} & \textcolor{lightgray}{5.00\%} & \textcolor{lightgray}{10.44\%} & \textcolor{lightgray}{15.93\%} & \textcolor{lightgray}{21.11\%} & \textcolor{lightgray}{25.93\%} & \textcolor{lightgray}{30.53\%} \\
        HaCON & 42.91 & 44.42 & 45.40 & 45.73 & 45.50 & 46.38 \\
        \textcolor{lightgray}{Subset size} & \textcolor{lightgray}{5.09\%} & \textcolor{lightgray}{10.18\%} & \textcolor{lightgray}{15.33\%} & \textcolor{lightgray}{20.52\%} & \textcolor{lightgray}{25.45\%} & \textcolor{lightgray}{30.39\%} \\
        T2I Retrieval & 43.22 & 43.53 & 44.27 & 45.01 & 45.79 & 46.13 \\
        \textcolor{lightgray}{Subset size} & \textcolor{lightgray}{5.06\%} & \textcolor{lightgray}{9.95\%} & \textcolor{lightgray}{15.51\%} & \textcolor{lightgray}{20.89\%} & \textcolor{lightgray}{26.04\%} & \textcolor{lightgray}{31.10\%} \\
        OFDS & \textbf{44.10} & \textbf{45.39} & \textbf{46.12} & \textbf{46.51} & \textbf{47.16} & \textbf{47.35} \\
        \textcolor{lightgray}{Subset size} & \textcolor{lightgray}{5.12\%} & \textcolor{lightgray}{9.93\%} & \textcolor{lightgray}{14.11\%} & \textcolor{lightgray}{20.29\%} & \textcolor{lightgray}{24.94\%} & \textcolor{lightgray}{29.67\%} \\
        \hline
    \end{tabular}%
    }
    \vspace{0.2cm}
    \caption{\textbf{Data Selection on LoveDA.}}
    \label{tab:res_loveda_seg}
\end{table}

\end{document}